%% file: nmm.tex
\documentclass[]{youtu}

\usepackage{mathpazo}
\usepackage{graphicx}
\usepackage{natbib}
\usepackage{times}
\usepackage{latexsym}
\usepackage[T1]{fontenc}
\usepackage{xcolor}
\usepackage{enumitem}
\usepackage{lipsum} 
\usepackage[utf8]{inputenc}
\usepackage{booktabs}
\usepackage{multirow}
\usepackage{amssymb}
\usepackage{adjustbox}
\usepackage{amsmath}
\usepackage{amssymb}
\usepackage{tabularx}

\usepackage{wrapfig}
\newcolumntype{Y}{>{\raggedright\arraybackslash}X}
\usepackage{threeparttable}
\usepackage{graphicx}
\usepackage{tcolorbox}
\usepackage{graphicx}
\usepackage{forest}
\useforestlibrary{edges}
\tcbuselibrary{skins, breakable, theorems, listings}
\tcbuselibrary{hooks}
\usepackage[table]{xcolor}
\usepackage{cuted}
\usepackage{placeins}
\usepackage{stfloats}
\usepackage{circledsteps}
\title{Toward Native Multimodal Modeling: A Roadmap}

\author{
  Siyu An\textsuperscript{1 $\clubsuit$}, 
  Junru Lu\textsuperscript{1 $\clubsuit$}, 
  Junnan Dong\textsuperscript{1 $\clubsuit~\heartsuit$},\\ 
  Qiufeng Wang\textsuperscript{1}, 
  Yinghui Li\textsuperscript{1}, 
  Weizhi Fei\textsuperscript{2}, 
  Zichao Yu\textsuperscript{3}, 
  Zheng Yuan\textsuperscript{1},\\ 
  Biao Liu\textsuperscript{1}, 
  Haopeng Wang\textsuperscript{1},
  Renzhao Liang\textsuperscript{1}, 
  Yixuan Yang\textsuperscript{4}, 
  Yunhang Shen\textsuperscript{1}, 
  Bo Ke\textsuperscript{1},\\
  Keyu Chen\textsuperscript{1},
  Linhao Luo\textsuperscript{5}, 
  Difan Zou\textsuperscript{3},
  Xiao Huang\textsuperscript{6}, 
  Di Yin\textsuperscript{1}, 
  Ruizhi Qiao\textsuperscript{1}, 
  Xing Sun\textsuperscript{1}
}

\affiliation{\textsuperscript{1}Tencent Youtu Lab\quad\textsuperscript{2}Tsinghua University\quad\textsuperscript{3}The University of Hong Kong\\ \textsuperscript{4}University of Warwick\quad\textsuperscript{5}Monash University\quad\textsuperscript{6}The Hong Kong Polytechnic University}

\sourcecode{https://nmm-roadmap.github.io}
\correspondence{$\clubsuit$~Equal Contribution; $\heartsuit$~Corresponding Author.}

\begin{document}

\abstract{Multimodal modeling represents a vital step from modality-agnostic reasoning toward world modeling. While early approaches predominantly rely on late-fusion that assembles encoders and frozen language backbones with output heads, recent efforts have shifted the paradigm toward native multimodal modeling (NMM) with the intrinsic integration of modalities for superior multimodal performance. Despite its potential, the design space of native architectures remains insufficiently defined. In this paper, we present the community with a formalized roadmap for this transition. Specifically, we formally define the architectural nativity, distinguishing \textit{mid-fusion} and \textit{early-fusion} from non-native paradigms. We further organize the existing native models through the lens of input-output duality into three categories: \((i)\) \textbf{Multi-to-Text} for cross-modal comprehension with text-only output; \((ii)\) \textbf{Multi-to-Target} for scenario-oriented generation, e.g., image, audio and video generation, and \((iii)\) \textbf{Multi-to-Multi} for unified modeling with symmetric input-output. We deliver a comprehensive and industrial-grade investigation into the transition toward the definitive NMM framework, where understanding and generation seamlessly coexist within a unified transformer paradigm. We systematically unpack the end-to-end pipeline from industrial perspectives from architectural coordination, massive data curation, to full-stack training recipes, inference \& deployment, and the comprehensive evaluation for truly native modeling.}

\maketitle
\vspace{1cm}
\begin{figure}[ht!]
    \centering
    \hspace*{-0.36cm} 
    \includegraphics[width=1.04\linewidth]{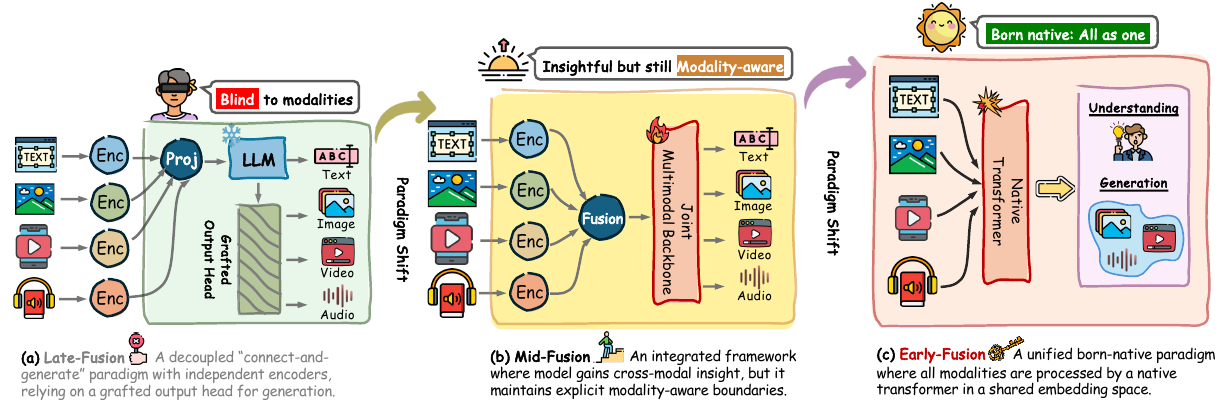}
    \caption{A sketched overview of the evolutionary landscape in the paradigms of multimodal modeling. In the final stage, the model achieves a born-native state where all modalities are processed within a unified transformer space, facilitating symmetric multi-to-multi understanding and generation.}
    \label{fig:placeholder}
\end{figure}

\tableofcontents

\vspace{-.1em}

\input{intro}
\input{formal}
\input{arch}
\input{data}
\input{train}
\input{inference}
\input{eval}
\input{future}

\setcitestyle{numbers,square}
\setcitestyle{square,numbers,comma}
\bibliography{youtu_bib}

\end{document}

%% file: intro.tex
\section{Introduction}
\label{sec:intro}
Large language models (LLMs) have increasingly demonstrated their capabilities for social good, showing remarkable performance in comprehension and reasoning~\cite{lu2025youtu, dong2024clrbenchevaluatinglargelanguage,liu2024deepseek,bai2023qwen}. Despite this success, LLMs remain fundamentally limited by a text-only interface to both users and the real world~\cite{bai2025qwen3vltechnicalreport,tong2026beyond,InternVL3.5_2025}. Consequently, the understanding is inherently indirect, lacking grounding in the rich sensory signals that characterize real-world environments. The quest for artificial general intelligence thus necessitates a transition from modality-agnostic text processors toward holistic world models~\cite{caffagni2024revolution,Yin_2024_survey,dong2024modality}. Multimodal modeling represents a pivotal leap in this trajectory, aiming to transform LLMs into versatile agents through unified cross-modal understanding and generation~\cite{zhao2025unified, cui2025emu35nativemultimodalmodels}. While early research predominantly focused on late-fusion paradigms, e.g., LLaVa~\cite{zhang2024llava}, DeepSeek-VL~\cite{lu2024deepseekvl} and Qwen-Image~\cite{wu2025qwenimage}, characterized by modularly assembling pre-trained encoders with frozen language backbones through shallow projectors. These non-native compositions often suffer from a fundamental blindness to raw sensory signals. Such architectural decoupling limits the depth of cross-modal interaction, preventing the model from achieving true synergy across disparate data forms.

\begin{table*}[!htb]
\scriptsize
\centering
\begin{adjustbox}{width=\textwidth}
\begin{tabular}{llc|cccc|cccc}
\toprule
\rowcolor[HTML]{EAF2F8}
& & & \multicolumn{4}{c|}{\textbf{Input Modalities}} & \multicolumn{4}{c}{\textbf{Output Modalities}} \\
\cline{4-11}
\rowcolor[HTML]{EAF2F8}

\multirow{-2}{*}{\textbf{Model Name}} & \multirow{-2}{*}{\textbf{Date}} & \multirow{-2}{*}{\textbf{Params of Flagship}} & \textbf{Text} & \textbf{Img} & \textbf{Aud} & \textbf{Vid} & \textbf{Text} & \textbf{Img} & \textbf{Aud} & \textbf{Vid} \\
\midrule
\multicolumn{11}{c}{\textit{\textbf{Multi-to-Text Unimodal Generation}}}\\
\midrule
MiniCPM-V-4.6~\cite{yu2025minicpmv45cookingefficient}& 2026.05 & 1B & \checkmark & \checkmark & -- & \checkmark & \checkmark & -- & -- & -- \\
Nemotron3-Nano-Omni~\cite{nvidia2026nemotron3nanoomni} & 2026.04 & 30BA3B & \checkmark & \checkmark & \checkmark & \checkmark & \checkmark & -- & -- & -- \\
MiMo-V2.5~\cite{xiaomi2026mimov25} & 2026.04 & 310BA15B & \checkmark & \checkmark & \checkmark & \checkmark & \checkmark & -- & -- & -- \\
Qwen3.6~\cite{qwen36_35b_a3b} & 2026.04 & 27B & \checkmark & \checkmark & -- & \checkmark & \checkmark & -- & -- & -- \\
Gemma-4-31B~\cite{Gemma4Team2026} & 2026.04 & 31B & \checkmark & \checkmark & -- & \checkmark & \checkmark & -- & -- & -- \\
Gemma-4-E4B~\cite{Gemma4Team2026} & 2026.04 & 4.5B(8B) & \checkmark & \checkmark & \checkmark & \checkmark & \checkmark & -- & -- & -- \\
Kimi K2.5~\cite{KimiK2_5_2026} & 2026.01 & 1TA32B & \checkmark & \checkmark & -- & \checkmark & \checkmark & -- & -- & -- \\
GLM-5V-Turbo~\cite{GLM5VTurbo2026} & 2026.04 & 744BA40B & \checkmark & \checkmark & -- & \checkmark & \checkmark & -- & -- & -- \\
Llama-4-Scout~\cite{Adcock2026TheL4} & 2025.04 & 109BA17B & \checkmark & \checkmark & -- & \checkmark & \checkmark & -- & -- & -- \\
Llama-4-Maverick~\cite{Adcock2026TheL4} & 2025.04 & 400BA17B & \checkmark & \checkmark & -- & \checkmark & \checkmark & -- & -- & -- \\
InternVL-3.5~\cite{InternVL3.5_2025} & 2025.08 & 241BA28B & \checkmark & \checkmark & -- & \checkmark & \checkmark & -- & -- & -- \\
Qwen3-VL~\cite{bai2025qwen3vltechnicalreport} & 2025.09 & 235BA22B & \checkmark & \checkmark & -- & \checkmark & \checkmark & -- & -- & -- \\
Qwen2.5-VL~\cite{bai2025qwen25vltechnicalreport} & 2025.02 & 72B & \checkmark & \checkmark & -- & \checkmark & \checkmark & -- & -- & -- \\
CogVLM~\cite{wang2023cogvlm} & 2023.11 & 17B & \checkmark & \checkmark & -- & -- & \checkmark & -- & -- & -- \\
Video-LLaVA~\cite{lin2023video} & 2023.11 & 13B & \checkmark & \checkmark & -- & \checkmark & \checkmark & -- & -- & -- \\
Qwen-Audio~\cite{chu2023qwenaudioadvancinguniversalaudio} & 2023.11 & 13B & \checkmark & \checkmark & -- & \checkmark & \checkmark & -- & -- & -- \\

\midrule
\multicolumn{11}{c}{\textit{\textbf{Multi-to-Target Scenario-based Generation}}}\\
\midrule
HiDream-O1-Image~\cite{hidreamolimage}& 2026.05 & 8B & \checkmark & \checkmark & -- & -- & -- & \checkmark & -- & -- \\
OmniVoice~\cite{zhu2026omnivoiceomnilingualzeroshottexttospeech}& 2026.04 & 0.8B & \checkmark & \checkmark & \checkmark & -- & -- & -- & \checkmark & -- \\
LTX-2.3~\cite{LightricksLTX2_2026} & 2026.03 & 19B & \checkmark & \checkmark & \checkmark & \checkmark & -- & -- & \checkmark & \checkmark \\
Ming-Flash-Omni-2.0~\cite{ai2026mingflashomnisparseunifiedarchitecture}& 2026.02 & 100BA6B & \checkmark & \checkmark & \checkmark  & \checkmark & \checkmark & \checkmark & \checkmark & -- \\
MiniCPM-o-4.5~\cite{cui2026minicpm} & 2026.02 & 9B & \checkmark & \checkmark & \checkmark & \checkmark & \checkmark & -- & \checkmark & -- \\
Kling-Omni~\cite{klingteam2025klingomnitechnicalreport} & 2025.12 & - & \checkmark & \checkmark & -- & \checkmark & -- & -- & -- & \checkmark \\
HunyuanVideo-1.5~\cite{wu2025hunyuanvideo15technicalreport} & 2025.12 & 8.3B & \checkmark & -- & -- & -- & -- & -- & -- & \checkmark \\
LTX-2.2~\cite{LightricksLTX2_2026} & 2025.10 & 19B & \checkmark & \checkmark & \checkmark &  \checkmark & -- & -- & \checkmark & \checkmark \\
Qwen3-Omni~\cite{Qwen3Omni2025} & 2025.09 & 30BA3B & \checkmark & \checkmark & \checkmark  & \checkmark & \checkmark & -- & \checkmark & -- \\
Wan2.2-T2V-A14B~\cite{wan22_2025} & 2025.07 & 27BA14B & \checkmark & -- & -- & -- & -- & -- & -- & \checkmark \\
Wan2.2-TI2V-5B~\cite{wan22_2025} & 2025.07 & 5B & \checkmark & \checkmark & -- & -- & -- & -- & -- & \checkmark \\
Seedream3.0~\cite{gao2025seedream30technicalreport} & 2025.04 & 12B & \checkmark & -- & -- & -- & -- & \checkmark & -- & -- \\
\midrule
\multicolumn{11}{c}{\textit{\textbf{Multi-to-Multi Symmetric Modeling}}}\\
\midrule
Lance~\cite{fu2026lanceunifiedmultimodalmodeling} & 2026.05 & 3B & \checkmark & \checkmark & -- & \checkmark & \checkmark & \checkmark & -- & \checkmark \\
Mamoda2.5~\cite{shi2026mamoda25enhancingunifiedmultimodal} & 2026.05 & 25BA3B & \checkmark & \checkmark & -- & -- & \checkmark & \checkmark & -- & -- \\
TUNA-2~\cite{liu2026tuna2pixelembeddingsbeat} & 2026.04 & 7B & \checkmark & \checkmark & -- & -- & \checkmark & \checkmark & -- & -- \\
SenseNova-U1-8B-MoT$^{*}$~\cite{diao2026sensenovau1unifyingmultimodalunderstanding} & 2026.04 & 8B(18B) & \checkmark & \checkmark & -- & -- & \checkmark & \checkmark & -- & -- \\
LLaDA2.0-Uni$^{*}$~\cite{ai2026llada20uniunifyingmultimodalunderstanding} & 2026.04 & 16BA1B & \checkmark & \checkmark & \checkmark & -- & \checkmark & \checkmark & \checkmark & -- \\
LongCat-Next$^{*}$~\cite{MeituanLongCat2026} & 2026.04 & 68.5BA3B & \checkmark & \checkmark & \checkmark & -- & \checkmark & \checkmark & \checkmark & -- \\
Emu3.5$^{*}$~\cite{cui2025emu35nativemultimodalmodels} & 2025.10 & 34.1B & \checkmark & \checkmark & -- & \checkmark & \checkmark & \checkmark & -- & \checkmark \\
Show-o2~\cite{xie2025showo2improvednativeunified} & 2025.09 & 7B & \checkmark & \checkmark & -- & \checkmark & \checkmark & \checkmark & -- & \checkmark \\
BAGEL~\cite{BAGEL7B2025} & 2025.05 & 14BA7B & \checkmark & \checkmark & -- & \checkmark & \checkmark & \checkmark & -- & \checkmark \\
OneCAT$^{*}$~\cite{OneCAT3B2025} & 2025.09 & 9BA3B & \checkmark & \checkmark & \checkmark & -- & \checkmark & \checkmark & \checkmark & -- \\
Janus-Pro$^{*}$~\cite{DeepSeekJanusPro2025} & 2025.01 & 7B & \checkmark & \checkmark & -- & -- & \checkmark & \checkmark & -- & -- \\
Moshi$^{*}$~\cite{defossez2024moshispeechtextfoundationmodel} & 2024.09 & 7B & \checkmark & -- & \checkmark & -- & \checkmark & -- & \checkmark & -- \\
Transfusion~\cite{zhou2024transfusion} & 2024.08 & 7B & \checkmark & \checkmark & -- & -- & \checkmark & \checkmark & -- & -- \\
Chameleon$^{*}$~\cite{team2024chameleon} & 2024.05 & 34B & \checkmark & \checkmark & -- & -- & \checkmark & \checkmark & -- & -- \\
AnyGPT$^{*}$~\cite{zhan2024anygpt} & 2024.02 & 7B & \checkmark & \checkmark & \checkmark & -- & \checkmark & \checkmark & \checkmark & -- \\
\bottomrule
\end{tabular}
\end{adjustbox}
\caption{Comprehensive comparison of recently released Native Multimodal Models. We limit our comparison to open-source models or technical reports with verified architecture and parameter transparency. *Indicates models employing the discrete unified scheme. () denotes effective(total) parameter counts of special architectural designs.}
\label{tab:multimodal_evolution}
\end{table*}

In response to these limitations, recent efforts have catalyzed a paradigm shift toward native multimodal modeling (NMM)~\cite{KimiK2_5_2026,cui2025emu35nativemultimodalmodels, klingteam2025klingomnitechnicalreport,BAGEL7B2025,DeepSeekJanusPro2025,OneCAT3B2025,xie2025showo2improvednativeunified}, where multiple modalities are intrinsically integrated into the core architecture. Unlike their predecessors, native models seek to internalize multimodal capabilities through joint multimodal backbones or unified transformer spaces, enabling more principled and robust cross-modal intelligence. However, as the field rapidly expands with diverse architectural choices ranging from deep feature injection to unified tokenization, the design space for NMM remains fragmented and insufficiently defined. This lack of formalization hinders the community's ability to evaluate the degree of nativity in emergent models and complicates the selection of optimal architectures for specific downstream tasks. There is a pressing need for a structured roadmap to formalize the transition from modular assembly to native convergence, clarifying the taxonomies that distinguish varying levels of architectural integration.

In this paper, we provide a comprehensive formalization of the NMM landscape by distinguishing two primary native regimes based on their integration depth: mid-fusion and early-fusion. We categorize mid-fusion models as a naturally interacted regime, where features from distinct encoders are injected into a joint multimodal backbone, allowing the model to be insightful across modalities while maintaining explicit modality-aware boundaries. This category is historical yet foundational, represented by classical pioneers such as CogVLM~\cite{wang2023cogvlm} and Qwen-Audio~\cite{chu2023qwenaudioadvancinguniversalaudio}. This paradigm has evolved into massive state-of-the-art architectures, including Qwen2.5-VL~\cite{bai2025qwen25vltechnicalreport}, Qwen3-VL~\cite{bai2025qwen3vltechnicalreport}, and InternVL-3.5~\cite{InternVL3.5_2025}, culminating in scaling attempts like GLM-5V-Turbo~\cite{GLM5VTurbo2026} and Kimi K2.5~\cite{KimiK2_5_2026}. Yet, early-fusion represents a native convergent regime where all modalities are modeled within a unified embedding space via one unified backbone. This born-native design, explored by Transfusion~\cite{zhou2024transfusion}, Chameleon~\cite{team2024chameleon}, and AnyGPT~\cite{zhan2024anygpt}, achieves omnipresent synergy by treating all modalities equivalently.

Building upon this structural taxonomy, we organize the existing NMM ecosystem through the lens of input-output duality into three functional categories to capture the full spectrum of modality flows. \((i)\) The first category, Multi-to-Text (M2T) unimodal generation, leverages native scaling to ground cross-modal inputs into purely linguistic responses for reasoning. This front is represented by dense models such as Nemotron3-Nano-Omni~\cite{nvidia2026nemotron3nanoomni}, MiMo-V2.5~\cite{xiaomi2026mimov25} and MiniCPM-V-4.6~\cite{yu2025minicpm}; \((ii)\) The second category, Multi-to-Target (M2G) scenario-based generation, bypasses traditional post-hoc generation decoders by synthesizing modality-specific outputs directly through native representations, which enables temporal and acoustic coherence in complex environments. Key milestones in this space include advanced video generators such as Wan2.2-T2V-A14B~\cite{wan22_2025}, HunyuanVideo-1.5~\cite{wu2025hunyuanvideo15technicalreport}, and Kling-Omni~\cite{klingteam2025klingomnitechnicalreport}, alongside speech-centric native frameworks like OmniVoice~\cite{zhu2026omnivoice}, MiniCPM-o-4.5~\cite{cui2026minicpm}, and Seedream3.0~\cite{gao2025seedream30technicalreport}; \((iii)\) The final and most comprehensive category is Multi-to-Multi (M2M) symmetric modeling, which establishes a symmetric input-output paradigm where understanding and generation naturally coexist within a single network. Early formulations in this direction, such as Moshi~\cite{defossez2024moshi} and Emu3.5~\cite{cui2025emu35nativemultimodalmodels}, have laid the foundation for complex architectural explorations. This includes interleaved modeling via BAGEL-7B~\cite{BAGEL7B2025}, OneCAT-3B~\cite{OneCAT3B2025}, and Show-o2-7B~\cite{xie2025showo2improvednativeunified}, as well as bidirectional unification in Janus-Pro~\cite{DeepSeekJanusPro2025}, TUNA-2~\cite{liu2026tuna2pixelembeddingsbeat}, and Mamoda2.5~\cite{shi2026mamoda25enhancingunifiedmultimodal}. \\\\
\textbf{Contributions}.
\begin{itemize}[leftmargin=*]
    \item \textbf{\texttt{Problem Formalization}}. We first present the formal, systemic definition of NMM, establishing a principled structural taxonomy based on \textbf{integration depth}, i.e. \textit{\{mid-, early-\} fusion} and \textbf{input-output duality}, i.e., \textit{Multi-to-\{Text, Target, Multi\}}  to clarify the fragmented design space.
    \item \textbf{\texttt{Technological Roadmap}}. We systematically analyze the full lifecycle of NMM, extracting and characterizing the core modal bottlenecks and cross-cutting technical solutions across architectural designs (\S\ref{arch}), data curricula (\S\ref{data}), training strategies (\S\ref{sec:training}), inference deployment (\S\ref{inference}), and holistic evaluation (\S\ref{eval}).
    \item \textbf{\texttt{Future Outlook}}. We carefully provide empirical insights from state-of-the-art implementations and paradigms to deliver a visionary projection of future trajectories, suggesting crucial strategic directions for the evolution toward advanced NMM. 
\end{itemize}

%% file: formal.tex
\begin{figure}[!htbp]
    \centering
    \vspace{-10mm}
    \includegraphics[width=1\linewidth]{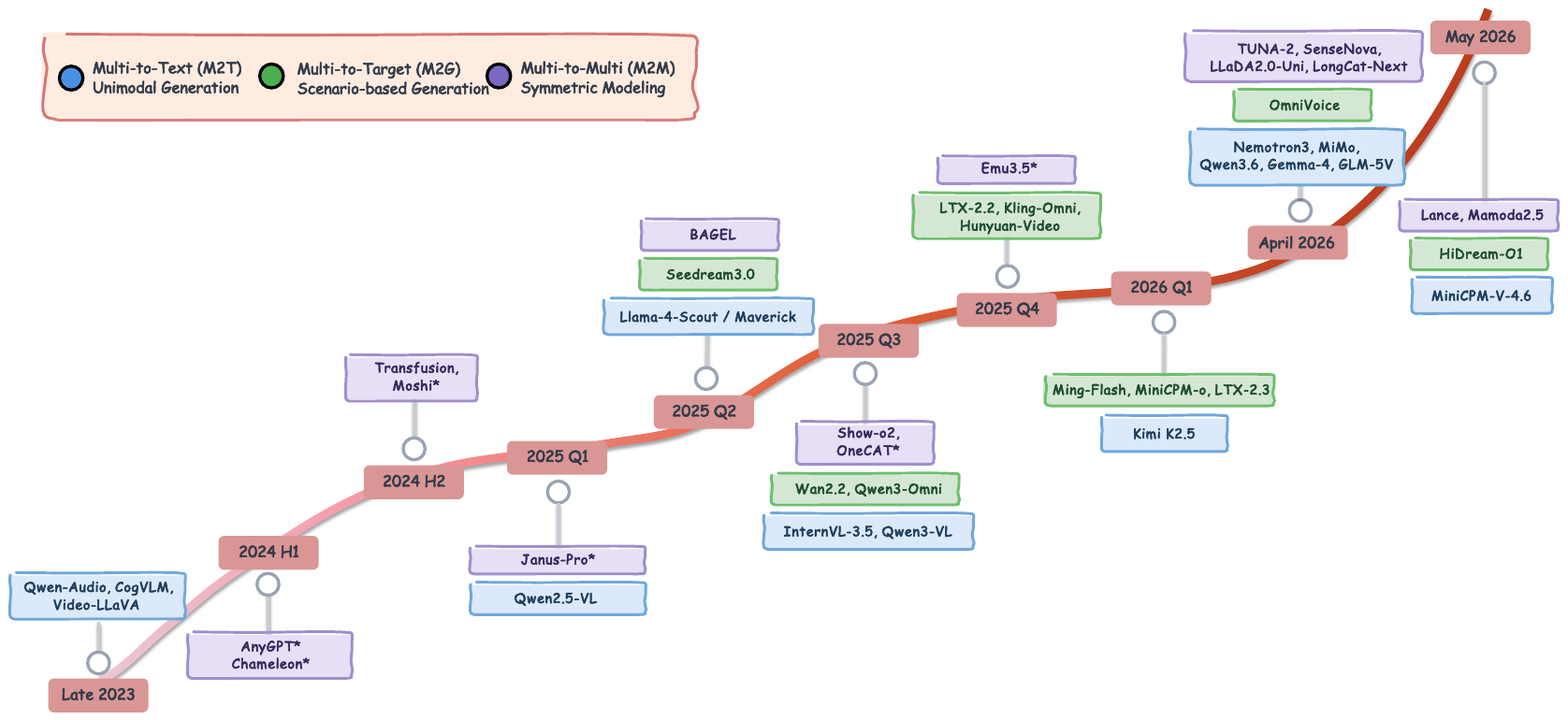}
    \vspace{-20mm}
    \caption{Evolutionary timeline and functional taxonomy of Native Multimodal Foundation Models (2023–2026). The upward trajectory charts the historical progression from early mid-fusion alignment to the early-fusion methods, i.e., born-native transformer.}
    \label{fig:timeline}
\end{figure}

\section{Task Formalization}
\label{sec:formal}

\subsection{What is Native? Formalizing Cross-modal Fusion Nativity}
To establish a rigorous boundary for native multimodal modeling, we formalize the architectural transition through a set of functional operators. Let the input modality set be $\mathcal{M} = \{m_1, m_2, \dots, m_n\}$. We denote $E_i$ as modality-specific encoders, $\mathcal{P}_i$ as projection/alignment layers, and $\mathcal{T}$ as a unified tokenization operator. Typically, the \textbf{Late-Fusion} paradigm, i.e., modular assembling~\cite{zhang2024llava,lu2024deepseekvl,wu2025qwenimage} is defined as $\mathcal{F}_{\text{late}} = \mathcal{G}\!\left( \text{LLM}\big( \{\mathcal{P}_i(E_i(m_i))\}_{i=1}^n \big) \right)$, where the backbone remains \textit{blind} to raw sensory signals and relies on a grafted output head $\mathcal{G}$.

In this paper, we explicitly exclude such post-hoc alignment schemes from the scope of native modeling. Instead, we define NMM as a paradigm where multimodal synergy is an intrinsic architectural property, categorized into the following two regimes:

\textbf{Mid-Fusion}: The first stage of transition to NMM, defined as $\mathcal{F}_{mid} = \text{Backbone}(\mathcal{C}(E_1(m_1), \dots, E_n(m_n)))$, where $\mathcal{C}$ denotes a cross-modal alignment or injection operator (e.g., cross-attention or deeply stacked adapters). In this regime, multimodal features are injected into the intermediate layers of a Joint Multimodal Backbone. While the model becomes \textit{insightful} regarding cross-modal correlations, it remains inherently \textit{modality-aware} due to the explicit architectural boundaries and structural asymmetry between the upstream encoders $E_i$ and the central backbone.

\textbf{Early-Fusion}: Representing the optimal pinnacle of native synergy, this paradigm is defined as $\mathcal{F}_{early} = \text{Transformer}(\bigcup_{i} \mathcal{T}(m_i))$. By bypassing independent, frozen encoders entirely, all modalities are mapped by a unified operator $\mathcal{T}$ into a single, shared embedding space from the outset. This \textit{born-native} architecture achieves a deep synergy, acting as an ideally unified world model that treats all modalities as fundamentally equivalent tokens.

\begin{wrapfigure}{l}{0.5\textwidth}
    \centering
    \includegraphics[width=\linewidth]{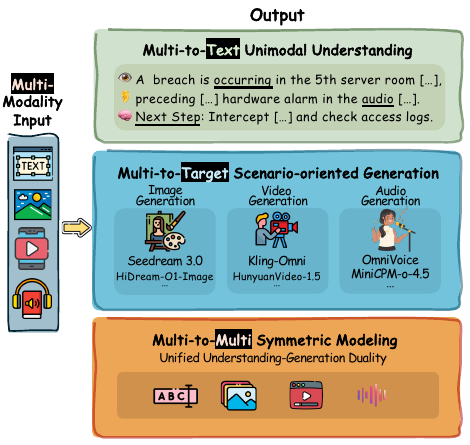}
    \caption{Illustrative examples of three primary NMM architectures considering input-output duality.}
    \label{fig:multi}
\end{wrapfigure}
\subsection{How Native? Taxonomy by Architectural Symmetry}
Beyond the depth of architectural integration, the degree of native capability is inherently bounded by the input-output modality flow. We formalize this taxonomy from the perspective of modality duality and structural symmetry, mapping the native landscape into three progressive paradigms.

\textbf{Multi-to-Text (M2T) Unimodal Generation}: This paradigm represents an asymmetric comprehension scheme, formalized as $\mathcal{F}_{M2T}: \mathcal{M} \rightarrow T$, where $T \in \mathcal{M}$ represents the text modality. In this configuration, whether utilizing a Mid-Fusion joint backbone or an Early-Fusion transformer, the model ingests arbitrary interleaved cross-modal streams to perform dense reasoning, ultimately collapsing the multimodal hidden states into a single linguistic space. The optimization bottleneck primarily lies in cross-modal alignment and perceptual grounding rather than textual synthesis.

\textbf{Multi-to-Target (M2G) Scenario-based Generation}: This paradigm shifts the architectural focus toward asymmetric generation, formalized as $\mathcal{F}_{M2G}: \mathcal{M} \rightarrow y_k$, where $y_k \in \mathcal{M}$ represents a single target non-textual modality (e.g., video voxels and audio waveforms). Native M2G architectures establish unified output pathways that directly decode the target modality from the core native hidden representations. This ensures that the generated targets retain high semantic coherence with the multimodal prompt, underscoring the superiority of unified output pathways over non-native grafting schemes.

\textbf{Multi-to-Multi (M2M) Symmetric Modeling}: Representing the ultimate phase of native convergence, this paradigm establishes a fully symmetric input-output flow, formalized as $\mathcal{F}_{M2M}: \mathcal{M}_{in} \rightarrow \mathcal{M}_{out}$, where both $\mathcal{M}_{in} \subseteq \mathcal{M}$ and $\mathcal{M}_{out} \subseteq \mathcal{M}$ can contain arbitrary combinations of co-existing modalities. In this regime, the concepts of separate perceptors and renderers disappear. The model serves as a unified world modeler where multimodal understanding and token-level next-step generation mutually coexist in a single Transformer. This symmetrical duality eliminates the informational bottlenecks present in asymmetric design, enabling fluid, real-time, any-to-any intelligence.

%% file: arch.tex
\section{Model Architecture}\label{arch}
NMM systems assign distinct functional roles to comprehend and generate different modalities. In this section, we dive into the three aforementioned paradigms as listed in Table~\ref{tab:multimodal_evolution}, outlining the respective technical challenges and approaches. 
The functional categories examined in this section are defined by their input-output modality configurations, whereas the architectural taxonomy of Section~\ref{sec:intro} (mid-fusion vs. early-fusion) captures the depth of cross-modal integration. As these two dimensions are orthogonal, each functional category contains representatives of both fusion paradigms. We annotate individual architectures as Mid-fusion or Early-fusion throughout.

\subsection{M2T Unimodal Generation}
\label{sec:multi-to-text} 
M2T models take multimodal inputs (\emph{text}, \emph{image}, \emph{audio}, \emph{video}) and produce text-only output. This design efficiently converts real-world signals into semantic representations, focusing on complex comprehension and reasoning. 

\subsubsection{Image Comprehension}
The integration of vision and text is the primary focus of multimodal comprehension models. Currently the core barriers are centered around three key challenges: 
1) \textit{Modality Unification}
2) \textit{Multi-image Reasoning}
3) \textit{Multi-scale Encoding}. 

\paragraph{Modality Unification.} Unifying disparate modalities natively into a single computational space often introduces architectural tensions and modality competition during joint training. To mitigate information loss from discrete quantization, current state-of-the-art models primarily pursue continuous projection routes. \textbf{$(i)$ Vision-Encoder-Based Fusion} remains the dominant paradigm, utilizing dedicated modules to project features into the LLM's latent space. Llama-4-Scout/Maverick utilizes an enhanced vision encoder to project images into continuous patch embeddings, enabling joint processing from the earliest transformer layers. Similarly, Kimi K2.5 employs a MoonViT encoder to transform images into embeddings that flow through a shared sparse MoE backbone, while Gemma-4-31B utilizes a hybrid-attention architecture to interleave continuous soft tokens with text. \textbf{$(ii)$ Unified Stream Mapping} seeks to reduce architectural fragmentation. Qwen3.6 represents this direction by treating all modalities as a unified token stream within a single transformer, while Nemotron3-Nano-Omni utilizes a compact, unified architecture to achieve low-latency cross-modal alignment.

\paragraph{Multi-image Reasoning.} In scenarios involving multiple images or long-form documents, visual tokens can overwhelm the attention, leading to attention saturation and quadratic computational growth. Current foundational models address this through four technical routes: \textbf{$(i)$ Extreme Visual Compression}: Kimi K2.5 and InternVL-3.5 employ the Visual Resolution Router and temporal pooling to reduce visual token counts without losing semantic density. \textbf{$(ii)$ Deep Feature Alignment}: Qwen3-VL and Qwen2.5-VL utilize deep-stack multi-level feature injection to strengthen synergy, while CogVLM maintains a dedicated Visual Expert module to preserve structural integrity. \textbf{$(iii)$ Advanced Positional Encoding}: To maintain spatio-temporal awareness across massive contexts, Llama-4 and Gemma-4-E4B have integrated iRoPE/p-RoPE, ensuring stable retrieval across interleaved sequences. \textbf{$(iv)$ Perception-Reasoning Decoupling}: Models like GLM-5V-Turbo and MiMo-V2.5 implement a thinking mode, which separates raw visual perception from the subsequent heavy-duty logical deduction to minimize latency and hallucination.

\paragraph{Multi-scale Encoding.} To resolve geometric distortion and loss of fine-grained detail in non-standard aspect ratios, models have converged on the following strategies: \textbf{$(i)$ Structure-Aware Tiling}: InternVL-3.5 and MiniCPM-V-4.6 partition high-resolution inputs into dynamic tiles, taking structural identifiers to help the model reconstruct 2D layouts from 1D token streams. \textbf{$(ii)$ Dimension-Decoupled Positional Encoding}: Qwen3-VL and GLM-5V-Turbo utilize 2D-RoPE, decomposing coordinates into $x$ and $y$ components to natively interpret any aspect ratio. \textbf{$(iii)$ Semantic-Driven Resampling}: InternVL-3.5 utilizes a perceiver-based architecture to adaptively compress background patches into a fixed latent space, preventing visual noise from drowning out text signals. \textbf{$(iv)$ Resolution-Agnostic Projection}: Gemma-4-31B and Llama-4-Maverick bypass fixed-grid constraints, allowing seamless reasoning over complex, variable-scale layouts such as ultra-wide tables and long-scroll documents.

\definecolor{root-navy}{HTML}{0F172A}
\definecolor{m2t-solid}{HTML}{1E40AF}
\definecolor{m2t-mid}{HTML}{DBEAFE}
\definecolor{m2t-light}{HTML}{EFF6FF}
\definecolor{m2t-leaf}{HTML}{F8FAFC}
\definecolor{m2g-solid}{HTML}{B45309}
\definecolor{m2g-mid}{HTML}{FEF3C7}
\definecolor{m2g-light}{HTML}{FFFBEB}
\definecolor{m2g-leaf}{HTML}{FAFAF9}
\definecolor{m2m-solid}{HTML}{991B1B}
\definecolor{m2m-mid}{HTML}{FEE2E2}
\definecolor{m2m-light}{HTML}{FEF2F2}
\definecolor{m2m-leaf}{HTML}{FFF9FA}
\definecolor{edge-darkgray}{HTML}{64748B}

\begin{figure*}[!th]
    \centering
    \vspace{-3mm}
    \resizebox{1\textwidth}{!}{
        \begin{forest}
            forked edges,
            for tree={
                grow=east,
                reversed=true,
                anchor=base west,
                parent anchor=east,
                child anchor=west,
                base=left,
                font=\Large,  
                rectangle,
                rounded corners=4pt,
                align=left,
                minimum width=1em,
                edge+={edge-darkgray, line width=1.0pt},
                inner xsep=10pt,
                inner ysep=7pt,
                l sep=12mm,       
                s sep=4.0mm,        
                ver/.style={rotate=90, child anchor=north, parent anchor=south, anchor=center},
            } 
            [ {Native Multimodal LLMs}, fill=root-navy, text=white, draw=root-navy, inner xsep=14pt, inner ysep=9pt, font=\Large\bfseries, ver
                [ {M2T Unimodal\\ Generation \\ (\S3.1)}, fill=m2t-solid, text=white, draw=m2t-solid, font=\Large\bfseries
                    [ {Image\\ Comprehension }, fill=m2t-mid, text=m2t-solid, draw=m2t-solid!60, font=\Large\bfseries
                        [ {Modality Unification}, fill=m2t-light, text=m2t-solid!90, draw=m2t-solid!30
                            [ {Vision-Encoder-Based Fusion}, fill=m2t-leaf, draw=m2t-solid!40
                                [ {Llama-4-Scout/Maverick, Kimi K2.5, Gemma-4-31B}, fill=white, draw=m2t-solid!60, font=\Large\itshape ] 
                            ] 
                            [ {Unified Stream Mapping}, fill=m2t-leaf, draw=m2t-solid!40
                                [ {Qwen3.6, Nemotron3-Nano-Omni}, fill=white, draw=m2t-solid!60, font=\Large\itshape ] 
                            ] 
                        ] 
                        [ {Multi-image Reasoning}, fill=m2t-light, text=m2t-solid!90, draw=m2t-solid!30
                            [ {Extreme Visual Compression}, fill=m2t-leaf, draw=m2t-solid!40
                                [ {Kimi K2.5, InternVL-3.5}, fill=white, draw=m2t-solid!60, font=\Large\itshape ] 
                            ] 
                            [ {Deep Feature Alignment}, fill=m2t-leaf, draw=m2t-solid!40
                                [ {Qwen2.5-VL, Qwen3-VL, CogVLM}, fill=white, draw=m2t-solid!60, font=\Large\itshape ] 
                            ] 
                            [ {Advanced Positional Encoding}, fill=m2t-leaf, draw=m2t-solid!40
                                [ {Llama-4-Scout/Maverick, Gemma-4-E4B}, fill=white, draw=m2t-solid!60, font=\Large\itshape ] 
                            ] 
                            [ {Perception-Reasoning Decoupling}, fill=m2t-leaf, draw=m2t-solid!40
                                [ {MiMo-V2.5, GLM-5V-Turbo}, fill=white, draw=m2t-solid!60, font=\Large\itshape ] 
                            ] 
                        ] 
                        [ {Multi-scale Encoding}, fill=m2t-light, text=m2t-solid!90, draw=m2t-solid!30
                            [{Structure-Aware Tiling}, fill=m2t-leaf, draw=m2t-solid!40
                                [ {InternVL-3.5, MiniCPM-V-4.6}, fill=white, draw=m2t-solid!60, font=\Large\itshape ] 
                            ] 
                            [ {Dimension-Decoupled Positional Encoding}, fill=m2t-leaf, draw=m2t-solid!40 
                                [ {Qwen3-VL, GLM-5V-Turbo}, fill=white, draw=m2t-solid!60, font=\Large\itshape ] 
                            ] 
                            [ {Semantic-Driven Resampling}, fill=m2t-leaf, draw=m2t-solid!40
                                [ {InternVL-3.5}, fill=white, draw=m2t-solid!60, font=\Large\itshape ] 
                            ] 
                            [ {Resolution-Agnostic Projection}, fill=m2t-leaf, draw=m2t-solid!40
                                [ {Gemma-4-31B, Llama-4-Maverick}, fill=white, draw=m2t-solid!60, font=\Large\itshape ] 
                            ] 
                        ] 
                    ]
                    [ {Audio\\ Comprehension}, fill=m2t-mid, text=m2t-solid, draw=m2t-solid!60, font=\Large\bfseries
                        [ {Semantic-Acoustic Conflict}, fill=m2t-light, text=m2t-solid!90, draw=m2t-solid!30
                            [ {MiMo-V2.5, Gemma-4-E4B}, fill=white, draw=m2t-solid!60, font=\Large\itshape ] 
                        ] 
                        [ {High Latency \& Computation.}, fill=m2t-light, text=m2t-solid!90, draw=m2t-solid!30        
                            [ {Gemma-4-E4B, Nemotron-3-Nano-Omni}, fill=white, draw=m2t-solid!60, font=\Large\itshape ] 
                        ] 
                    ]
                    [ {Video\\ Comprehension}, fill=m2t-mid, text=m2t-solid, draw=m2t-solid!60, font=\Large\bfseries
                        [ {Computational Explosion}, fill=m2t-light, text=m2t-solid!90, draw=m2t-solid!30
                            [ {Compression \& Feature Aggregation}, fill=m2t-leaf, draw=m2t-solid!40 
                                [ {Kimi K2.5, GLM-5V-Turbo}, fill=white, draw=m2t-solid!60, font=\Large\itshape ] 
                            ] 
                            [ {Dynamic Token Allocation}, fill=m2t-leaf, draw=m2t-solid!40
                                [ {InternVL-3.5, Gemma-4-31B}, fill=white, draw=m2t-solid!60, font=\Large\itshape ] 
                            ] 
                        ] 
                        [ {Temporal \& Logical Inconsistency}, fill=m2t-light, text=m2t-solid!90, draw=m2t-solid!30
                            [ {Temporal Coordinate Encoding}, fill=m2t-leaf, draw=m2t-solid!40 
                                [ {Qwen3.6, Qwen3-VL}, fill=white, draw=m2t-solid!60, font=\Large\itshape ] 
                            ] 
                            [ {Explicit Time Tokens}, fill=m2t-leaf, draw=m2t-solid!40
                                [ {GLM-5V-Turbo}, fill=white, draw=m2t-solid!60, font=\Large\itshape ] 
                            ] 
                        ] 
                        [ {Long-range Dependency}, fill=m2t-light, text=m2t-solid!90, draw=m2t-solid!30
                            [{Modular Long-Term Memory}, fill=m2t-leaf, draw=m2t-solid!40 
                                [ {InternLM-XComposer2.5}, fill=white, draw=m2t-solid!60, font=\Large\itshape ] 
                            ] 
                            [{Distributed Clustering}, fill=m2t-leaf, draw=m2t-solid!40
                                [ {Kimi K2.5}, fill=white, draw=m2t-solid!60, font=\Large\itshape ] 
                            ] 
                        ] 
                    ]
                ]
                [ {M2G Scenario-\\based Generation \\ (\S3.2)}, fill=m2g-solid, text=white, draw=m2g-solid, font=\Large\bfseries
                    [ {Image \\Generation}, fill=m2g-mid, text=m2g-solid, draw=m2g-solid!60, font=\Large\bfseries
                        [ {High Visual Fidelity}, fill=m2g-light, text=m2g-solid!90, draw=m2g-solid!30
                            [ {Ming-Flash-Omni-2.0}, fill=white, draw=m2g-solid!60, font=\Large\itshape ] 
                        ] 
                        [ {Compositional Controllability}, fill=m2g-light, text=m2g-solid!90, draw=m2g-solid!30
                            [ {Seedream3.0, HiDream-O1-Image}, fill=white, draw=m2g-solid!60, font=\Large\itshape ] 
                        ] 
                    ]
                    [ {Audio\\Generation}, fill=m2g-mid, text=m2g-solid, draw=m2g-solid!60, font=\Large\bfseries
                        [ {Semantic-prosody Alignment}, fill=m2g-light, text=m2g-solid!90, draw=m2g-solid!30
                            [ {LTX-2, CosyVoice}, fill=white, draw=m2g-solid!60, font=\Large\itshape ] 
                        ]  
                        [ {Latency Control}, fill=m2g-light, text=m2g-solid!90, draw=m2g-solid!30 
                            [ {Qwen3-Omni, GLM-4-Voice, MiniCPM-o 4.5}, fill=white, draw=m2g-solid!60, font=\Large\itshape ] 
                        ]
                        [ {Reasoning-Streaming Synergy}, fill=m2g-light, text=m2g-solid!90, draw=m2g-solid!30
                            [ {Ming-Flash-Omni 2.0, Qwen3.5-Omni}, fill=white, draw=m2g-solid!60, font=\Large\itshape ] 
                        ]                                      
                    ]
                    [ {Video\\Generation}, fill=m2g-mid, text=m2g-solid, draw=m2g-solid!60, font=\Large\bfseries
                        [ {Physics Understanding}, fill=m2g-light, text=m2g-solid!90, draw=m2g-solid!30
                            [ {Training with Explicit Physics Rules}, fill=m2g-leaf, draw=m2g-solid!40 
                                [ {Wan2.2}, fill=white, draw=m2g-solid!60, font=\Large\itshape ] 
                            ]
                            [ {Implicit Emergence via Intelligent Reasoning}, fill=m2g-leaf, draw=m2g-solid!40 
                                [ {Kling-Omni, HunyuanVideo-1.5}, fill=white, draw=m2g-solid!60, font=\Large\itshape ] 
                            ] 
                        ]
                        [ {Token Explosion}, fill=m2g-light, text=m2g-solid!90, draw=m2g-solid!30
                            [ {Extreme Spatiotemporal VAE Compression}, fill=m2g-leaf, draw=m2g-solid!40
                                [ {LTX-2.3, Wan2.2}, fill=white, draw=m2g-solid!60, font=\Large\itshape ] 
                            ]
                            [ {Dynamic Sparse Attention Pruning}, fill=m2g-leaf, draw=m2g-solid!40 
                                [ {HunyuanVideo-1.5, Ming-Flash-Omni-2.0, Seedream 3.0}, fill=white, draw=m2g-solid!60, font=\Large\itshape ] 
                            ]
                        ]
                        [ {Audio-visual Alignment}, fill=m2g-light, text=m2g-solid!90, draw=m2g-solid!30
                            [ {Strict Audio-Visual Anchoring via Unified Timelines}, fill=m2g-leaf, draw=m2g-solid!40
                                [ {MiniCPM-o 4.5, Qwen3-Omni}, fill=white, draw=m2g-solid!60, font=\Large\itshape ] 
                            ]
                            [ {Synchronous Generation via Deep Architectural Coupling}, fill=m2g-leaf, draw=m2g-solid!40
                                [ {LTX-2.3, Seedance 2.0, OmniVoice}, fill=white, draw=m2g-solid!60, font=\Large\itshape ] 
                            ]
                        ]
                    ]
                ]
                [ {M2M Modeling \\ (\S3.3)}, fill=m2m-solid, text=white, draw=m2m-solid, font=\Large\bfseries
                    [ {Fully Discretized Unified}, fill=m2m-mid, text=m2m-solid, draw=m2m-solid!60, font=\Large\bfseries
                        [ {Loss from Discretizing}, fill=m2m-light, text=m2m-solid!90, draw=m2m-solid!60
                            [ {LongCat-Next, Moshi, AnyGPT}, fill=white, draw=m2m-solid!60, font=\Large\itshape ] 
                         ]
                        [ {Competition‑Driven Latency}, fill=m2m-light, text=m2m-solid!90, draw=m2m-solid!30 
                            [ {Chameleon, LLaDA2.0-Uni, Emu3.5}, fill=white, draw=m2m-solid!60, font=\Large\itshape ] 
                        ]
                     ]
                    [ {Modality-Specificity Preserving}, fill=m2m-mid, text=m2m-solid, draw=m2m-solid!60, font=\Large\bfseries
                        [ {Comprehension‑Generation Dilemma}, fill=m2m-light, text=m2m-solid!90, draw=m2m-solid!60
                            [ {Physical Decoupling}, fill=m2m-leaf, draw=m2m-solid!60
                                [ {Janus-Pro, BAGEL}, fill=white, draw=m2m-solid!60, font=\Large\itshape ] 
                            ]
                            [ {Encoder-Free Modeling}, fill=m2m-leaf, draw=m2m-solid!60
                                [ {TUNA-2, SenseNova-U1}, fill=white, draw=m2m-solid!60, font=\Large\itshape ] 
                            ]
                         ]                 
                        [ {Bridging AR and Diffusion}, fill=m2m-light, text=m2m-solid!90, draw=m2m-solid!60 
                                [ {Transfusion, Show-o2, OneCAT-3B, Mamoda2.5}, fill=white, draw=m2m-solid!60, font=\Large\itshape 
                                ] 
                            ]
                        ]
                    ]
                ]
            ]
        \end{forest}
    }
    \vspace{-2mm}
    \caption{A hierarchical taxonomy of the major technical challenges, core design axes, and representative NMM systems (as listed in Table~\ref{tab:multimodal_evolution}), which is derived from the discussion in Section 3.}
    \label{fig:taxonomy}
    \vspace{-4mm}
\end{figure*}
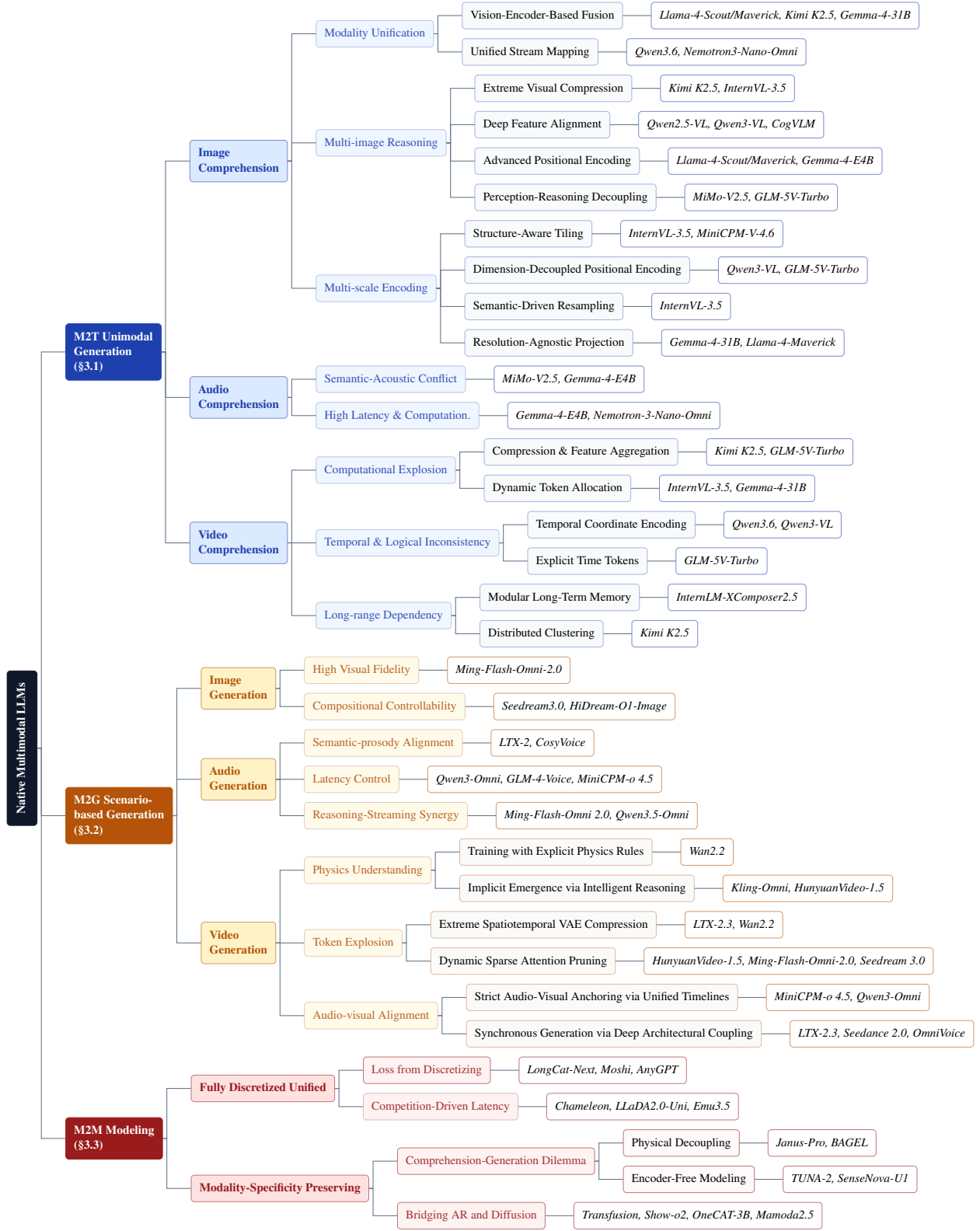

\subsubsection{Audio Comprehension}
NMM systems for audio understanding aim to process audio waveforms or acoustic features through the underlying representational space, achieving end-to-end cross-modal comprehension. In the course of this evolution, the core challenges are 1) \textit{Semantic-Acoustic Conflict} and 2) \textit{High Latency \& Computation}. 

\paragraph{Semantic-Acoustic Conflict.} Continuous audio signals are inherently incompatible with the highly structured, discrete textual semantics. MiMo-V2.5 employs MiMo-Audio-Tokenizer to generate semantic and acoustic features within a shared latent space. Its RVQ system prioritizes semantic structure in the initial layers, while the later layers refine acoustic details, thereby minimizing representation conflicts in the discrete token space. Gemma-4-E4B directly processes log-Mel spectrograms through a Conformer-based audio encoder, which outputs continuous embedding vectors that preserve complete acoustic information. 
Further bridging discrete and continuous paradigms, Nemotron-3-Nano-Omni adopts a non-linear alignment strategy: it extracts deep acoustic features via a FastConformer encoder and projects them into the language backbone through a 2-layer MLP, preserving fine-grained continuous details while enabling robust semantic grounding in the shared latent space.

\paragraph{High Latency \& Computation.} High latency and computational costs present another major challenge in audio comprehension. Gemma-4-E4B adopts a long frame duration in its acoustic encoder, compressing each second of audio input to vectors, which are then directly injected into the backbone through a projection layer. This approach significantly reduces the cost of forward propagation and enables real-time speech interaction with extremely low latency. To address this computational bottleneck at scale, Nemotron-3-Nano-Omni implements an algorithmic-architectural co-optimization framework spanning the entire processing pipeline. On the encoder side, it processes log-mel spectrogram features followed by three convolutional subsampling layers, yielding an 8$\times$ temporal downsampling rate. Furthermore, its underlying TDT decoder dynamically skips frames based on predicted token durations during inference, effectively filtering out silent or redundant acoustic periods before projection. On the backbone side, Nemotron-3-Nano-Omni is built on a 31B Mamba2-Transformer hybrid MoE that only activates 3B parameters per forward pass. The linear complexity $O(N)$ of the Mamba2 layers replaces the quadratic attention complexity $O(N^2)$ for long-context sequences, allowing the model to scale efficiently while delivering higher system throughput at equivalent interactivity thresholds.
 
\subsubsection{Video Comprehension}
Introducing the video modality from static images expands the input space from $H \times W$ to $T \times H \times W$, the increase in dimension triggers a series of non‑linearly scaling difficulties. Based on the analysis of current mainstream NMM systems, the core bottlenecks in video input support can be summarized into three points: 1) \textit{Computational Explosion} 2) \textit{Temporal and Logical Inconsistency} 3) \textit{Long-range Dependency}.

\paragraph{Computational Explosion.} For videos, the number of tokens generated per second is significantly redundant, which not only approaches memory capacity limits but also results in computational costs that scale quadratically with sequence length in transformer-based models.
One approach is \textbf{\((i)\) Compression \& Feature Aggregation}, which leverages the high similarity between video frames to reduce redundancy before feeding the representation into an LLM.
Kimi K2.5 packs consecutive frames into a spatiotemporal volume and performs temporal averaging at the patch level, enabling processing of videos longer under the same computational budget.
GLM-5V-Turbo uses 3D convolutions instead of 2D in the encoder to perform downsampling along the temporal axis during feature extraction, significantly improving efficiency for long video processing. 
\textbf{\((ii)\) Dynamic Token Allocation} based on an image's resolution and semantic density can also address this issue. For instance, InternVL-3.5 introduces a Visual Resolution Router to assign 256 tokens to semantically rich patches while compressing backgrounds to 64 tokens, cutting overall token redundancy by 50\%, whereas Gemma-4-31B allows users to manually set a token per task, using a high budget for complex work like OCR and a low budget for simple recognition.

\paragraph{Temporal \& Logical Inconsistency.} Unlike static images, video understanding requires the incorporation of a temporal dimension. Lacking temporal awareness is susceptible to temporal hallucinations and may fail to maintain consistent object identities across frames.
One typical strategy is \textbf{\((i)\) Temporal Coordinate Encoding}.
To enhance the model’s physical perception, Qwen3.6/Qwen3‑VL decomposes the RoPE position encoding into three interleaved dimensions, giving each token a unique representation in 3D spatiotemporal coordinates, thereby enabling accurate localization of events at timeline.
Another approach is \textbf{\((ii)\) Explicit Time Tokens}. GLM‑5V-Turbo inserts explicit time tokens into the video‑frame sequence, allowing the model to perceive physical time as it reads the sequence like understanding natural language. This is critical to long‑video summarization tasks that require precise time localization (e.g., soccer matches).

\paragraph{Long-range Dependency.} Processing video streams that last for hours requires the model to maintain efficient working memory. Raw video features can quickly exhaust the context window, making the effective management, rather than simple forgotten a key challenge to achieve native long‑video understanding. One solution is \textbf{\((i)\) Modular Long-Term Memory}. InternLM‑XComposer2.5~\cite{internlmxcomposer2} builds an independent memory pool to compress and store perceptual video features into a long‑term memory bank, retrieving information on‑demand during Q\&A. This supports unlimited‑length streaming interaction. Another solution involves \textbf{\((ii)\) Distributed Clustering}. Kimi K2.5 introduces an agent‑swarm mode. A central dispatcher decomposes long‑video tasks and assigns them to hundreds of specialized sub‑agents for parallel analysis. This distributed parsing improves processing efficiency compared to monolithic models.

\subsection{M2G Scenario-based Generation}
\label{sec:multi-to-target}
\subsubsection{Image Generation}
For image generation, traditional workflows rely on LLMs to generate prompts and feed into standalone diffusion models~\cite{wu2026visualgenerationnewera}. However, this approach struggles with maintaining spatial consistency. Native image generation technology has moved beyond this piecemeal paradigm, establishing the joint modeling of text and images as the mainstream approach for this stage. For image generation, we identified two primary challenges: 1) \textit{High Visual Fidelity} and 2) \textit{Compositional Controllability}.

\paragraph{High Visual Fidelity.} Frameworks such as Ming-Flash-Omni-2.0 combine Transformer and Diffusion models in a shared latent space. Taking Mask-based Discrete Diffusion as a unified mask-aware architecture, it learns the joint distribution of cross-modal tokens. The hidden layers predict the next text token and output continuous features which guide the image denoising. Through unified self-attention, the structure of text and the spatial layout of images are aligned at an early stage of feature fusion, leading to superior pixel-level fidelity. This approach also reduces artifacts such as spelling errors when generating text within images, establishing a robust foundation for high-quality generation.

\paragraph{Compositional Controllability.} The second major hurdle in native image generation is strictly adhering to complex compositional instructions, particularly when prompts involve multiple interacting entities or precise positional constraints. While early unified models often suffer from attribute leakage or spatial misalignment, recent architectures have introduced dedicated spatial grounding mechanisms. For instance, Seedream3.0 implements spatial perception through cross-modality RoPE, which helps the model better align the spatial positioning logic in text instructions with visual tokens. Taking explicit control a step further, HiDream-O1-Image integrates coordinate-aware representations, allowing the model to project discrete layout instructions directly into localized generation processes. 

\subsubsection{Audio Generation} 
For audio generation, the primary challenges lie in \textit{\((i)\) Semantic-Prosody Alignment}, \textit{\((ii)\) Latency Control} and \textit{(iii) Reasoning-Streaming Synergy}. While many models natively support robust audio comprehension, minimizing generation latency remains critical, particularly in full-duplex conversational scenarios.

\paragraph{Semantic-Prosody Alignment.} Similar to image generation tasks, audio generation can also be approached via two routes: one maps sound into continuous latent vectors, while the other discretizes audio signals and generates via discrete token prediction. The former approach often leads to higher acoustic fidelity and semantic-prosody alignment.
Specifically, LTX-2 utilizes RoPE to process audio, leveraging bidirectional cross-attention layers to capture transient dependencies that correspond to visual impacts triggering acoustic features. CosyVoice~\cite{du2024cosyvoice} focuses on semantic-acoustic decoupling and employs a supervised semantic tokenizer to handle content control, while a flow-matching module is used to render timbre and emotion.

\paragraph{Latency Control.} Similar to discrete tokenization in image generation, the core value of discrete audio generation lies in converting continuous sound signals into symbols that are essentially identical to text. This allows LLMs to directly leverage their powerful autoregressive prediction and instruction-following capabilities.
Qwen3-Omni adopts a Multi-Token Prediction (MTP)~\cite{gloeckle2024better} strategy to balance efficiency and quality, it takes the MTP module to output residual codebooks simultaneously, paired with a Code2Wav renderer for frame-level streaming synthesis—achieving a first-packet latency. GLM-4-Voice~\cite{zeng2024glm} utilizes a Single-codebook approach and introduces an ASR encoder (e.g., Whisper-v3~\cite{Radford2022RobustSR}) into the VQ bottleneck. MiniCPM-o 4.5 prioritizes high token density optimization. By compressing audio into an extremely small number of tokens per second, it is specifically tailored to accommodate the computational bandwidth limitations of mobile devices.

\paragraph{Reasoning-Streaming Synergy.}
Beyond the trade-off between quality and speed, a paradigm focuses on the synergy between internal reasoning and external streaming. The Thinker-Talker architecture has emerged as a leading solution for sophisticated voice chat. It allows a high-capacity Thinker to perform long-form reasoning in the background while a lightweight Talker (e.g., OmniVoice) delivers speech with ultra-low latency. Ming-Flash-Omni 2.0 extends this by integrating ambient sounds and background music into a single autoregressive DiT~\cite{peebles2023scalable} head, enabling precise control over environmental atmosphere via natural language. To address the inherent speed mismatch between text reasoning and audio synthesis, Qwen3.5-Omni introduces Adaptive Rate Interleave Alignment, preventing accuracy drifts during streaming. Furthermore, Mini-Omni-Reasoner~\cite{xie2025mini} achieves a thinking-in-speaking mechanism by maintaining hidden reasoning tokens while concurrently outputting audio tokens, effectively bridging the gap between slow thinking intelligence and fast talking responsiveness.

\subsubsection{Video Generation}
Compared to static images or one-dimensional audio, video generation demands exponentially greater resources in terms of both output quality and computational complexity. As a result, the task faces extreme engineering and mathematical hurdles regarding compute power, memory consumption, and spatio-temporal consistency. According to our research, the primary challenges in current video generation are concentrated in 1) \textit{Physics Understanding}, 2) \textit{Token Explosion}, and 3) \textit{Audio-visual Alignment}.

\paragraph{Physics Understanding.} Generative video models often struggle with frequent violations of basic physical laws. While diffusion models could produce highly realistic images through pixel-level noise reduction, they typically lacked an abstract understanding of concepts like rigid body dynamics, gravity and collision physics. This resulted in phenomena such as objects floating unnaturally, moving without external forces, or appearing to melt or pass through one another instead of colliding solidly. To address this, \textbf{\((i)\) Training with Explicit Physics Rules} is one of the most efficient methods to instill physical understanding and rules with explicit constraints on objects. 
Frameworks like NewtonRewards~\cite{le2025gravityvideogenerationposttraining} employ frozen visual networks to extract measurable physical metrics, translating Newtonian motion laws and mass conservation directly into mathematical penalty terms for reinforcement learning. For handling complex rigid-body collisions, systems like PhysRVG~\cite{zhang2026physrvgphysicsawareunifiedreinforcement} use foundational segmentation networks (e.g., SAM2~\cite{ravi2025sam}) to derive motion masks frame by frame, accurately tracking object trajectories in generated videos. These trajectories are then compared with real-world physical paths to compute errors. 
With optical flow-based Newtonian penalties, Wan2.2 significantly improved temporal consistency in scenes showing free fall, projectile motion, and inclined plane sliding, proving models can internalize Newtonian structures. 
\textbf{\((ii)\) Implicit Emergence via Intelligent Reasoning} is another method which models adopt an end-to-end understand-reason-generate architecture and trained on vast amounts of data annotated with precise physical labels. Kling-Omni, for instance, bridges the gap between visual-language input and physical simulation through an intelligent prompt enhancer that interprets physical intent, paired with a DiT-based Omni-Generator refined with large-scale fine-tuning and DPO~\cite{rafailov2024directpreferenceoptimizationlanguage}. This implicitly builds a physics engine inside the model, ensuring rigid-body stability and identity consistency in multi-agent interactions. Similarly, HunyuanVideo-1.5 was trained on massive real-world videos with highly accurate multimodal captions. Without explicit RL physics rewards, the model naturally developed strong temporal coherence and long-term physical reasoning simply by learning from the data distribution.

\paragraph{Token Explosion.} After video generation models transitioned from U-Net~\cite{ronneberger2015u} to DiT, the massive number of tokens from high-resolution and long videos caused self-attention computation to grow quadratically, leading to OOM errors and slower generation. To achieve low memory usage during generation, a common approach is \textbf{\((i)\) Extreme Spatiotemporal VAE Compression}, which uses a customized VAE to compress video pixels into a compact latent space before diffusion computation begins.
For example, LTX-2.3 moves the patchify operation to the VAE input, enabling single-step denoising to generate native 4K resolution with minimal memory overhead. Wan2.2 employs a highly optimized Wan-VAE,  which helps reducing the number of spatial tokens. Combined with the Flow Matching paradigm, this approach significantly lowers memory pressure when generating arbitrarily long 1080P videos.
Another technical route is \textbf{\((ii)\) Dynamic Sparse Attention Pruning}, which dynamically identifies and removes redundant information that contributes little to generation, transforming global dense computation into local sparse attention. HunyuanVideo-1.5 introduces the SSTA mechanism, which automatically prunes redundant spatiotemporal blocks such as static backgrounds during generation, boosting end-to-end inference speed and enabling smooth operation on consumer-grade GPU memory. 
Ming-Flash-Omni 2.0 employs a MoE architecture with modality-level routing. This design enables the model to handle complex audiovisual generation tasks with very low latency while retaining its vast knowledge capacity. 

\paragraph{Audio-visual Alignment.} In the final stage of video generation, achieving millisecond-level synchronization between audio and visuals in both timing and physics is a key challenge. Currently, the industry is advancing mainly in two directions. The first approach is \textbf{\((i)\) Strict Audio-Visual Anchoring via Unified Timelines}. This method focuses on building a unified audiovisual coordinate system at the underlying level, ensuring all modalities are locked in the time dimension. MiniCPM-o 4.5 introduces the Omni-Flow full-duplex framework, which forces audio-visual inputs and text/speech outputs to align at the token level on a single timeline. This not only achieves millimeter-level sync but also allows the model to proactively speak based on visual changes. Qwen3-Omni adopts TM-RoPE anchored to absolute time, abandoning relative segment alignment and taking explicit time IDs to lock all audio-visual features, eliminating temporal drift in long sequences.
The second approach is \textbf{\((ii)\) Synchronous Generation via Deep Architectural Coupling}, which emphasizes building an audio-visual handshake within the model. Through cross-modal attention bridges or non-autoregressive acoustic mapping, sound is generated in real time alongside visuals rather than as a post-processing step. LTX-2.3 employs a highly asymmetric dual-stream architecture with bidirectional cross-modal attention layers for dense interaction. Combined with cross-modal AdaLN~\cite{peebles2023scalable} and modality-CFG~\cite{LightricksLTX2_2026}, it ensures sound effects correspond precisely to visual actions. Seedance2.0 constructs a dedicated Attention Bridge at every millisecond of the diffusion process, action intensity from the visual branch is passed to the audio branch, while audio emotion and rhythm influence visual lighting. For real-time speech generation, OmniVoice and Qwen3-Omni take non-autoregressive discrete codec-based acoustic mapping, skipping complex two-stage pipelines. OmniVoice directly maps text to multi-codec acoustic tokens, while Qwen3-Omni replaces diffusion with a lightweight causal ConvNet~\cite{liu2022convnet}, achieving ultra-low TTFT and instant sync in interactive scenarios.

\subsection{M2M Symmetric Modeling}
\label{sec:multi-to-multi} 
The third category of models we summarize is Symmetrical Multi-Modal, which are capable of understanding multiple modalities and also generating them symmetrically within the same framework. At the architectural level, such models can be divided into two main technical camps. 
The first is \textit{Fully Discretized Unified}, which aims to compress and map continuous signals from all modalities into discrete tokens, and then train them under a unified autoregressive generation objective. 
The second is \textit{Modality-Specificity Preserving}, which argues that different modalities, such as the spatial continuity of images or the temporal dynamics of audio, possess inherent structures that cannot be losslessly expressed through a discrete vocabulary. 
While still adhering to a unified Transformer backbone, these models preserve continuous feature spaces, decoupled visual encoders, or hybrid loss functions. 
\subsubsection{Fully Discretized Unified}
The fully discretized architecture offers the ultimate advantage of extreme simplicity, yet it also brings two challenges. 1) \textit{Loss from Discretizing}, 2) \textit{Competition‑Driven Latency}.

\paragraph{Loss from Discretizing.} When continuous signals from the physical world are transformed into a discrete vocabulary, it inherently entails lossy compression. When compressing high-resolution images or audio into a limited set of discrete IDs, traditional codecs permanently discard the low-level features essential for fine-grained intensive tasks, severely limiting the performance ceiling of fully discretized models in perceptual tasks.
To mitigate this information loss, models strive to develop tokenizers that minimize semantic and acoustic loss.
LongCat‑Next proposes the Semantic Completeness principle, designing dNaViT~\cite{MeituanLongCat2026} as its visual tokenizer. Its codebook embeddings are not fixed but randomly initialized and co‑evolve with language tokens under a shared autoregressive objective.
Moshi tackles the discrete bottleneck in speech with its in‑house neural audio codec, Mimi, which uses RVQ to decompose continuous audio. Through a knowledge‑distillation mechanism, its early acoustic tokens are forced to match the semantic representations of self‑supervised speech models.  
AnyGPT adopts a multilingual strategy, deploying highly specialized discrete tokenizers for each continuous modality.

\paragraph{Competition‑Driven Latency.} When a model forces high‑information‑density discrete text tokens and extremely sparse visual/audio tokens into the single discrete vocabulary and computes cross‑entropy in the same Softmax layer, features from modalities with different entropy levels compete for weight. 
On large‑scale data, this competition can cause output norms to explode exponentially, leading to gradient divergence. Moreover, relying entirely on autoregressive step‑by‑step prediction of thousands of image tokens results in intolerable inference latency. 
Chameleon modifies the standard attention mechanism by introducing QK‑Norm to suppress representational competition, applying layer normalization to Query and Key vectors before computing dot products. 
LLaDA2.0‑Uni equips its inference engine with Sprint Inference, which breaks the latency bottleneck of single‑step decoding via Adaptive Unmasking and confidence‑based Batch Acceptance.
To solve the minutes‑long serial inference pain point for single‑image generation, Emu3.5 proposes Discrete Diffusion Adaptation. This shifts the model’s inference behavior from strictly token‑by‑token serial decoding to bidirectional parallel prediction, delivering roughly 20× acceleration in single‑image inference without sacrificing performance.

\subsubsection{Modality-Specificity Preserving}
Unlike unified architectures based on discrete tokens, an alternative approach argues visual spatial continuity cannot be captured losslessly by a discrete vocabulary. This school of thought favors continuous feature spaces, decoupled encoders, and hybrid loss functions (e.g., AR for text and Diffusion for images). However, preserving modality-specific traits creates two fundamental conflict: 
1) \textit{Comprehension‑Generation Dilemma}, 2) \textit{Bridging AR and Diffusion}.

\paragraph{Comprehension‑Generation Dilemma.} Understanding requires highly compressed, high-level semantic abstraction, whereas Generation demands fine-grained, low-level pixel features for reconstruction. 
When a shared representation tries to serve both, the network suffers from Task Interference, caught in a conflict between compressing semantics and preserving detail.
To resolve this, researchers are pursuing two main strategies:
\textbf{\((i)\) Physical Decoupling}. Janus-Pro uses separate visual encoders for understanding and generation, allowing each to evolve independently. BAGEL extends this into the backbone via a Mixture-of-Transformer-Experts (MoT) architecture, using hard routing to direct tokens to specialized Understanding or Generation experts. This enables advanced world-modeling such as 3D navigation.
\textbf{\((ii)\) Encoder-Free Modeling}. TUNA-2 and SenseNova-U1 take a more radical path by removing traditional CLIP~\cite{radford2021learning} encoders and VAEs. By feeding raw image patches directly into the network, they eliminate pre-trained inductive biases. This allows for native pixel-level coordination; SenseNova-U1, for instance, can reconstruct precise microscopic textures using raw pixel streams even when its understanding branch is frozen.

\paragraph{Bridging AR and Diffusion.} Furthermore, to retain modality specificity, the model must operate across discrete and continuous representations, integrating both AR and Diffusion paradigms. Seamlessly fusing these disparate spaces within a single network and bridging logical planning with high-speed rendering during inference.
Transfusion utilizes a unified Transformer that applies discrete NTP loss for text and a continuous denoising Diffusion loss for image patches. To bridge these paradigms, it employs a hybrid attention mechanism: causal masking for text to maintain logic, and bidirectional attention for image patches to capture spatial continuity. Show-o2 introduces Spatial-Temporal Fusion by 3D Causal VAE. It extracts high-level information via independent semantic layers and fuses them with low-level features through cascading and MLPs. Separate AR and Flow-Matching heads at the top manage heterogeneous text and video flows with minimal parameter overhead. OneCAT-3B implements Modality-MoE within a pure decoder architecture. It introduces a multi-scale visual AR mechanism which bypasses serial bottlenecks and boosts generation speed. Mamoda2.5 bridges AR and Diffusion by MetaQueries. Instead of relying on slow, error-prone visual token prediction, the AR backbone generates highly condensed logical plans. These continuous features are then bridged directly to a backend DiT-MoE module for high-speed, fine-grained pixel rendering.

%% file: data.tex
\section{Dataset}\label{data}
Data plays a central role in shaping the capabilities of NMM systems. Unlike earlier vision-language systems that mainly relied on image-text pairs for cross-modal alignment, recent native multimodal models are trained on heterogeneous data mixtures covering text, images, videos, audio, documents, GUI states, tool-use traces, and preference signals. These data sources differ not only in modality coverage, but also in their input-output structure and supervision granularity. Some data are designed for multimodal understanding, such as image captioning, visual question answering, OCR, document parsing, chart reasoning, grounding, and multi-image reasoning. Others target multimodal generation and editing, including text-to-image, image-to-image, text-to-video, speech generation, and interleaved image-text generation. More recently, interaction-oriented and preference-oriented data have become increasingly important, enabling models to operate in visual environments, follow complex instructions, and align their responses with human preferences. Overall, this section organizes the training data of NMM systems according to their functional roles and supervision formats.

\begin{table*}[!t]
\centering
\scriptsize
\renewcommand{\arraystretch}{1.15}
\setlength{\tabcolsep}{3pt}
\definecolor{UndBadge}{HTML}{2E86C1}
\definecolor{GenBadge}{HTML}{16A34A}
\definecolor{IntBadge}{HTML}{CA8A04}
\definecolor{PrefBadge}{HTML}{8B5CF6}
\definecolor{UndRow}{HTML}{F4F9FD}
\definecolor{GenRow}{HTML}{EAF7EE}
\definecolor{IntRow}{HTML}{FBF5E6}
\definecolor{PrefRow}{HTML}{F5F3FF}

\setlength{\fboxsep}{3pt}
\newcommand{\modbadge}[2]{%
  {\setlength{\fboxsep}{2.5pt}\colorbox{#1}{\textbf{\color{white}\scriptsize #2}}}%
}
\newcommand{\taskbadge}[2]{%
  {\setlength{\fboxsep}{2pt}\fcolorbox{#1}{#1!12}{\textbf{\color{#1}\scriptsize #2}}}%
}

\begin{tabularx}{\textwidth}{@{} l l p{3.8cm} p{1.8cm} Y @{}}
\toprule
\textbf{Category} & \textbf{Sub-type} & \textbf{Representative Datasets} & \textbf{Modalities} & \textbf{Key Supervision / Description} \\
\midrule

\rowcolor{UndRow}
\modbadge{UndBadge}{Understand} & \taskbadge{UndBadge}{Image-Text Alignment} & LAION-5B~\cite{schuhmann2022laion}, COCO Captions~\cite{chen2015microsoft}, CC3M/CC12M~\cite{sharma2018conceptual}, YFCC100M~\cite{thomee2016yfcc100m}, DataComp~\cite{gadre2023datacomp} & T, I & Weakly-aligned web-scale pairs for cross-modal mapping; provides image-level semantics. \\
\addlinespace[2pt]
\rowcolor{UndRow}
& \taskbadge{UndBadge}{VQA \& Instruction Tuning} & VQA v2~\cite{antol2015vqa}, GQA~\cite{hudson2019gqa}, OK-VQA~\cite{marino2019ok}, ScienceQA~\cite{lu2022learn}, LLaVA-Instruct~\cite{Liu2023VisualIT}, InstructBLIP~\cite{dai2023instructblip} & T, I & Task-driven QA and multi-turn instruction-following; converts perception into dialogue. \\
\addlinespace[2pt]
\rowcolor{UndRow}
& \taskbadge{UndBadge}{Interleaved \& Multi-Image} & MMC4~\cite{zhu2023multimodal}, OBELICS~\cite{laurenccon2023obelics}, OmniCorpus~\cite{li2025omnicorpus}, MANTIS~\cite{penha2019introducing}, NLVR2~\cite{suhr2019corpus}, MuirBench~\cite{wang2025muirbench}, BLINK~\cite{fu2024blink} & T, I & Natural interleaved documents and multi-image comparison/reasoning; long-range dependency. \\
\addlinespace[2pt]
\rowcolor{UndRow}
& \taskbadge{UndBadge}{Document, Chart \& Grounding} & DocVQA~\cite{mathew2021docvqa}, InfographicVQA~\cite{mathew2022infographicvqa}, ChartQA~\cite{masry2022chartqa}, TextVQA~\cite{singh2019towards}, Flickr30k Entities~\cite{plummer2015flickr30k}, RefCOCO~\cite{kazemzadeh2014referitgame} & T, I & Parsing text, layouts, tables, charts; region-level bounding box and referring expressions. \\
\addlinespace[2pt]
\rowcolor{UndRow}
& \taskbadge{UndBadge}{Video \& Audio Understanding} & MSR-VTT~\cite{chen2022msr}, ActivityNet~\cite{krishna2017dense}, WebVid~\cite{Bain21}, AudioSet~\cite{jort_audioset_2017}, LibriSpeech~\cite{panayotov2015librispeech}, Common Voice~\cite{ardila2020common}, Clotho~\cite{drossos2020clotho} & T, I, V, A & Temporal action/event recognition, speech transcription, acoustic scene understanding. \\
\midrule

\rowcolor{GenRow}
\modbadge{GenBadge}{Generate} & \taskbadge{GenBadge}{Text-to-Image \& Editing} & LAION-5B~\cite{schuhmann2022laion}, DiffusionDB~\cite{wang2023diffusiondb}, InstructPix2Pix~\cite{brooks2022instructpix2pix}, MagicBrush~\cite{Zhang2023MagicBrush}, HQ-Edit~\cite{hui2024hq}, UltraEdit~\cite{zhao2024ultraeditinstructionbasedfinegrainedimage} & T, I & Synthesizing or editing images from prompts; controlled transformation with masks/instructions. \\
\addlinespace[2pt]
\rowcolor{GenRow}
& \taskbadge{GenBadge}{Controllable Generation} & ControlNet~\cite{zhang2023adding}, GLIGEN~\cite{li2023gligen}, T2I-Adapter~\cite{mou2024t2i}, Composer & T, I + cond. & Grounded generation with depth, sketch, layout, bounding boxes, pose. \\
\addlinespace[2pt]
\rowcolor{GenRow}
& \taskbadge{GenBadge}{Interleaved Image-Text Gen.} & VIST~\cite{kim2021conditional}, OpenLEAF~\cite{an2023openleaf}, CoMM~\cite{chen2025comm}, InterSyn~\cite{ma2025intersyn} & T, I & Generating coherent multimodal sequences (stories, tutorials) with entity consistency. \\
\addlinespace[2pt]
\rowcolor{GenRow}
& \taskbadge{GenBadge}{Video Generation} & WebVid-10M~\cite{Bain21}, Panda-70M~\cite{chen2024panda70m}, OpenVid-1M~\cite{nan2024openvid}, VidGen-1M~\cite{tan2024vidgen} & T, I, V & Text/video-to-video; requires temporal coherence, motion quality, caption recaptioning. \\
\addlinespace[2pt]
\rowcolor{GenRow}
& \taskbadge{GenBadge}{Audio \& Speech Generation} & LibriTTS~\cite{zen2019libritts}, VCTK~\cite{yamagishi2019cstr}, GigaSpeech~\cite{chen2021gigaspeech}, Emilia~\cite{he2024emilia}, AudioCaps~\cite{kim-NAACL-HLT-2019}, WavCaps~\cite{mei2023wavcaps}, MusicCaps~\cite{agostinelli2023musiclm} & T, A (Sp) & Text-to-speech, voice cloning, music and environmental sound generation. \\
\midrule

\rowcolor{IntRow}
\modbadge{IntBadge}{Interact} & \taskbadge{IntBadge}{Web Interaction} & WebShop~\cite{yao2022webshop}, Mind2Web~\cite{deng2023mind2web}, WebArena~\cite{zhou2024webarena}, VisualWebArena~\cite{koh2024visualwebarena}, WebLINX~\cite{lù2024weblinx}, WebVoyager~\cite{he2024webvoyager} & T, I (GUI) & Goal-driven web navigation: searching, clicking, form filling on real/simulated websites. \\
\addlinespace[2pt]
\rowcolor{IntRow}
& \taskbadge{IntBadge}{Mobile \& Desktop GUI} & AITW~\cite{rawles2023androidinthewild}, RICO~\cite{deka2017rico}, ScreenAI~\cite{baechler2024screenai}, SeeClick~\cite{cheng2024seeclick}, OSWorld~\cite{OSWorld}, Windows Agent Arena~\cite{bonatti2024windows} & T, I (GUI) & Screenshot/UI-tree to action (tap, type, drag); covers mobile and OS environments. \\
\addlinespace[2pt]
\rowcolor{IntRow}
& \taskbadge{IntBadge}{Embodied Interaction} & ALFWorld~\cite{ALFWorld20}, BridgeData V2~\cite{walke2023bridgedata}, Open X-Embodiment~\cite{open_x_embodiment_rt_x_2023}, Magma~\cite{yang2025magma} & T, I, V (robot) & Language-conditioned manipulation from visual observations and robot states. \\
\midrule

\rowcolor{PrefRow}
\modbadge{PrefBadge}{Align} & \taskbadge{PrefBadge}{Hallucination \& Faithfulness} & LLaVA-RLHF~\cite{2023llavarlhf}, RLHF-V~\cite{yu2024rlhfvtrustworthymllmsbehavior}, VLFeedback~\cite{li2024vlfeedback}, RLAIF-V~\cite{yu2024rlaifv}, HA-DPO~\cite{zhang2025mm}, V-DPO~\cite{xie2024v} & T, I & Comparative or span-level feedback to reduce visual hallucinations; AI-assisted labels. \\
\addlinespace[2pt]
\rowcolor{PrefRow}
& \taskbadge{PrefBadge}{Safety Alignment} & SPA-VL~\cite{zhang2024spavl}, Safe RLHF-V~\cite{ji2026safe} & T, I & Safe/unsafe response pairs under multimodal harmful prompts. \\
\addlinespace[2pt]
\rowcolor{PrefRow}
& \taskbadge{PrefBadge}{Generation Quality Preference} & ImageReward~\cite{xu2023imagereward}, Pick-a-Pic~\cite{kirstain2023pick}, HPS v2~\cite{wu2023human}, VBench~\cite{huang2023vbench}, VBench++~\cite{huang2024vbench++} & I, V & Human preference scores for aesthetics, alignment, temporal consistency, motion quality. \\
\addlinespace[2pt]
\rowcolor{PrefRow}
& \taskbadge{PrefBadge}{Agentic Preference} & Environment Rewards, Human Demos & T, I, A & Env-aligned Action correctness, efficiency, recovery. \\
\bottomrule
\end{tabularx}
\caption{Training data for NMM, categorized as discussed in \S\ref{data}. The four main categories align with the section structure: Understanding-Oriented (\modbadge{UndBadge}{Understand}), Generation-Oriented (\modbadge{GenBadge}{Generate}), Interaction-Oriented (\modbadge{IntBadge}{Interact}), and Preference \& Alignment (\modbadge{PrefBadge}{Align}). Modalities: T = Text, I = Image, V = Video, A = Audio/Speech.}
\label{tab:native_mllm_data_engineering}
\end{table*}

\subsection{Understanding-Oriented Data}
Understanding-oriented data aims to train NMM systems to interpret multimodal inputs and produce textual or structured semantic outputs. In contrast to generation-oriented data, where the target may be an image, video, or speech signal, understanding-oriented data usually follows an input-to-text or input-to-structure paradigm, such as image captioning, visual question answering, OCR, document parsing, chart reasoning, grounding, and video/audio understanding. Its role is to establish the perceptual and reasoning foundation of native multimodal models, enabling them to recognize visual content, read text, localize evidence, compare multiple inputs, and reason over temporal or acoustic signals.

The most fundamental form of understanding-oriented data is image-text alignment pairs. Large-scale image-text pairs provide weak but scalable supervision for mapping visual semantics into language space. Early frameworks such as CLIP and ALIGN proposed that noisy web-scale image-text pairs offer strong transferable visual representations. Open datasets such as YFCC100M~\cite{thomee2016yfcc100m}, Conceptual Captions~\cite{sharma2018conceptual}, COCO Captions~\cite{chen2015microsoft}, and LAION-5B~\cite{schuhmann2022laion} further shaped this paradigm, while DataComp~\cite{gadre2023datacomp} studied how data filtering and mixture design affect contrastive vision-language training. Although these pairs are effective for learning objects, scenes, attributes, and general semantic descriptions, they are insufficient for learning fine-grained spatial localization, multi-step reasoning, document understanding, and long-context multimodal comprehension.

To go beyond generic captioning, visual question answering and visual instruction data introduce more task-oriented supervision. Datasets such as VQA~\cite{antol2015vqa}, VQA v2, GQA~\cite{hudson2019gqa}, OK-VQA~\cite{marino2019ok}, A-OKVQA, ScienceQA~\cite{lu2022learn}, and VizWiz~\cite{gurari2018vizwiz} require models to answer questions based on visual evidence, external knowledge, compositional relations, or real-world visual scenarios. Compared with captioning data, these datasets force models to selectively attend to relevant parts of the input rather than describe the whole image. More recent instruction-tuning datasets, represented by LLaVA and InstructBLIP~\cite{dai2023instructblip}, convert visual understanding tasks into natural language instruction-following formats, often using strong language models or multimodal models to synthesize questions, answers, rationales, and conversations. This shift is important for native multimodal models because it aligns perception with open-ended dialogue and instruction following, which are central to modern multimodal assistants.

Another important direction is interleaved and multi-image understanding data. Unlike isolated image-text pairs, interleaved data preserves the natural ordering of images and text in web pages, tutorials, documents, and multimodal articles. Flamingo~\cite{alayrac2022flamingo} showed the importance of web-scale interleaved image-text data for multimodal in-context learning, while datasets such as MMC4~\cite{zhu2023multimodal}, OBELICS~\cite{laurenccon2023obelics}, and OmniCorpus~\cite{li2025omnicorpus} provide large-scale open resources for training models on multimodal sequences. This data format changes the supervision unit from a single image-text pair to a multimodal context, allowing models to learn cross-image reference, long-range dependency, and contextual reasoning. Multi-image understanding data further extends this idea by requiring models to compare, aggregate, or reason over multiple visual inputs. Datasets and benchmarks such as MANTIS~\cite{penha2019introducing}, NLVR2~\cite{suhr2019corpus}, MuirBench~\cite{wang2025muirbench}, and BLINK~\cite{fu2024blink} evaluate capabilities such as image comparison, co-reference, temporal ordering, visual difference recognition, and multi-view reasoning. These data are especially relevant for native multimodal models because real-world tasks often involve sets or sequences of images rather than a single static input.

Structured visual understanding data further enriches the supervision signal by requiring models to parse text, layout, tables, charts, and other symbolic structures embedded in images. OCR-related datasets such as TextVQA~\cite{singh2019towards} train models to read scene text and combine it with visual context. Document-oriented datasets such as DocVQA~\cite{mathew2021docvqa} and InfographicVQA~\cite{mathew2022infographicvqa} require models to understand layout, reading order, forms, figures, and document-level semantics. Chart and table understanding datasets such as ChartQA~\cite{masry2022chartqa}, FigureQA~\cite{kahou2017figureqa}, PlotQA~\cite{methani2020plotqa}, and DVQA~\cite{kafle2018dvqa} introduce numerical, logical, and arithmetic reasoning over visualized data. These datasets bridge visual perception and symbolic reasoning: the model must not only detect visual elements, but also recover their structural relations and use them to answer questions. This type of data is crucial for models such as Qwen3-VL and MiniCPM-V, where OCR, document parsing, chart reasoning, and layout understanding are central parts of the training recipe.

Region-level grounding and spatial reasoning data provide more fine-grained supervision between language expressions and visual regions. Datasets such as Visual Genome, Flickr30k Entities~\cite{plummer2015flickr30k}, RefCOCO~\cite{kazemzadeh2014referitgame}, RefCOCO+, and RefCOCOg connect objects, phrases, attributes, relationships, and referring expressions to bounding boxes or regions. More recent works such as Kosmos-2~\cite{peng2024grounding} and GLaMM~\cite{rasheed2024glamm} extend this idea to grounded image-text pairs and pixel-level grounding conversations. This kind of data moves multimodal understanding from image-level recognition to evidence-level localization. It helps models answer not only ``what is in the image'', but also ``where it is,'' ``which object is being referred to,'' and ``which visual evidence supports the answer.'' Such grounding ability is essential for reliable visual question answering, GUI understanding, robotics, and agentic multimodal systems.

Finally, video and audio understanding data introduce temporal and acoustic supervision. Video-text datasets such as MSR-VTT~\cite{chen2022msr}, ActivityNet Captions~\cite{krishna2017dense}, HowTo100M~\cite{miech2019howto100m}, WebVid~\cite{Bain21}, and VideoInstruct-100K~\cite{Maaz2023VideoChatGPT} train models to recognize actions, events, temporal order, scene transitions, and long-range dependencies. Unlike static image understanding, video understanding requires models to reason about state changes and event progression. Audio understanding data, including AudioSet~\cite{jort_audioset_2017}, Common Voice~\cite{ardila2020common}, LibriSpeech~\cite{panayotov2015librispeech}, Clotho~\cite{drossos2020clotho}, FSD50K~\cite{fonseca2021fsd50k}, SALMONN~\cite{tang2024salmonn}, and Qwen2-Audio-style~\cite{chu2024qwen2} training corpora, extends multimodal comprehension to speech, environmental sounds, music, speaker characteristics, and paralinguistic cues. These data sources allow native multimodal models to move from image-language understanding toward broader world understanding across visual, textual, temporal, and acoustic channels.

In a nutshell, understanding-oriented data has evolved from coarse image-text alignment to task-specific reasoning, long-context multimodal comprehension, structured document understanding, fine-grained grounding, and temporal/audio understanding. This evolution reflects a broader shift in NMM systems: the goal is no longer merely to associate images with captions, but to build models that can inspect multimodal evidence, integrate information across inputs, reason over structure and time, and produce reliable textual or structured responses grounded in the input.

\subsection{Generation-Oriented Data}

Generation-oriented data is designed to train native multimodal models to produce non-textual or mixed-modality outputs, such as images, edited images, videos, speech, audio, or interleaved image-text sequences. Compared with understanding-oriented data, which usually maps multimodal inputs to textual or structured semantic responses, generation-oriented data defines a more demanding input-output relationship: the model must synthesize perceptually plausible content while preserving semantic alignment, visual fidelity, temporal coherence, and controllability. As native multimodal models move toward unified understanding and generation, this category of data becomes increasingly important for connecting language, perception, and content creation within a single model.

The most basic form of generation-oriented data is text-to-image data, where natural language prompts or captions are paired with images. Although such data may overlap with image-text pairs used for contrastive understanding, its function in generation is different: it teaches the model to map textual descriptions into visual distributions rather than merely align image and text embeddings. Large-scale image-text corpora such as LAION-5B~\cite{schuhmann2022laion} provide broad coverage of visual concepts and styles, while higher-quality caption datasets such as COCO Captions~\cite{chen2015microsoft} are often used for evaluation or fine-tuning. More recent prompt-image datasets, such as DiffusionDB~\cite{wang2023diffusiondb} and JourneyDB~\cite{sun2023journeydb}, capture real user prompts and AI-generated images, making them useful for studying prompt distributions, aesthetic preferences, and the mismatch between natural captions and generation-oriented prompts. These datasets show that text-to-image generation requires not only semantic alignment, but also control over composition, style, object relations, and visual quality.

Image editing data further extends generation from open-ended synthesis to controllable visual transformation. The typical data triplet is a source image, an editing instruction, and a target image. InstructPix2Pix~\cite{brooks2022instructpix2pix} pioneered large-scale instruction-guided image editing by using LLMs and text-to-image diffusion models to synthesize editing triples. MagicBrush~\cite{Zhang2023MagicBrush} improved this direction with human-annotated editing data, including both single-turn and multi-turn edits. HQ-Edit~\cite{hui2024hq} and UltraEdit~\cite{zhao2024ultraeditinstructionbasedfinegrainedimage} scale instruction-based editing with higher-quality source-target pairs, more diverse edit types, and region-level constraints. Unlike text-to-image data, editing data requires the model to preserve irrelevant regions while modifying only the target content given instructions. Thus, it provides supervision for locality, identity preservation, style transfer, object replacement, and instruction faithfulness.

Another important branch is controllable or grounded generation data, where text prompts are augmented with explicit structural conditions such as bounding boxes, masks, sketches, depth maps, edge maps, layouts, segmentation maps, or human poses. Works such as ControlNet~\cite{zhang2023adding} and T2I-Adapter~\cite{mou2024t2i} demonstrate that adding external visual conditions can significantly improve spatial control in image generation. GLIGEN~\cite{li2023gligen} introduces grounded text-to-image generation with caption and bounding-box conditions, while Composer decomposes visual generation into multiple controllable factors such as depth, sketch, color, and layout. This kind of data addresses a central limitation of pure text-conditioned generation: natural language alone is often insufficient for precise spatial arrangement and local control. By providing structured conditions, controllable generation data helps models learn where objects should appear, how they should be arranged, and how local constraints interact with global semantics.

For NMM systems, interleaved image-text generation data is especially important because the target is no longer a single image, but a coherent multimodal sequence. In this setting, a model may be asked to generate alternating text and images, visual stories, illustrated explanations, or multi-step multimodal outputs. Datasets and frameworks such as VIST, OpenLEAF, CoMM, and InterSyn~\cite{kim2021conditional,an2023openleaf,ma2025intersyn,chen2025comm} explore this direction by organizing generation targets as sequences of text and images. Unified models such as Emu3.5, BAGEL, and LLaDA2.0-Uni also rely on interleaved generation data to connect understanding, reasoning, and generation. Compared with isolated text-to-image pairs, interleaved generation data requires stronger entity consistency, discourse coherence, visual style consistency, and long-range dependency modeling. It is therefore a key data format for models that aim to generate multimodal documents, tutorials, stories, or step-by-step visual outputs.

Video generation data introduces temporal supervision into multimodal generation. Its common formats include text-to-video, image-to-video, text-image-to-video, speech-to-video, and animation data. Compared with image generation, video generation data must encode not only visual semantics and aesthetics, but also motion quality, object persistence, temporal transitions, camera movement, and physical plausibility. Large-scale video-text datasets such as WebVid-10M~\cite{Bain21} and Panda-70M~\cite{chen2024panda70m} provide broad video-caption supervision, while OpenVid-1M~\cite{nan2024openvid} and VidGen-1M~\cite{tan2024vidgen} focus more on high-quality video-text pairs for generative training. Recent video generation systems such as Wan2.1/Wan2.2 and HunyuanVideo-1.5 further highlight the importance of data filtering, caption rewriting, aesthetic scoring, motion quality assessment, bilingual text-video alignment, and progressive training strategies. These examples show that video generation data is not simply an extension of image-text data to more frames; it requires explicit consideration of temporal consistency and dynamic world modeling.

Audio and speech generation data expands generation-oriented supervision beyond visual outputs. Typical tasks include text-to-speech, speech response generation, voice cloning, text-to-audio, and music generation. Datasets such as LibriTTS~\cite{zen2019libritts}, LibriTTS-R~\cite{koizumi2023libritts}, VCTK~\cite{yamagishi2019cstr}, GigaSpeech~\cite{chen2021gigaspeech}, and Emilia~\cite{he2024emilia} provide speech data for modeling pronunciation, speaker identity, prosody, multilingual speech, and naturalness. For general audio generation, datasets and models such as AudioCaps~\cite{kim-NAACL-HLT-2019}, WavCaps~\cite{mei2023wavcaps}, AudioLDM~\cite{liu2023audioldm}, MusicCaps~\cite{agostinelli2023musiclm}, and LP-MusicCaps~\cite{doh2023lp} connect text descriptions with environmental sounds, acoustic events, or music. In omni-modal systems such as MiniCPM-o, Ming-Omni, and OmniVoice, speech generation data is especially important since the model is expected not only to understand audio, but also to respond with natural, expressive, and context-aware speech. These data introduces supervision for real-time interaction, speaker consistency, emotion, rhythm, and paralinguistic expression.

Overall, generation-oriented data evolves from text-conditioned image synthesis to controllable editing, structured generation, interleaved multimodal generation, temporal video generation, and speech/audio generation. This progression reflects a broader shift in native multimodal models: they are no longer limited to perceiving and describing multimodal inputs, but are increasingly expected to create, modify, and organize multimodal content. The main challenge is that generative supervision must satisfy multiple constraints simultaneously, including semantic alignment, perceptual quality, controllability, temporal coherence, identity preservation, and human preference. As a result, generation-oriented data is often combined with preference and reward data, but its core role remains distinct: it defines the input-output mappings through which native multimodal models learn to synthesize new multimodal content.

\subsection{Interaction-Oriented Data}

Interaction-oriented data is designed to train native multimodal models to act in external environments rather than only understand or generate content. Its defining feature is that the supervision target is no longer a textual answer or a generated image, but an executable action, tool call, or action trajectory. The typical data format can be described as a task goal, an observation history, and a sequence of actions, where the observation may include webpages, screenshots, UI hierarchies, documents, tool outputs, videos, or embodied visual states. This data category is central to the development of multimodal agents because it connects perception, reasoning, planning, and execution.

Web interaction data is one of the earliest and most representative forms of interaction-oriented supervision. In this setting, the model receives a user goal and must operate a browser environment through actions such as searching, clicking, typing, scrolling, selecting options, and navigating across web pages. WebShop~\cite{yao2022webshop} introduced a simulated e-commerce environment for language-grounded web interaction, requiring agents to search and purchase products according to natural language instructions. Mind2Web~\cite{deng2023mind2web} extended this direction to real websites, collecting human action sequences across diverse domains and tasks. WebArena~\cite{zhou2024webarena} further emphasized executable evaluation in reproducible web environments, while VisualWebArena~\cite{koh2024visualwebarena} showed that many web tasks require visual grounding over rendered webpages rather than relying only on HTML or text. WebLINX~\cite{lù2024weblinx} and WebVoyager~\cite{he2024webvoyager} also highlight the importance of multi-turn navigation, real websites, screenshots, and open-ended task completion. These datasets shift web page understanding to goal-directed operation.

Mobile and desktop GUI interaction data further broadens the action space from browser-specific operations to general interface control. Such data usually pairs screenshots or UI trees with natural language instructions and low-level actions such as tap, click, type, drag, scroll, or open an application. RICO provides large-scale mobile UI screens and hierarchy information, forming an early foundation for UI understanding. Android in the Wild collects large-scale Android operation episodes with natural language goals, screenshots, and human demonstrations, making it a representative dataset for mobile device control. ScreenAI~\cite{baechler2024screenai} focuses on UI and infographic understanding, while SeeClick~\cite{cheng2024seeclick} and ScreenSpot emphasize GUI grounding, where models must locate actionable elements from screenshots~\cite{cheng2024seeclick}. CogAgent, OmniACT, OSWorld, and Windows Agent Arena~\cite{hong2023cogagent,kapoor2024omniact,OSWorld} further extend this direction to more general desktop and operating-system environments, where agents must complete multi-step tasks across real applications. Compared with web-only data, GUI data requires stronger visual grounding, coordinate prediction, layout understanding, and long-horizon action planning.

Embodied and robotic interaction data extends the same idea from digital environments to physical action. Here the observation may be an image, video, robot state, or language instruction, while the output is a robot action or manipulation trajectory. ALFWorld connects language planning with embodied execution, providing a bridge between abstract instructions and grounded actions. RT-1~\cite{rt12022arxiv} demonstrates large-scale language-conditioned robot control from real-world demonstrations. BridgeData V2 and Open X-Embodiment further scale robot learning by aggregating diverse manipulation trajectories across environments, tasks, and embodiments. Recent multimodal agent frameworks such as Magma also try to unify GUI navigation and robotic manipulation under a shared action-grounding formulation. Although embodied data is often treated separately from web or GUI data, it shares the same core supervision pattern: models must map multimodal observations and task goals to executable actions.

Overall, interaction-oriented data marks a transition from passive modeling to active agency. Web data teaches models to navigate online environments; GUI data teaches them to operate visual interfaces; tool-use data teaches symbolic action and API invocation; embodied data teaches physical control. Across these settings, the key supervision signal is the action trajectory, not merely the final answer. This makes interaction-oriented data essential for training NMM systems that can perceive an environment, understand user goals, plan intermediate steps, execute actions, and adapt based on feedback.

\subsection{Preference and Alignment Data}

Preference and alignment data is mainly used in the post-training stage to calibrate the behavior of native multimodal models. Unlike understanding-oriented data, which teaches models to interpret multimodal inputs, or generation-oriented data, which teaches them to synthesize images, videos, or speech, preference data teaches models which outputs should be preferred under human, factual, safety, and task-specific criteria. Its supervision format is usually comparative or reward-based, such as a prompt with two candidate responses, a ranking among multiple generations, a human or AI preference label, a reward score, or a critique explaining why one output is better than another. Therefore, this type of data does not primarily expand modality coverage; rather, it improves helpfulness, faithfulness, safety, controllability, instruction following, and output quality.

For multimodal understanding, one of the central goals of preference data is to reduce hallucination and improve visual faithfulness. Large vision-language models often generate fluent answers that are not supported by the image, especially when questions require fine-grained visual evidence or when the model over-relies on language priors. LLaVA-RLHF~\cite{2023llavarlhf} is an early representative work that introduces RLHF into large multimodal models by collecting human preferences between candidate answers and incorporating factual information into reward modeling. RLHF-V~\cite{yu2024rlhfvtrustworthymllmsbehavior} further uses segment-level correctional feedback, where annotators identify hallucinated spans in model responses rather than only selecting a better answer. VLFeedback~\cite{li2024vlfeedback} and Silkie~\cite{2023vlfeedback} scale this direction by collecting responses from multiple vision-language models and using preference signals based on helpfulness, visual faithfulness, and ethical considerations. RLAIF-V~\cite{yu2024rlaifv} reduces the cost of alignment by replacing part of human feedback with AI feedback. More recent works such as MM-RLHF, HA-DPO, V-DPO, and CLIP-DPO~\cite{zhang2025mm, zhang2025mm,xie2024v,ouali2024clip} also construct preference pairs or reward signals to penalize hallucinated answers and encourage visually grounded responses. Together, these works show that multimodal preference data is especially important for teaching models not only to answer, but to answer based on the visual evidence actually present in the input.

Safety alignment is another major focus of preference and alignment data. Multimodal models may receive harmful requests in the form of text, images, screenshots, or image-text combinations, and they need to distinguish benign visual understanding from unsafe compliance. Safety preference data typically compares safe and unsafe responses under multimodal prompts, or separately annotates helpfulness and harmlessness. Works such as SPA-VL~\cite{zhang2024spavl} and Safe RLHF-V~\cite{ji2026safe} build vision-language safety preference data to train models that can refuse harmful multimodal instructions while still providing helpful responses for safe tasks. This direction is important because safety failures in multimodal models can arise not only from text prompts, but also from visual content, hidden text in images, screenshots, or combinations of visual and textual cues. As a result, safety preference data must consider both the semantic content of the input and the model’s response behavior.

Preference data is also widely used to align multimodal generation with human judgments. For text-to-image generation, datasets and reward models such as ImageReward, Pick-a-Pic, HPS v2, and HPD v2~\cite{xu2023imagereward,kirstain2023pick,wu2023human} collect human preferences over generated images, focusing on prompt-image alignment, realism, aesthetics, composition, and overall visual quality. Diffusion-DPO, DDPO, and DPOK~\cite{wallace2024diffusion,black2024training,fan2023dpok,yu2026advancing} further show how preference or reward signals can be used to optimize diffusion models directly. In this setting, the preferred output is not simply the one that matches the caption most literally, but the one that better satisfies human expectations of visual appeal, style, object fidelity, and controllability. Similar ideas are being extended to video generation, where preference data and reward models evaluate temporal consistency, motion quality, subject preservation, physical plausibility, and prompt-video alignment. Benchmarks and reward-oriented works such as VBench, VBench++, VBench-2.0~\cite{huang2023vbench,huang2024vbench++,zheng2025vbench}, and human-feedback-based video generation studies reflect the growing importance of multi-dimensional video preference signals. For audio and speech generation, preference data focuses on naturalness, intelligibility, speaker similarity, prosody, music quality, and text-audio alignment, as seen in emerging reward models and human preference datasets for text-to-audio and speech generation.

In addition to response and generation quality, preference and alignment data increasingly supports agentic behavior. For interaction-oriented multimodal models, the preferred output may be a better action, a more reliable tool call, or a more successful trajectory. Although much of the existing literature focuses on text-only tool use or GUI benchmarks, the same preference principle applies to multimodal agents: models should prefer actions that are correct, efficient, recoverable, and aligned with user intent. Feedback can be collected from human demonstrations, execution results, environment rewards, or critiques of failed trajectories. This suggests that future native multimodal models will likely combine visual grounding, tool use, and preference optimization in a unified post-training pipeline.

Overall, preference and alignment data acts as a behavioral calibration layer for native multimodal models. It improves factual grounding in visual question answering, reduces hallucination, enhances safety, aligns generated images, videos, and speech with human preferences, and supports more reliable tool use and action selection. Its development also shows several clear trends: from human feedback to AI-assisted feedback, from scalar preference labels to fine-grained critiques and rationales, from pairwise comparisons to multi-dimensional reward models, and from response-level alignment to generation- and action-level alignment. As native multimodal models become increasingly capable of understanding, generating, and acting, preference and alignment data becomes essential for ensuring that these capabilities are reliable, controllable, and aligned with human expectations.

\subsection{Data Mixture Across Training Stages}
The dataset categories above are not usually mixed with a single fixed ratio. In recent native multimodal systems, data proportion is stage-dependent~\cite{BAGEL7B2025, cui2026minicpm, cui2025emu35nativemultimodalmodels,ai2026llada20uniunifyingmultimodalunderstanding,ai2025ming,xie2026emergentbridge}. Early alignment stages mainly use clean paired data, such as image-caption, OCR-image-text, speech-transcription, or short video-caption pairs, so that visual or acoustic tokens become compatible with the language backbone. Later pre-training stages broaden the mixture to text-only data, image-text data, video-text data, documents, grounding, VQA, interleaved sequences, and generation data. For example, Qwen3-VL first performs vision-language alignment and then uses a large multimodal pre-training stage that roughly balances text-only and vision-language data, while Emu3.5 emphasizes long video-interleaved sequences in its native autoregressive pre-training.

For unified understanding-generation models, the central issue is how to balance discriminative understanding data with generative data. Pair data such as image-to-text or text-to-image is useful for local alignment and basic generation, but interleaved data becomes more important when the model is expected to produce coherent multimodal sequences. BAGEL makes this pattern explicit: its recipe moves from alignment data to mixtures of text, T2I, I2T, interleaved understanding, video-interleaved generation, and web-interleaved generation, with later stages increasing the weight of interleaved and generative data. LLaDA2.0-Uni and Mamoda2.5 follow the same broad motivation from diffusion or AR-diffusion modeling, where image understanding, text-to-image generation, image editing, and video generation/editing must be learned under a unified objective.

Video and omni-modal models further show why raw sample ratios are insufficient. For video generation, data quality and curriculum often matter as much as scale: HunyuanVideo-1.5 filters massive raw video collections into high-quality clips and progressively mixes T2I, T2V, and I2V data, while Wan2.2 emphasizes deduplication, caption rewriting, visual-quality filtering, motion-quality filtering, and progressive training. For omni-modal models, image, video, text, audio, and speech data are measured in different units and converge at different speeds. Ming-Omni therefore reports stage-wise mixtures over image-text, audio-text, text, video-text, and audio QA data, while OmniVoice frames data scale in speech hours and language coverage rather than image/video sample counts.

The final supervised stage is usually smaller but more behavior-oriented. Instead of preserving the pre-training mixture, SFT data is organized around target capabilities such as OCR and document understanding, chart reasoning, multi-image comparison, text-to-image generation, image editing, video generation, speech response, or interleaved dialogue. Emu3.5, for instance, reports SFT data by task families such as any-to-image generation, visual narrative, visual guidance, world exploration, and embodied manipulation, while Kling-Omni emphasizes unified video generation, editing, and instruction following. Overall, native multimodal data proportion should be viewed as a curriculum over modality, sequence length, resolution, generation target, and instruction format, rather than as a static dataset pie chart.

%% file: train.tex
\section{Training}
\label{sec:training}
This section argues that training strategies are not independent of architecture; rather, each fusion regime imposes a distinct training signature. At pretraining time, that signature is captured by five dimensions---freezing topology, learning-rate topology, loss formulation, stability prescription, and curriculum scheduling over resolutions, sequence lengths, and modality mixtures---each of which we trace across regimes. SFT and RL inherit this signature and add two further regime-specific axes: an SFT-time \emph{freezing rewiring} that only mid-fusion can perform, and an RL-time \emph{policy scope} that the fusion regime, not the algorithm, dictates. Our discussion centers on mid- and early-fusion, while late-fusion is treated mainly as a baseline against which the native regimes diverge.

As multimodal modeling has migrated from late- through mid- to early-fusion, the training stack shifts. Section~\ref{sec:training:pt} traces the fusion-coupled evolution of pre-training across the three regimes; Sections~\ref{sec:training:sft} and~\ref{sec:training:rl} show how the same fusion gradient propagates into supervised fine-tuning and reinforcement learning, respectively; and Section~\ref{sec:training:opd} closes with a discussion of \emph{On-Policy Distillation}, an emerging post-RL training paradigm.

\subsection{Pre-Training (PT)}
\label{sec:training:pt}
One paradigm to all fusion regimes: \emph{modal quantizers and VAEs are pretrained independently and remain frozen}, which applies to discrete visual tokenizers (VQ-VAE, MAGVIT-v2~\cite{Yu2023LanguageMB}, Make-a-Scene~\cite{Gafni2022MakeASceneST}), discrete audio codecs (Mimi, SpeechTokenizer, Encodec~\cite{Defossez2022HighFN}), and continuous VAEs (3D Causal VAE~\cite{Wu2024ImprovedVV}, Video DC-AE~\cite{Peng2025OpenSora2T}, Wan-VAE). The reasoning is consistent across fusion types: these components define the latent space, so changing them mid-training would invalidate all learned representations. Thus, the regime-specific differences concern only the \emph{trainable} components (e.g., continuous encoders, projectors, and the backbone) and how they connect to the loss.

\subsubsection{Late-Fusion PT}
In the late-fusion regime, such as LLaVA and Video-LLaVA, modality encoders are connectivity peripherals: features flow through a thin projector into a frozen LLM and gradients do not reach the encoder. All five training-signature dimensions are therefore degenerate---a single global learning rate suffices, the loss is text-only autoregressive cross-entropy, no stabilizers are needed beyond standard practice, and the resolution/mixture schedule is absorbed into the frozen encoder subgraph and the dataset rather than the optimizer. Visual tokens simply form a prefix to the text sequence. The trade-off is explicit: late-fusion buys training simplicity at the price of \emph{capped cross-modal capacity}, since the encoder cannot adapt to the language objective.

\begin{figure*}[t]
  \centering
  \includegraphics[width=\textwidth]{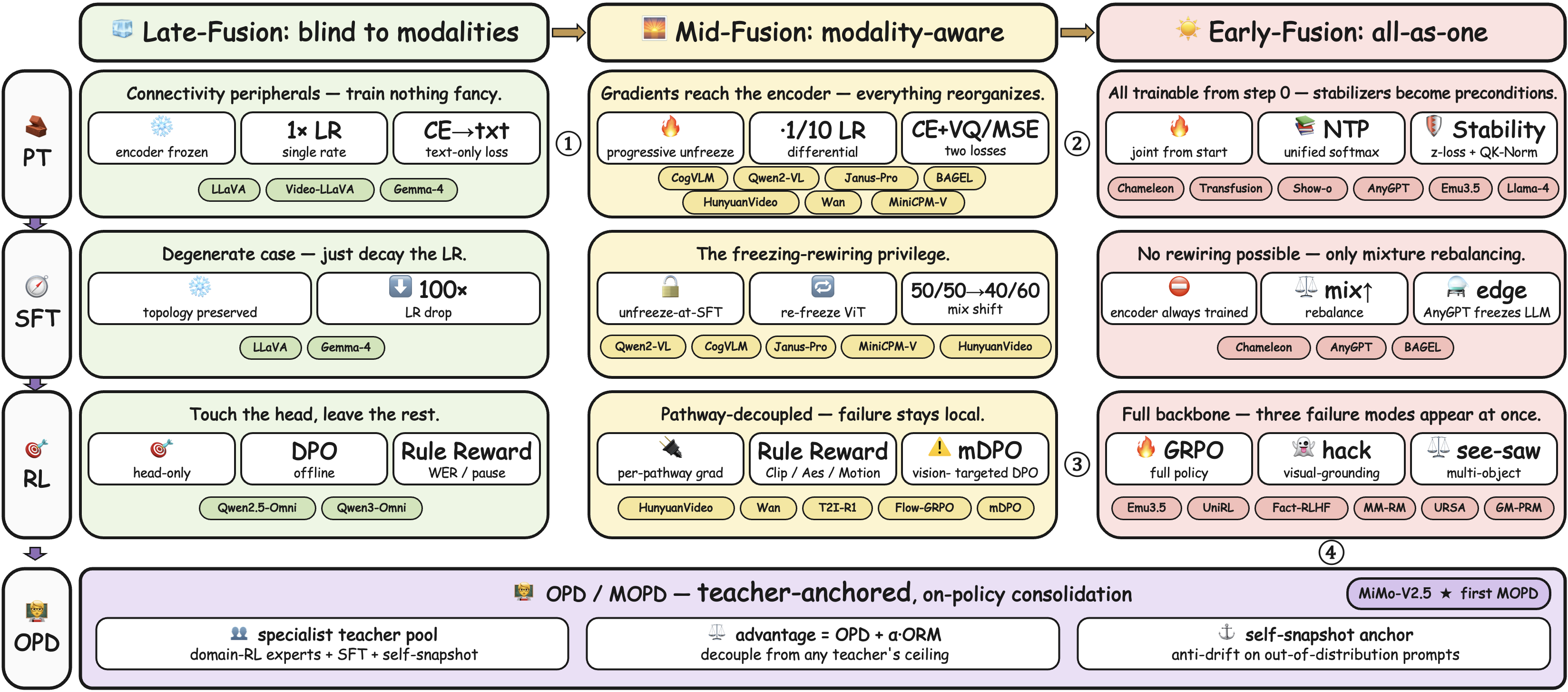}
  \caption{\textbf{Training $\times$ Fusion: a stage-by-regime grid.} Rows = training stages (PT/SFT/RL/OPD); columns = fusion regimes (late/mid/early). \textbf{Every rightward arrow is an architectural necessity, not a stylistic choice.} \Circled{1} PT late$\rightarrow$mid: differential LR becomes mandatory once the encoder receives gradients. \Circled{2} PT mid$\rightarrow$early: z-loss and QK-Norm become preconditions; the modality-mixture schedule replaces differential LR. \Circled{3} RL mid$\rightarrow$early: pathway-locality collapses, so the policy must cover the full backbone---and the three failure modes on the right (grounding hack, see-saw, perception/logic gap) co-emerge \emph{because of} this collapse, not alongside it. \Circled{4} RL$\rightarrow$OPD: teacher-anchored on-policy distillation (bottom bar, fusion-agnostic) is the structural response to \Circled{3}, combining a specialist teacher pool, a hybrid OPD$+$ORM advantage, and a self-snapshot anchor against drift.}
  \label{fig:training}
\end{figure*}

\subsubsection{Mid-Fusion PT}
Mid-fusion is the regime in which gradients first reach the encoder, and every element of the training signature is a response to that single change. Key techniques of mid-fusion are progressive unfreezing, differential rates, and decoupled losses.

\textbf{The rise of progressive unfreezing.} The defining mid-fusion pattern is the \emph{progressive unfreezing schedule}: the encoder is frozen during initial alignment, then unlocked at a later stage, sometimes with the LLM frozen during the encoder-warmup phase. Qwen2-VL exemplifies the symmetric variant, where ViT trained while LLM frozen in Stage 1 and both unfrozen in Stage 2. CogVLM, Janus-Pro, and MiniCPM-V defer encoder unfreezing to SFT. A genuinely SFT-specific variant also appears here: Qwen2-VL \emph{re-freezes} its ViT at the chat-tuning stage, signalling that once vision-text alignment is consolidated by 1.4T-token joint pretraining, further encoder updates during instruction tuning are unnecessary.

\textbf{Differential rates become mandatory.} Once gradients reach the encoder, a uniform learning rate destabilizes the system, as a single rate is typically too high for the encoder yet too low for the LLM. CogVLM applies $1/10$ of the base rate to its EVA2-CLIP-E encoder upon SFT-time unfreezing, establishing the canonical mid-fusion prescription. Stage-wise global decay performs the same role implicitly: Janus-Pro decays $10^{-3}\!\to\!10^{-4}\!\to\!4{\times}10^{-5}$ across its three stages for a $25\times$ total reduction, whereas the largest rate corresponds to adapter-only training and the smallest to full-model SFT. Although audio-centric, Moshi exhibits the identical logic, training its temporal and depth transformers at $3{\times}10^{-5}$ versus $2{\times}10^{-4}$ to create a $\sim7\times$ gap that reflects their distinct convergence dynamics. Differential learning rates are indeed constitute the fundamental condition for the mid-fusion regime.

\textbf{Decoupled loss of understanding/generation.} Mid-fusion models that handle both understanding and generation typically maintain two separate loss terms over a shared backbone. Janus-Pro computes cross-entropy on text tokens for understanding and on discrete VQ tokens for generation, with two independent visual encoders feeding a shared LLM. BAGEL routes understanding (cross-entropy on text given SigLIP features) and generation (MSE on continuous VAE latents via Next-Group-Token Prediction) through MoT, using task-specific batch toggles to prevent the two pathways from interfering. The two losses share parameters but not gradient signal at the modality-specific layers, which indicates partial rather than full unification.

\textbf{Resolution and context-length curricula become critical.} Once the encoder receives gradients, its operating resolution is no longer fixed by the pretraining checkpoint and becomes a key scheduling variable. Understanding models progress through resolution curricula across pretraining stages, in order to avoid the optimization shock of unfreezing a high-capacity encoder at full resolution from the start. For instance, MiniCPM-V $224 \rightarrow 448 \rightarrow 1344^+$ over three stages, CogVLM increasing its input from 224 to 490 in late pretraining. Omni systems extend this logic to the temporal axis: Qwen2.5-Omni grows its context window from 8{,}192 to 32{,}768 tokens in its final pretraining stage to handle long audio and video, and Emu3 raises generation resolution from 512 to 720\,px (understanding up to 1024\,px) during post-training. The key insight is that mid-fusion links \emph{which} parameters are unfrozen with \emph{at what resolution} they are unfrozen; using one schedule without the other destabilizes training. Modality mixture remains externally specified at this stage, since two pathways still have their own loss heads, so that its scheduling is at most coarse-grained (e.g., Janus-Pro's 50/50 generation/understanding split during PT).

\subsubsection{Early-Fusion PT}
Early-fusion eliminates the architectural firewall between modalities, and the training signature follows: every component carries gradients from step zero, the loss collapses to a single objective over a shared vocabulary, and the absence of any modality-specific buffer makes stabilizers essential rather than optional. Key techniques of early-fusion includes joint-from-start, unified NTP, and mandatory stabilization.

\textbf{Joint-from-start.} Early-fusion models optimize all modules simultaneously from the first step with no freeze-to-unfreeze transitions. For discrete-token early-fusion architectures such as Chameleon, Emu3.5, and AnyGPT, the vocabulary is expanded by the codebook size, including an increase of 8,192 for Chameleon, 32,768 for Emu3.5, and a combined addition of 8,192 image, 1,024 speech, and 8,192 music tokens for AnyGPT. Consequently, the new embeddings are trained end-to-end with the unified objective from the onset. These models also typically return to a single global learning rate, not because differential rates would be incorrect, but because the unified vocabulary yields a unified loss whose gradient statistics are homogeneous across token types. Specifically, Chameleon and Transfusion apply a single global rate, with Chameleon decaying from $10^{-4}$ to $10^{-5}$ and Transfusion from $3{\times}10^{-4}$ to $1.5{\times}10^{-5}$ via cosine decay, thereby relying on architectural stabilizers rather than rate engineering. The only exception is Llama-4's MetaP, which moves in the opposite direction by descending to algorithmically determined per-layer rates.

\textbf{Unified NTP and modal-aware attention.} Discrete-token early-fusion collapses every modality into a single cross-entropy loss over a shared vocabulary. Systems such as Chameleon, Emu3.5, AnyGPT, and LongCat-Next all adopt this pattern, ensuring that images, audio, and text receive identical gradient treatment within the same softmax layer. Hybrid variants emerge when continuous modalities resist tokenization. For instance, Transfusion combines language modeling with DDPM diffusion via a joint loss expressed as $\mathcal{L} = \mathcal{L}_{\text{LM}} + 5\cdot\mathcal{L}_{\text{DDPM}}$, where the scaling coefficient $\lambda = 5$ is determined via preliminary search. Similarly, Show-o combines Mask Token Prediction for images with Next Token Prediction for text, whereas LLaDA2.0-Uni adopts a discrete diffusion masked-denoising objective uniformly across text and image tokens to replace the autoregressive loss entirely. Expanding on unified NTP, OneCAT introduces a multi-scale visual autoregressive mechanism to predict image tokens from coarse to fine resolution. Within the RVQ-based audio path of Moshi, the semantic codebook receives a loss weight $\alpha_k = 100$, which stands in contrast to the acoustic codebook weight $\alpha_k = 1$ to ensure that linguistic content is prioritized over acoustic detail. Attention patterns closely track these objective functions. Pure NTP models apply causal attention uniformly across modalities, utilizing special structural tokens to demarcate modality boundaries. Conversely, hybrid models relax this constraint where their loss demands it, allowing Transfusion and Show-o to permit bidirectional attention within image regions because diffusion processes inherently benefit from full image context.

\textbf{Z-loss and QK-Norm for stability.} The unified softmax also establishes the unified divergence surface. Chameleon's ablations are unequivocal: without QK-Norm, the model diverges after approximately 20\% of training. Furthermore, z-loss regularization expressed as $10^{-5}\cdot\log^2 Z$, where $Z$ is the softmax partition function, is required to keep logits bounded across the heterogeneous token distribution. These interventions are not generic transformer tricks, but rather the engineering preconditions for scaling discrete-token early-fusion. This constitutes the clearest evidence that early-fusion is a distinct training regime, not a stylistic refinement of mid-fusion.

\textbf{Modality-mixture scheduling.} The key change at the early-fusion boundary is that separate loss heads per modality are eliminated, meaning the modality mixture in each batch directly determines the gradient direction. This transition forces every early-fusion system to treat mixture scheduling as a core training hyperparameter.

Transfusion fixes a 1:1 text-to-image token ratio with captions preceding their corresponding images 80\% of the time, using BOI and EOI tokens as attention-pattern triggers that switch between causal and bidirectional regions on the fly. Chameleon interleaves image and text tokens with special delimiters, mixing image-text pairs and text-only documents in the same batch while fixing each image at 1{,}024 tokens derived from a 512$\times$512 center-cropped crop regardless of resolution, which ensures that the gradient per image remains constant. Moshi advances this schedule further by interleaving text and audio at the 12.5\,Hz frame level, combining one text position and eight audio codebook positions per timestep. Crucially, half of the pretraining batches are allocated as text-only data to serve as an explicit anti-forgetting buffer for language capability under the unified loss. Video generators with unified objectives such as Open-Sora 2.0 schedule the task from text-to-video to image-to-video by prepending reference-frame latents, and run resolution curricula in parallel starting from 256\,px and scaling up to 768\,px. This strategy is mirrored in HunyuanVideo's escalation from 256\,px to 960\,px and Wan's transition from 256\,px to 720\,px, maintaining mixed image-video batches throughout the process.

Chameleon's documented warning explicitly highlights the primary failure mode, stating that imbalanced modality mixtures cause early-fusion models to learn degenerate unconditional priors that distort generation. This stands as a degenerate behavior that late-fusion with its inert generation pathway and mid-fusion with its decoupled losses avoid structurally. Modality-mixture scheduling is thus the early-fusion equivalent of differential learning rates, representing a fundamental precondition for successful training rather than just an optimization refinement.

\subsection{Supervised Fine-Tuning (SFT)}
\label{sec:training:sft}
The five-dimensional PT signature carries forward into SFT, but two new regime-specific axes appear on top of it. First, mid-fusion gains a free freezing rewiring privilege, meaning that SFT may unfreeze the encoder that PT had kept frozen or re-freeze a component that PT had been training. Neither late-fusion nor early-fusion can exercise this option, since nothing was trainable to begin with in late-fusion and the joint-throughout commitment forecloses re-freezing in early-fusion. Second, the curriculum-scheduling dimension that PT introduced now expresses itself as distribution rebalancing, reflecting the reality that SFT corpora are dramatically smaller and skew text-heavy. Consequently, recovering an appropriate modality mixture on this smaller corpus stands as a regime-specific design step. Layered on top of these is a regime-independent universal layer encompassing prompt-token loss masking, learning-rate decay relative to PT, and light dropout, which we will not repeat per regime. The discussion below emphasizes the regime-specific aspect exclusively.

\subsubsection{Late-Fusion SFT}
Late-fusion SFT represents the degenerate case of the aforementioned framework because neither freezing rewiring nor distribution rebalancing applies, rendering the universal layer essentially the entire story. The pretraining freezing topology is preserved intact, keeping the encoder frozen while the projector and LLM are tuned. This preservation is accompanied by a sharp learning-rate decay between stages, exemplified by LLaVA's $100\times$ times drop during the transition from projector-only Stage 1 to end-to-end Stage 2.

\subsubsection{Mid-Fusion SFT}
Mid-fusion represents the unique regime in which SFT is permitted to rewire the freezing topology, yielding two distinct rewiring strategies.

\textbf{(i) Unfreeze-at-SFT.} CogVLM, Janus-Pro, and MiniCPM-V maintain a frozen encoder during pretraining and unlock it exclusively at the SFT stage, utilizing a reduced learning rate such as one-tenth of the base rate for CogVLM. Janus-Pro selectively keeps the generation tokenizer frozen even after the understanding encoder is unlocked, thereby exemplifying the asymmetric component-by-component thaw that mid-fusion permits.

\textbf{(ii) Train-then-re-freeze.} Qwen2-VL re-freezes its ViT at the SFT stage after training it through both pretraining stages, leaving only the LLM tuned on ChatML conversations. This pattern stands as an exclusive characteristic of mid-fusion, since the encoder in late-fusion was never trained to begin with and the joint-throughout commitment in early-fusion forecloses any potential re-freezing.

Beyond freezing modifications, mid-fusion SFT inherits the decoupled structure of mid-fusion PT and continues to schedule the two pathways independently. On the understanding side, this architecture manifests as fine-grained modality-mix rebalancing. For instance, Janus-Pro shifts its generation to understanding data ratio from 50/50 during Stage II PT to 40/60 during Stage III SFT to bias the model toward understanding without losing generation capability. This shift illustrates that pretraining establishes the resolution curriculum while SFT subsequently adjusts the modality mix. On the generation side, the same pathway-specific logic produces an SFT recipe that operates entirely on the diffusion transformer with all VAEs and text encoders frozen. This recipe concentrates instead on two coupled curricula, namely resolution escalation and quality-filtered data tightening. In this context, HunyuanVideo's final stage utilizes approximately 1M human-annotated samples scored on aesthetic and motion criteria, Wan applies resolution-dependent quality filters, and Open-Sora 2.0 simultaneously shifts its task curriculum from text-to-video to image-to-video by prepending reference-frame latents. Across both pathways, the SFT-time tightening that the universal layer expresses on the optimizer side through a lower learning rate, prompt-token loss masking, and dropout is mirrored on the data side as a narrower distribution, stricter quality bar, and more targeted task mix. Crucially, the two pathways are tightened on their own independent schedules, which represents the exact privilege that decoupled losses confer.

\subsubsection{Early-Fusion SFT}
Early-fusion forecloses every freezing rewiring available to mid-fusion, because the joint-from-start commitment of PT means there is no frozen component left to thaw and no trained component that can be safely re-frozen without breaking the unified softmax surface. SFT therefore reduces to operating purely on the universal layer, encompassing a lower learning rate, prompt-token loss masking, and additional dropout, such as the 0.05 dropout added by Chameleon at the 34B scale. Alongside these adjustments, the system executes one regime-specific responsibility by re-balancing the modality mixture for the SFT data distribution. Because instruction-tuning corpora are dramatically smaller than pretraining corpora and skew heavily toward text-heavy conversational data, the identical imbalance that PT already had to manage resurfaces with sharper consequences at SFT. Consequently, recovering the PT-time mixture ratios on the smaller SFT corpus stands as a critical regime-specific design step.

A characteristic edge case lives directly at the boundary of this regime, where AnyGPT inverts the typical pattern by freezing its LLM backbone at the SFT stage and updating only the newly added multimodal embedding and prediction layers over 5{,}000 steps. This strategy preserves pretrained language capabilities while adapting only the modality interface, representing the extreme of training conservatism, whereas BAGEL's all-trainable, all-stage joint optimization stands as its exact opposite.

\subsection{Reinforcement Learning (RL)}
\label{sec:training:rl}
Where SFT modulates the \emph{trainable scope} through freezing rewirings in mid-fusion or their absence in early-fusion, RL inherits the same scope question and answers it under a different objective, asking what subset of parameters a reward signal touches and at what cost. The fusion regime rather than the algorithm dictates the answer, rendering this scope decision the single most consequential RL design choice for native multimodal models.

Before turning to regime-specific scopes, we note the regime-independent toolkit on which all three regimes draw. Algorithmically, three families dominate: DPO, PPO~\cite{schulman2017proximalpolicyoptimizationalgorithms}, and GRPO~\cite{shao2024deepseekmath}. Online methods incur higher sampling costs but offer stronger exploration when preference data only sparsely covers the target distribution~\cite{xu2024dposuperiorppollm,song2024importanceonlinedataunderstanding}. Reward design likewise spans three families, encompassing \emph{outcome-level} feedback with one scalar per response that remains vulnerable to hacking, \emph{process-level} feedback via multimodal PRMs, and \emph{rule-based} deterministic verifiers such as math match, code execution, CLIP or ImageReward, and various aesthetic and motion scorers. Multimodal-specific variants such as mDPO, Fact-RLHF, MM-RM~\cite{li2025devildetailstacklingunimodal}, URSA, and GM-PRM~\cite{zhang2025gmprmgenerativemultimodalprocess} are introduced where they appear in the regime-specific discussion below. The toolkit itself is shared across regimes, whereas what changes is which subset is admissible, the parameters it acts on, and the failure modes that surface.

\subsubsection{Late-Fusion RL}
Late-fusion RL is structurally minimal because when the architecture cleanly separates a quality-localizable head from the rest of the model, RL targets that head alone. Specifically, Qwen2.5-Omni and Qwen3-Omni apply DPO only to the Talker over word-error-rate or pause-error-ranked triplets, leaving the Thinker and all encoders untouched. The toolkit consequently collapses to its simplest configuration of offline DPO with rule-based scoring, making the regime largely immune to multimodal failure modes by construction. Because the projector is too thin to overwhelm visual evidence, the policy cannot drift far from its visual conditioning even under naive optimization. Late-fusion thus serves as the trivial reference case, just as it does for PT and SFT.

\subsubsection{Mid-Fusion RL}
Mid-fusion RL inherits the pathway-decoupled trainable set from mid-fusion SFT and applies the gradient only to the pathway being optimized. Video and image generators such as HunyuanVideo, Wan, T2I-R1~\cite{jiang2025t2ir1reinforcingimagegeneration}, and Flow-GRPO~\cite{liu2025flowgrpotrainingflowmatching} keep the VAE and text encoder frozen and route RL gradients only into the diffusion transformer. This routing typically employs rule-based rewards including CLIP, ImageReward, and aesthetic and motion scorers whose pathway-locality matches the pathway-locality of the update. The decoupled losses that defined mid-fusion PT and SFT thus translate directly into pathway-specific RL.

The first regime-specific failure mode also surfaces here on the understanding pathway. Naive DPO on multimodal preference pairs may rely too heavily on language priors and ignore the image condition, ultimately learning text-only preferences~\cite{wang2024mdpoconditionalpreferenceoptimization,rao2026understandinggenerationfightdiagnostic}. To counteract this, mDPO~\cite{wang2024mdpoconditionalpreferenceoptimization} explicitly conditions the preference loss on the image so that the chosen-versus-rejected gap depends directly on visual input. The pathway decoupling that simplifies mid-fusion RL also localizes this failure, ensuring it is contained within the understanding pathway and does not propagate to the generation side, which is precisely why mid-fusion RL remains tractable in practice.

\subsubsection{Early-Fusion RL}
The pathway-locality that made mid-fusion RL tractable is unavailable in early-fusion RL. Under a unified softmax, there is no isolated head whose update leaves the rest of the model unchanged, meaning the RL scope necessarily expands to the full backbone. Omni-generation systems such as Emu3.5 and UniRL~\cite{mao2025unirlselfimprovingunifiedmultimodal} update the entire policy with GRPO, and the same unified softmax that forces this expansion also delivers its main reward, allowing a single scalar arriving at any output token to credit-assign across modalities through the shared parameters, a capability that remains impossible in the decoupled pathways of mid-fusion. Reward models are typically initialized from the SFT checkpoint so that the policy and reward share representations natively.

This expanded scope, however, also exposes the policy to two failure modes that the decoupling of mid-fusion had structurally suppressed.

\textbf{Visual-grounding hacking.} The full-policy update can drive textual reward proxies such as length, formatting, and certainty up without grounding claims in the image. Fact-RLHF~\cite{sun2023aligninglargemultimodalmodels} and shortcut-aware MM-RM address this from the reward side, while complementary policy-side responses add explicit visual-faithfulness terms to the RL objective.

\textbf{Perceptual versus logical errors in process supervision.} Multimodal CoT errors split into logical mistakes encompassing computation and derivation and perceptual mistakes such as misreading a chart or mislocalizing a region. Outcome-only RL conflates these two types, but process-level rewards via multimodal PRMs such as URSA~\cite{luo2025unlockingmultimodalmathematicalreasoning} and GM-PRM separate them. This distinction becomes essential in early-fusion, where both error types route through the identical set of parameters.

Both failure modes share a single underlying mechanism, because under the unified softmax, language priors compete with visual evidence on equal footing, and naive RL lets the priors win. A second, equally structural cost compounds this issue, since when each capability including math, code, agentic tool-use, instruction following, and safety is improved by its own specialized RL run, the resulting checkpoints trade off against one another. Improving one capability regresses others, creating the well-known \emph{see-saw} effect of multi-objective post-training. Both costs motivate the post-RL primitive discussed next.

\subsection{On-Policy Distillation (OPD)}
\label{sec:training:opd}
OPD, and its multi-teacher form (MOPD), is emerging as the response. The method is a single-line modification of GRPO, by replacing the group relative advantage with a stop-gradient reverse-KL log-ratio against a teacher: $\hat{A}_{i,t}=\mathrm{sg}\!\left[\log\pi_{\text{teacher}}(y_{i,t}\!\mid\!x,y_{i,<t})/\pi_{\text{student}}(y_{i,t}\!\mid\!x,y_{i,<t})\right]$. Therefore, every token sampled from the student receives dense, per-position teacher supervision while remaining on-policy.

\textbf{MOPD on native multimodal models.} MiMo-V2.5 provides the first publicly reported deployment of MOPD on a native multimodal model: text PT $\rightarrow$ projector warmup $\rightarrow$ multimodal PT $\rightarrow$ SFT and agentic post-training (context progressively extended from 32K to 1M) $\rightarrow$ \emph{RL and MOPD}. MiMo-V2.5 places MOPD as the terminal consolidation step, explicitly tasked with strengthening perception, reasoning, and agentic capabilities in one shared backbone, consisting of three structural pieces: (i) a pool of specialist teachers obtained by independent domain RL; (ii) an outcome-reward augmentation $\hat{A}_{i,t}=\hat{A}^{\text{OPD}}_{i,t}+\alpha\,\hat{A}^{\text{ORM}}_{i,t}$ that decouples the student from any single teacher's ceiling; and (iii) a permissive teacher pool that admits domain SFT models, RL specialists, and a frozen snapshot of the student itself, where the snapshot acts as an anti-drift anchor on prompts where the other teachers would push the student into unfamiliar territory.

%% file: inference.tex
\section{Inference \& Deployment}\label{inference}

\subsection{Mitigating Sequence Explosion in Long-Context Multimodal Inference}

Native multimodal pretraining substantially amplifies the classical long-context problem. A high-resolution image, a multi-image document, or a long video is no longer a compact side feature, but is converted into hundreds, thousands, or even millions of visual and temporal tokens that must coexist with language tokens in the same context window. Consequently, inference efficiency is governed not only by the number of model parameters, but also by prefill cost, KV-cache capacity, memory bandwidth, and cross-device communication. Systems such as Gemini 1.5~\cite{gemini2024gemini15} and Gemini 2.5~\cite{gemini2025gemini25} show that multimodal contexts are already moving toward the million-token regime. Recent work therefore attacks sequence explosion from two complementary directions: reducing the number of multimodal tokens that enter the backbone, and redesigning the backbone or serving system so that very long streams can be processed without exhausting device memory~\cite{shao2026surveytokencompressionefficient}.

\paragraph{Visual Resampling and Token Compression.}
The first line of work compresses visual features before, during, or immediately after visual encoding. Fixed-budget resamplers and pooling modules map dense patch grids into a small number of latent tokens, thereby stabilizing prefill latency regardless of the original image resolution. This idea appears in production-oriented models such as MiniCPM-V 4.5 and Gemma3~\cite{gemma2025gemma3}, where image and video features are summarized into compact visual sequences before being passed to the language backbone. More adaptive methods further observe that most visual tokens are redundant for a given query. VisionZip~\cite{visionzip2025}, SparseVLM~\cite{sparsevlm2025}, FitPrune~\cite{fitprune2025}, and LLaVA-PruMerge~\cite{shang2024llavaprumerge} select, prune, recycle, or merge visual tokens according to information density, attention behavior, or similarity structure, while trainable methods such as VisionSelector~\cite{visionselector2025} and LaCo~\cite{laco2025} move compression into the learned visual pathway itself. The intuition is that multimodal reasoning rarely requires preserving every patch with equal fidelity: global semantics, task-relevant regions, and fine-grained details should receive different token budgets.

\paragraph{Dynamic Resolution and Spatially Sparse Perception.}
The second line avoids generating unnecessary visual tokens in the first place. Dynamic-resolution models encode images according to their native aspect ratio and information density, rather than forcing all inputs into a fixed square canvas. Qwen2-VL~\cite{wang2024qwen2} and Qwen2.5-VL introduce dynamic visual tokenization and multimodal rotary position encodings so that image and video tokens remain spatially and temporally grounded under arbitrary resolutions. Qwen3-VL extends this trajectory with longer contexts, improved interleaved position modeling, and stronger temporal grounding for video. Related systems such as LLaVA-UHD~\cite{llavauhd2024}, LLaVA-OneVision~\cite{llavaonevision2024}, Oryx~\cite{oryx2024}, and InternVL 2.5~\cite{internvl25_2024} use AnyRes-style slicing, spatial schemas, or on-demand compression to preserve high-resolution details while preventing token counts from growing mechanically with pixel count. More recent query-aware approaches, including Q-Zoom~\cite{shi2026q}, further make the resolution decision conditional on the user instruction: the model first reasons over a coarse view, then spends high-resolution tokens only on regions likely to affect the answer.

\subsection{Addressing the Dual Challenges of Heterogeneity and Scale in MLLMs}

In the progression toward artificial general intelligence, MLLMs must reconcile the dual challenges of heterogeneity and scale \cite{zhao2026unifiedmultimodalunderstandinggeneration, 10.1145/3718958.3750472}. Heterogeneity manifests as a fundamental representational chasm, reflecting the reality that human language is abstract, discrete, and symbolically structured, whereas visual, auditory, and sensory signals remain high-dimensional, continuous, and grounded in physical observables. This disparity extends beyond mere modality-specific encoding schemes, for it encompasses divergent information densities, temporal granularities, and noise characteristics, creating a profound semantic alignment gap that complicates unified reasoning. Meanwhile, scale introduces its own formidable barrier. As these models expand to trillion-parameter scales and attempt to process increasingly long multimodal contexts spanning thousands of high-resolution images, video streams, or audio sequences, the quadratic computational complexity of attention mechanisms \cite{10123038} escalates into a prohibitive bottleneck. The resulting surge in activation memory, inter-layer communication volume, and gradient synchronization overhead collides directly with the physical constraints of modern acceleration hardware, where high-bandwidth memory capacity and interconnect bandwidth remain strictly finite resources. Therefore, the field faces a tension that is simultaneously algorithmic, centered on how to bridge fundamentally incompatible signal modalities within a shared representational space, and systems-level, focusing on how to sustain scalable training and inference without violating the power, memory, and throughput limits of contemporary computing clusters.

\paragraph{Resolving Mismatches through Pure Discrete Tokenization} Early multimodal systems relied on continuous embedding paradigms that severely exacerbated memory bandwidth congestion~\cite{liang2025comprehensivesurveyguidemultimodal, kong2026tokenreductionefficiencygenerative}. To resolve this, frontier research has decisively shifted toward pure discrete tokenization~\cite{11455337}, a strategy that vector-quantizes high-dimensional continuous signals into finite discrete integer identifiers. Chameleon exemplifies this by employing an 8,192-entry independent image codebook to process unified one-dimensional sequences, eliminating hardware-level branching overhead. To handle extreme resolutions, Emu3.5 radically expands the discrete image vocabulary and utilizes feature distillation. By completely abandoning diffusion models, Emu3.5 proves that a single transformer can achieve mixed-modality sequence training purely through next-token prediction. Seedance 2.0 extends this system-level advantage by standardizing up to 12 channels of mixed inputs into unified spatiotemporal and waveform tokens~\cite{mousavi2025discreteaudiotokenssurvey} for shared parallel processing. AnyGPT similarly validates the universality of this discrete data-level preprocessing for arbitrary-modality dialogue.

\paragraph{Optimizing Routing via MoE and Hybrid Paradigms} Although discretization mitigates bandwidth issues, the sequence explosion from high-dimensional flattening still poses severe computational bottlenecks~\cite{shao2026surveytokencompressionefficient}. To bypass physical inference limits, architectures are transitioning to MoE~\cite{10937907}. Kimi2.5 leverages rigorous routing strategies to prune activations, supporting multi-agent concurrent inference at an extremely low cost. To address structural modality differences, Janus-Pro introduces fine-grained, isolated experts~\cite{dai-etal-2024-deepseekmoe} to achieve modality-aware implicit computational bifurcation. Despite the dominance of discrete representations, continuous diffusion retains irreplaceable advantages in high-fidelity visual synthesis~\cite{Yin_2025_CVPR}. Transfusion pioneers a hybrid method that simultaneously optimizes discrete AR and continuous denoising objectives. However, forcibly nesting causal~\cite{chen2024tokenpredictionmultimodalintelligence} and bidirectional~\cite{10123038} masks breaks the memory alignment assumptions of foundational operators like FlashAttention~\cite{dao2022flashattentionfastmemoryefficientexact, dao2023flashattention2}. To salvage this mask topology conflict, FlexAttention~\cite{dong2024flexattentionprogrammingmodel,liu2025efficienttrainingdiffusionmixtureofexperts} employs just-in-time compilation to dynamically generate fused computation graphs, while FlashMask~\cite{wang2025flashmaskefficientrichmask} enables rapid foundational switching between causal and bidirectional blocks. The liberation of these tensor operations ultimately actualizes the enterprise-scale deployment of hybrid multimodal architectures~\cite{hu2026evolutionvideogenerativefoundations}.

\subsection{Real-Time Streaming and Full-Duplex Deployment of NMM systems}

To address the latency bottleneck and first-token delay caused by dynamically arriving multimodal streams, current NMM systems are gradually shifting from static offline generation toward a unified inference paradigm centered on streaming decoding, duplex concurrency, and resource-adaptive serving. In this setting, TTFT, sustained latency, and real-time responsiveness become first-class optimization targets rather than secondary deployment considerations.

\paragraph{Incremental Multimodal Token Decoding.}
A first technical route is incremental multimodal token decoding, which avoids the conventional wait-for-completion paradigm. Instead of deferring responses until an entire visual or acoustic sequence has been encoded, recent models progressively emit visual or audio tokens in an autoregressive manner, enabling patch-by-patch, frame-by-frame, and streaming generation for lower TTFT and smoother interaction continuity~\cite{defossez2024moshi,fang2025llama,xu2025qwen3,cheng2026ar,bruce2024genie,pang2026next,ren2025next}. In practice, this route is increasingly coupled with adaptive visual granularity and dynamic input reduction, so that only the most task-relevant visual tokens are retained during streaming inference~\cite{wang2024qwen2,lan2024avg,lin2025adaptvision,liao2026resadapt}.

\paragraph{Full-Duplex State Management.}
A second route is full-duplex state management, designed to support concurrent inference over incoming sensory streams and outgoing generation streams. In real-time M2M settings, the model must process visual and audio inputs while simultaneously producing text, speech, or images, which has motivated duplex dialogue control, streaming state prediction, and dynamic KV-cache management to mitigate cache contention and sequential blocking~\cite{ma2025language,zhang2025llm,chen2025fireredchat,wang2025end,zhang2026think,lu2026aura,ning2024inf}. The key challenge here is no longer unimodal decoding speed alone, but stable coordination between input accumulation, intermediate state updates, and output generation under continuous streams.

\paragraph{Inference-Time Adaptive Bitrate Control.}
A third route is inference-time adaptive bitrate control, which aims to trade off fidelity and latency by dynamically reducing the granularity of discrete visual codes under runtime bandwidth constraints. Although direct RVQ-layer switching remains underexplored in native multimodal generation, related efforts have begun to connect adaptive tokenized perception with multimodal semantic communication and bandwidth-aware token transmission~\cite{jiang2025m4sc,qiao2025token}. Broadly, this route is closely related to runtime visual token budgeting, where adaptive resolution selection and visual token compression serve as approximations to bitrate-aware streaming.

\paragraph{Modality-Aware Mixed Quantization and Resource-Adaptive Compression.}
A fourth route focuses on modality-aware mixed quantization and resource-adaptive compression. Rather than uniformly compressing the entire model, recent works assign different precisions to the visual encoder, projector, and language backbone, thereby reducing memory and latency while preserving multimodal fidelity~\cite{yu2025mquant,li2025mbq,zhangmodality,wang2026lqa,xue2025vlmq,wang2024q,das2026towards,wang2025bi,qin2026veq}. This line is increasingly combined with runtime-aware visual simplification---including dynamic resolution degradation, adaptive preprocessing, token pruning, and energy-aware visual reduction---so that edge systems can lower the number of visual input tokens according to latency, energy, or hardware pressure instead of relying on fixed training-time resizing rules~\cite{xu2025learning,cahyani2025input,he2026energy,debnath2026llmind,zhang2025adaptinfer,liang2025dynamic,shi2026q}.

%% file: eval.tex
\section{Evaluation}
\label{eval}
Evaluating native multimodal models requires benchmarks that span both \textit{understanding} (perception, reasoning, grounding) and \textit{generation} (synthesis, editing, controllability) across modalities. Unlike earlier modular systems whose evaluation focused primarily on image--text comprehension, native architectures demand assessment of whether deep cross-modal fusion translates into improved performance on \emph{both} axes simultaneously without degradation on either. We organize the evaluation landscape by modality---image (\S\ref{sec:eval_image}), audio (\S\ref{sec:eval_audio}), and video (\S\ref{sec:eval_video})---and within each modality distinguish understanding from generation benchmarks. Table~\ref{tab:eval_benchmarks} provides a consolidated summary.

\subsection{Image}
\label{sec:eval_image}

\paragraph{Understanding Benchmarks.}
Image understanding evaluation for NMM systems follow a hierarchy of capabilities: general perception, knowledge-intensive reasoning, hallucination diagnosis, and document comprehension.

At the \textit{general perception} level, VQAv2 and GQA assess open-ended visual question answering with balanced answer distributions, while SEED-Bench extends evaluation to 12 dimensions including spatial reasoning, action recognition, and instance interaction. MMBench and MMStar adopt bilingual multiple-choice formats that reduce evaluation noise from free-form generation and specifically address data leakage concerns. For \textit{knowledge-intensive reasoning}, MMMU curates college-level problems spanning 30 subjects that require joint domain knowledge and visual interpretation, whereas MathVista isolates mathematical reasoning grounded in visual contexts. These benchmarks are particularly diagnostic for native models: because understanding and generation share a single backbone, verifying that generation-capable models preserve strong comprehension is critical~\cite{rao2026understandinggenerationfightdiagnostic}.

\textit{Hallucination evaluation} is especially pertinent for unified architectures, where generative priors in the shared representation space may leak into discriminative predictions. POPE probes object hallucination through polling-based binary questions, while RLHF-V provides segment-level hallucination annotation for finer-grained diagnosis. Training-time mitigation methods such as mDPO and HA-DPO~\cite{zhao2024hallucinationsenhancinglvlmshallucinationaware} leverage these evaluation signals as preference data, demonstrating the tight coupling between benchmark design and alignment objectives. \textit{Document and OCR} benchmarks---DocVQA, ChartQA, InfoVQA, and OCRBench---further evaluate fine-grained textual perception within images, a capability critical for practical deployment of native models.

\paragraph{Generation Benchmarks.}
Image generation evaluation has evolved from distribution-level metrics to compositional, semantic-level assessment. Fr\'{e}chet Inception Distance (FID) measures distributional similarity but is insensitive to compositional accuracy. GenEval addresses this by decomposing text-to-image generation into attribute binding, spatial relationships, and counting sub-tasks. DPG-Bench evaluates dense prompt following with long, compositionally complex descriptions. T2I-CompBench provides multi-dimensional metrics covering attribute binding, object relationships, and complex composition. CLIPScore offers a reference-free text--image alignment metric, though its sensitivity to fine-grained generation quality is limited---\citet{rao2026understandinggenerationfightdiagnostic} show that DPO on VQ-based unified models fails to improve CLIPScore even when understanding metrics improve, revealing that discrete tokenization creates a structural bottleneck for offline preference optimization.

\begin{table*}[!t]
\centering
\scriptsize
\renewcommand{\arraystretch}{1.15}
\setlength{\tabcolsep}{3pt}

\definecolor{ImgBadge}{HTML}{2E86C1}
\definecolor{AudBadge}{HTML}{CA8A04}
\definecolor{VidBadge}{HTML}{16A34A}
\definecolor{ImgRow}{HTML}{F4F9FD}
\definecolor{AudRow}{HTML}{FBF5E6}
\definecolor{VidRow}{HTML}{EAF7EE}

\setlength{\fboxsep}{3pt}
\newcommand{\modbadge}[2]{%
  {\setlength{\fboxsep}{3.5pt}\colorbox{#1}{\textbf{\color{white}\scriptsize #2}}}%
}
\newcommand{\taskbadge}[2]{%
  {\setlength{\fboxsep}{3pt}\fcolorbox{#1}{#1!12}{\textbf{\color{#1}\scriptsize #2}}}%
}
\begin{tabularx}{\textwidth}{@{} l l l l Y @{}}
\toprule
\textbf{Modality} & \textbf{Task Group} & \textbf{Benchmark} & \textbf{Metric} & \textbf{Key Characteristics} \\
\midrule

\rowcolor{ImgRow}
\modbadge{ImgBadge}{Image} & \taskbadge{ImgBadge}{General Perception} & VQAv2~\cite{goyal2017vqav2}     & Acc.    & Open-ended VQA with balanced answer distribution. \\
\rowcolor{ImgRow}
&                                                  & GQA~\cite{hudson2019gqa}        & Acc.    & Compositional questions grounded on scene graphs. \\
\rowcolor{ImgRow}
&                                                  & SEED-Bench~\cite{li2024seedbench} & Acc.  & 12 evaluation dims across spatial \& temporal reasoning. \\
\rowcolor{ImgRow}
&                                                  & MMBench~\cite{liu2023mmbench}   & Acc.    & Bilingual multi-choice with circular evaluation. \\
\rowcolor{ImgRow}
&                                                  & MMStar~\cite{chen2024mmstar}    & Acc.    & Vision-indispensable, leakage-controlled selection. \\
\addlinespace[2pt]
\rowcolor{ImgRow}
& \taskbadge{ImgBadge}{Knowledge Reasoning}        & MMMU~\cite{yue2024mmmu}         & Acc.    & College-level reasoning over 30 disciplines. \\
\rowcolor{ImgRow}
&                                                  & MathVista~\cite{lu2023mathvista} & Acc.   & Mathematical reasoning grounded in visual contexts. \\
\addlinespace[2pt]
\rowcolor{ImgRow}
& \taskbadge{ImgBadge}{Hallucination}              & POPE~\cite{li2023evaluatingobjecthallucinationlarge} & F1 & Polling-based binary probing of object hallucination. \\
\rowcolor{ImgRow}
&                                                  & RLHF-V~\cite{yu2024rlhfvtrustworthymllmsbehavior}  & Hall. Score & Segment-level fine-grained hallucination evaluation. \\
\addlinespace[2pt]
\rowcolor{ImgRow}
& \taskbadge{ImgBadge}{Document \& OCR}            & DocVQA~\cite{mathew2021docvqa}      & ANLS  & Question answering on document images. \\
\rowcolor{ImgRow}
&                                                  & ChartQA~\cite{masry2022chartqa}     & Acc.  & Visual and logical reasoning over charts and plots. \\
\rowcolor{ImgRow}
&                                                  & InfoVQA~\cite{mathew2022infographicvqa} & ANLS & Multi-hop reasoning over infographic layouts. \\
\rowcolor{ImgRow}
&                                                  & OCRBench~\cite{liu2024ocrbench}     & Acc.  & Comprehensive OCR perception across 29 sub-tasks. \\
\addlinespace[2pt]
\rowcolor{ImgRow}
& \taskbadge{ImgBadge}{Generation}                 & GenEval~\cite{ghosh2024geneval}      & Comp. Score & T2I: attribute binding, counting, relations. \\
\rowcolor{ImgRow}
&                                                  & DPG-Bench~\cite{hu2024dpgbench}      & Alignment & Dense, long-prompt following with structured grading. \\
\rowcolor{ImgRow}
&                                                  & T2I-CompBench~\cite{huang2023t2icompbench} & Multi & Attribute binding, relations, complex composition. \\
\rowcolor{ImgRow}
&                                                  & FID~\cite{heusel2017gans}            & Distrib.    & Fr\'{e}chet Inception Distance to real-image distribution. \\
\rowcolor{ImgRow}
&                                                  & CLIPScore~\cite{hessel2021clipscore} & Alignment   & CLIP-embedded Reference-free text-image alignment. \\
\midrule

\rowcolor{AudRow}
\modbadge{AudBadge}{Audio} & \taskbadge{AudBadge}{Speech Recognition} & LibriSpeech~\cite{panayotov2015librispeech} & WER & Read English speech, clean and other splits. \\
\rowcolor{AudRow}
&                                                  & CommonVoice~\cite{ardila2020common}  & WER   & Crowdsourced multilingual ASR across diverse accents. \\
\rowcolor{AudRow}
&                                                  & FLEURS~\cite{conneau2023fleurs}      & WER   & Few-shot ASR across 102 languages. \\
\addlinespace[2pt]
\rowcolor{AudRow}
& \taskbadge{AudBadge}{Speech Synthesis}           & MOS-Bench~\cite{huang2026mos}        & MOS   & Subjective rating of naturalness and prosody. \\
\addlinespace[2pt]
\rowcolor{AudRow}
& \taskbadge{AudBadge}{Full-Duplex Interaction}    & Moshi Eval~\cite{defossez2024moshispeechtextfoundationmodel} & Latency & Real-time full-duplex with 200\,ms target latency. \\
\rowcolor{AudRow}
&                                                  & SoulX-Duplug-Eval~\cite{yan2026soulx} & Lat./Acc. & Bilingual streaming turn detection at 240\,ms latency. \\
\rowcolor{AudRow}
&                                                  & Full-Duplex-Bench~\cite{lin2025full} & Multi & Turn-taking, barge-in handling, false-interruption rate. \\
\midrule
\rowcolor{VidRow}
\modbadge{VidBadge}{Video} & \taskbadge{VidBadge}{Offline Understanding} & VideoMME~\cite{fu2025video} & Acc. & General video QA spanning short to long durations. \\
\rowcolor{VidRow}
&                                                  & EgoSchema~\cite{mangalam2023egoschema} & Acc. & Long-form egocentric video QA w/ temporal reasoning. \\
\rowcolor{VidRow}
&                                                  & MVBench~\cite{li2024mvbench}         & Acc.  & 20 fine-grained temporal tasks. \\
\rowcolor{VidRow}
&                                                  & PerceptionTest~\cite{patraucean2023perception} & Acc. & Multimodal perception and causal-reasoning skill probe. \\
\rowcolor{VidRow}
&                                                  & LongVideoBench~\cite{wu2024longvideobench} & Acc. & Hour-long referring and reasoning over long contexts. \\
\rowcolor{VidRow}
&                                                  & MLVU~\cite{zhou2025mlvu}             & Acc.  & Multi-task long-video understanding. \\
\addlinespace[2pt]
\rowcolor{VidRow}
& \taskbadge{VidBadge}{Streaming Understanding}    & OVO-Bench~\cite{niu2025ovo}          & Multi & Online perception with backward tracing of past events. \\
\rowcolor{VidRow}
&                                                  & StreamingBench~\cite{lin2026streamingbench} & Acc./Lat. & Video comprehension under latency constraints. \\
\rowcolor{VidRow}
&                                                  & OmniMMI~\cite{wang2025omnimmi}       & Multi & Multimodal streaming interaction evaluation. \\
\addlinespace[2pt]
\rowcolor{VidRow}
& \taskbadge{VidBadge}{Generation}                 & UCF-101~\cite{soomro2012ucf101}      & FVD   & Action-class video generation distributional metric. \\
\rowcolor{VidRow}
&                                                  & Kinetics-600~\cite{carreira2018short} & FVD  & Large-scale action distribution for video FVD. \\
\rowcolor{VidRow}
&                                                  & VBench~\cite{huang2023vbench}        & Multi & Temporal consistency, motion smoothness, aesthetics. \\
\rowcolor{VidRow}
&                                                  & SeedVideoBench~2.0~\cite{seedance2026seedance} & 6-dim & Motion, prompt adherence, A/V sync. \\
\rowcolor{VidRow}
&                                                  & Arena.AI~\cite{arenaai}              & Elo   & Community-scale human-preference Elo ranking. \\
\bottomrule
\end{tabularx}
\caption{Summary of major evaluation benchmarks for native multimodal models. Each benchmark is shown on its own row to preserve its unique characteristics; the modality and task badges (\modbadge{ImgBadge}{Image}, \modbadge{AudBadge}{Audio}, \modbadge{VidBadge}{Video}) act as visual anchors and apply to all rows below until the next badge.}
\label{tab:eval_benchmarks}
\end{table*}

\subsection{Audio}
\label{sec:eval_audio}

Audio evaluation for NMM systems spans three capability axes: speech recognition, speech synthesis, and full-duplex interactive conversation.

\paragraph{Speech Understanding.}
Automatic Speech Recognition (ASR) is evaluated by Word Error Rate (WER) on standard corpora including LibriSpeech, CommonVoice, and FLEURS. For native omni models such as Qwen3-Omni and Ming-Flash-Omni, ASR benchmarks serve as regression tests to verify that multimodal integration does not degrade core speech perception. Beyond transcription, audio understanding benchmarks increasingly assess semantic comprehension of paralinguistic cues---emotion recognition, speaker identification, and environmental sound classification---capabilities that are critical for models processing raw audio natively rather than through an ASR cascade.

\paragraph{Speech Generation.}
Text-to-Speech (TTS) quality is predominantly evaluated through Mean Opinion Score (MOS), a subjective rating capturing naturalness, prosody, and speaker similarity. For native models with speech output, additional metrics include first-token latency, word-level synchronization accuracy, and voice cloning fidelity. Low-latency streaming TTS has been demonstrated within autoregressive multimodal frameworks by SyncSpeech~\cite{sheng2025syncspeech} and AR-Omni~\cite{cheng2026ar}. SeedVideoBench~2.0 further extends audio evaluation to three dimensions---audio expressiveness, audio-visual synchronization, and audio prompt adherence, establishing a more comprehensive protocol for joint audio-video generation.

\paragraph{Full-Duplex Interaction.}
A core capability of native audio models is full-duplex conversation: simultaneous listening and speaking with natural turn-taking. Evaluation here beyond traditional ASR/TTS: \textit{turn-taking accuracy} (predicting when to speak), \textit{barge-in handling} (gracefully yielding when interrupted), \textit{response latency} (time from user silence to system response), and \textit{false interruption rate}. Moshi pioneered real-time full-duplex evaluation with 200ms latency targets. SoulX-Duplug introduces a bilingual evaluation suite achieving 240ms average streaming turn detection latency. FireRedChat~\cite{chen2025fireredchat} and ELLSA~\cite{wang2025end} evaluate cascaded and end-to-end full-duplex implementations respectively, while LLM-enhanced dialogue management~\cite{zhang2025llm} tests LLM-based approaches to turn prediction. These benchmarks capture the real-time interaction quality that distinguishes native audio models from traditional pipeline systems.

\subsection{Video}
\label{sec:eval_video}

\paragraph{Understanding Benchmarks.}
Video understanding evaluation tests both offline comprehension and real-time streaming capabilities, reflecting the dual deployment modes of native multimodal models.

For \textit{offline video understanding}, benchmarks assess progressively harder temporal reasoning. VideoMME and EgoSchema evaluate general video QA across durations from seconds to hours. MVBench and PerceptionTest probe fine-grained temporal perception including action sequencing, state changes, and causal reasoning. Long-video benchmarks such as LongVideoBench and MLVU specifically target the long-range dependency challenges addressed by native models with extended context. Qwen3-VL supports 256K tokens and FAR~\cite{gu2025long} explores long-context autoregressive video modeling. Kimi K2.5 introduces an agent-swarm paradigm for distributed video analysis, achieving 4.5$\times$ processing efficiency gains on long-video tasks.

\textit{Streaming video understanding} represents a frontier evaluation paradigm uniquely suited to native models. OVO-Bench evaluates real-time visual perception (OCR, action recognition, spatial understanding) alongside backward tracing capabilities, testing whether models can identify the appropriate temporal moment to respond. StreamingBench tests continuous video comprehension under strict latency constraints. ThinkStream~\cite{liu2026thinking} introduces the Watch--Think--Speak protocol, judging models not only on answer accuracy but on \textit{response timing}---whether sufficient evidence has accumulated before responding. AURA~\cite{lu2026aura} further extends streaming evaluation to proactive QA (responding when relevant events occur without explicit queries) and multi-response QA (tracking evolving events over time), requiring joint assessment of response quality and temporal appropriateness.

Complementing accuracy-focused evaluation, recent work demonstrates the need for \textit{efficiency-aware} protocols. ResAdapt~\cite{liao2026resadapt} shows that adaptive input-side visual budget allocation can eliminate over 90\% of visual tokens while processing 16$\times$ more frames, achieving $>$15\% relative gains on complex long-video reasoning. This highlights an emerging evaluation dimension that jointly assesses accuracy and computational cost.

\paragraph{Generation Benchmarks.}
Video generation evaluation encompasses both automated distributional metrics and human preference assessment. Fr\'{e}chet Video Distance (FVD) on UCF-101 and Kinetics-600 remains the standard distributional metric. VBench provides comprehensive multi-dimensional evaluation covering temporal consistency, motion smoothness, subject identity preservation, and aesthetic quality. HunyuanVideo-1.5 reports on these standard benchmarks, assessing motion quality, visual fidelity, and text-video alignment.

For native multimodal video generators, evaluation increasingly emphasizes \textit{controllability} and \textit{multimodal conditioning}. Seedance 2.0 establishes SeedVideoBench~2.0 with six evaluation dimensions---motion quality, video prompt adherence, aesthetics, audio quality, audio-visual synchronization, and audio prompt following---assessed across text-to-video, image-to-video, and reference-to-video tasks. Notably, it achieves top rankings on both Arena.AI T2V and I2V leaderboards, providing community-scale human preference validation. Next Block Prediction~\cite{ren2025next} introduces evaluation protocols for semi-autoregressive video generation, where spatial-temporal coherence of block-level predictions must be explicitly assessed. LTX-2~\cite{LightricksLTX2_2026} evaluates joint audio-visual generation quality, testing temporal synchronization between synthesized audio and video. These emerging benchmarks reflect the expanding scope of native multimodal generation beyond traditional text-to-video metrics.

%% file: future.tex
\section{Future Outlook}
\label{sec:future}

The roadmap presented in the previous sections paints a clear arc: NMM systems has progressed from \textit{modular assembly} of frozen encoders, through \textit{mid-fusion} backbones with explicit modality boundaries, toward an emerging \textit{early-fusion} regime in which understanding and generation co-exist within a single transformer space. While this trajectory is increasingly well-defined at the architectural level, our investigation across \S\ref{arch}--\S\ref{eval} reveals that translating it into deployable, industrial-grade systems remains an open frontier. In this section, we synthesize the open problems and outline what we view as the most consequential research directions toward truly born-native world models.

\subsection{Toward Architectural Convergence: From M2T/M2G to Symmetric M2M}

The taxonomy in \S\ref{arch}, organized through the lens of input--output duality, shows that the field is still divided across three regimes: M2T unimodal generation, M2G scenario-based generation, and M2M symmetric modeling. We expect this fragmentation to gradually collapse, and we identify three convergence axes that warrant systematic investigation.

\paragraph{Unifying understanding and generation in a single backbone.} Most current ``unified'' models still rely on hybrid objectives---next-token prediction for textual reasoning paired with diffusion or flow-matching heads for visual/audio synthesis, e.g., Transfusion~\cite{zhou2024transfusion}, Show-o2~\cite{xie2025showo2improvednativeunified}, and BAGEL~\cite{BAGEL7B2025}. A central open question is whether a \textit{single} probabilistic objective, a unified tokenization scheme, or a continuous latent grammar can support both fronts without quality regression on either side. Bridging discrete-token unification (e.g., Chameleon~\cite{team2024chameleon}, AnyGPT~\cite{zhan2024anygpt}, Janus-Pro~\cite{DeepSeekJanusPro2025}) and continuous-latent paths (e.g., TUNA-2~\cite{liu2026tuna2pixelembeddingsbeat}, Mamoda2.5~\cite{shi2026mamoda25enhancingunifiedmultimodal}) remains an unresolved design choice, with strong implications for scaling laws and downstream controllability.

\paragraph{Scaling sparsity and modality-aware experts.} As Table~\ref{tab:multimodal_evolution} shows, flagship NMMs are increasingly MoEs at the trillion-parameter scale (e.g., Kimi K2.5~\cite{KimiK2_5_2026}, GLM-5V-Turbo~\cite{GLM5VTurbo2026}, Ming-Flash-Omni-2.0~\cite{ai2026mingflashomnisparseunifiedarchitecture}). Yet, modality-aware routing, expert specialization across vision/audio/video, and the interplay between sparsity and cross-modal attention remain poorly understood. Future work should formalize \textit{expert nativity}---the degree to which experts are jointly trained across modalities versus specialized---as a counterpart to the architectural nativity defined in \S\ref{sec:formal}.

\paragraph{Beyond the four canonical modalities.} The M2M paradigm should ultimately extend beyond text/image/audio/video to embodied signals, including proprioception, depth, tactile, action sequences, and structured modalities such as code, graphs, and 3D scenes. We anticipate that the formal definition of nativity introduced in \S\ref{sec:formal} will need to be generalized to such heterogeneous, possibly continuous-time signals.

\subsection{Data: From Curated Corpora to Self-Generating Multimodal Streams}

The data curriculum surveyed in \S\ref{data} already organizes corpora by understanding-, generation-, interaction-, and preference-oriented purposes. Three open problems stand out.

\paragraph{Cross-modal data scarcity and synthesis.} Aligned multi-stream data---particularly long-horizon video paired with synchronized audio, transcripts, actions, and reasoning traces---remains the hardest bottleneck. Synthetic data generated by NMM systems themselves is becoming feasible, yet rigorous methodologies for \textit{filtering}, \textit{de-biasing}, and \textit{preventing model collapse} in self-distilled multimodal pipelines are still missing.

\paragraph{Interaction-grounded data at scale.} Full-duplex audio dialogue (Moshi~\cite{defossez2024moshispeechtextfoundationmodel}), streaming video (ThinkStream~\cite{liu2026thinking}, AURA~\cite{lu2026aura}), and proactive agent traces require data that captures not only \textit{what} to respond, but also \textit{when} to respond. Curating such temporally annotated corpora, ideally through scalable instrumentation of real deployments rather than offline labeling, is a prerequisite for the next generation of native interactive systems.

\paragraph{Preference data for generative modalities.} While preference data for text is mature, scalable preference signals for image/audio/video generation, such as aesthetics, factuality, and audio-visual synchronization, remain comparatively under-developed. We expect cross-modal reward modeling, jointly trained with the policy, to become a central data-engineering effort.

\subsection{Training: Joint PT/SFT/RL/OPD Recipes for Native Models}

The training stack surveyed in \S\ref{sec:training}, spanning pre-training, supervised fine-tuning, reinforcement learning, and on-policy distillation, was largely inherited from text-only LLMs. Native models impose new demands.

\paragraph{Modality-balanced optimization.} Mixing tokens of vastly different information density (e.g., a 32K-token long-document SFT sample vs. a sequence-packed image grid) creates loss-scale and gradient-norm asymmetry. Principled \textit{token-budget allocation}, \textit{per-modality loss weighting}, and \textit{curriculum scheduling} across the joint corpus are still underexplored. The continued reliance on heuristic mixture ratios (\S\ref{data}) suggests an opportunity for theoretically grounded, modality-aware training laws.

\paragraph{RL for cross-modal generation.} RLHF and RLVR for text are well established, but extending verifiable rewards to image, audio, video generation, and interleaved interaction traces remains open. We expect a unification of policy-gradient methods with diffusion/flow-based generative objectives, possibly via \textit{stepwise multimodal advantage estimation}, to be a central technical thrust.

\paragraph{On-policy distillation for omni capabilities.} OPD has emerged as an efficient lever to transfer capabilities from large teachers to small students. For NMM, beyond the M2T projection, distilling \textit{symmetric} M2M behavior into compact deployable models is largely uncharted, especially under streaming and full-duplex constraints.

\subsection{Inference and Deployment: Streaming, Long-Context, and System Co-Design}

\S\ref{inference} surfaces three orthogonal pressures on NMM serving: \textit{sequence explosion} from long video and document inputs, \textit{heterogeneity and scale} from MoE-augmented multimodal stacks, and \textit{real-time streaming with full-duplex interaction}. We highlight three forward-looking directions.

\paragraph{Native long-context and adaptive perception.} Beyond model-level remedies such as 256K context windows and long-context autoregressive video modeling (e.g., FAR~\cite{gu2025long}), \textit{adaptive perception} will be essential, which refers to selectively spending compute on informative regions of the input stream. Recent results such as ResAdapt~\cite{liao2026resadapt} eliminating $>90\%$ of visual tokens while expanding the temporal horizon $16\times$ point toward an emerging \textit{accuracy-efficiency Pareto frontier} that benchmarks must explicitly assess.

\paragraph{System--algorithm co-design for sparse multimodal MoE.} Disaggregated prefill/decoding, expert offloading, and modality-aware KV-cache management are becoming first-class concerns. The interaction between MoE sparsity, sparse attention, and multimodal sequence packing opens a rich space of co-design problems that span the kernel layer up to the scheduling layer.

\paragraph{Born-streaming, born-duplex deployment.} Truly native interactive agents require streaming \textit{by construction}, not as a post-hoc wrapper around an autoregressive backbone. End-to-end full-duplex frameworks (Moshi~\cite{defossez2024moshispeechtextfoundationmodel}, ELLSA~\cite{wang2025end}, FireRedChat~\cite{chen2025fireredchat}) and Watch--Think--Speak protocols (ThinkStream~\cite{liu2026thinking}) hint at this future, but stable, deployable, low-latency systems with consistent quality across modalities are still an industrial open problem.

\subsection{Evaluation: From Static Benchmarks to Holistic, Temporally-Aware Protocols}

The benchmarks summarized in \S\ref{eval} reveal two systemic gaps. First, most existing benchmarks evaluate modalities in isolation; few jointly assess understanding \textit{and} generation, or cross-modal grounding under interaction. Second, accuracy-only metrics ignore the dimensions that matter most for native deployment: \textit{when} a model responds, \textit{how much compute} it consumes, and how gracefully it handles \textit{streaming} and \textit{interruption}. We see four open directions:
\begin{itemize}[leftmargin=1.5em, itemsep=2pt, topsep=2pt]
    \item \textbf{Symmetric M2M benchmarks} that grade a single model on aligned understanding--generation pairs (e.g., describe-then-render, listen-then-speak, watch-then-act), penalizing inconsistency across the two directions.
    \item \textbf{Temporally-aware metrics}, generalizing the Watch--Think--Speak protocol of ThinkStream~\cite{liu2026thinking} and the proactive QA setup of AURA~\cite{lu2026aura}, that jointly score answer quality and response timing.
    \item \textbf{Efficiency-aware protocols} that report accuracy alongside token budget, latency, and energy, in the spirit of ResAdapt~\cite{liao2026resadapt}, so that the community can compare native models on a meaningful Pareto frontier.
    \item \textbf{Robustness and safety under multimodal attack surfaces}, including adversarial cross-modal prompts, jailbreaks via images/audio, and hallucination of generated content, all of which are only partially covered by current single-modality safety benchmarks.
\end{itemize}

\subsection{Toward Native World Models}

Looking further ahead, we expect NMM to evolve beyond a system that consumes and produces multimodal tokens into a genuine \textit{world model}: a unified backbone that perceives raw sensory streams, maintains persistent state across long horizons, and acts in continuous time. The roadmap from late-fusion stitching to early-fusion convergence is increasingly clear at the architectural level, but the path to deployable, born-native world models is not. We hope the formalization, taxonomy, and open problems consolidated in this paper provide the community with a structured starting point for the next phase of the journey: a unified, symmetric, streaming, and embodied multimodal intelligence.

%% file: nmm.bbl
\begin{thebibliography}{297}
\providecommand{\natexlab}[1]{#1}
\providecommand{\url}[1]{\texttt{#1}}
\expandafter\ifx\csname urlstyle\endcsname\relax
  \providecommand{\doi}[1]{doi: #1}\else
  \providecommand{\doi}{doi: \begingroup \urlstyle{rm}\Url}\fi

\bibitem[Lu et~al.(2025)Lu, Qin, Qiao, Li, Dai, Ke, He, Qiao, Yin, Sun, et~al.]{lu2025youtu}
Junru Lu, Jiarui Qin, Lingfeng Qiao, Yinghui Li, Xinyi Dai, Bo~Ke, Jianfeng He, Ruizhi Qiao, Di~Yin, Xing Sun, et~al.
\newblock Youtu-llm: Unlocking the native agentic potential for lightweight large language models.
\newblock \emph{arXiv preprint arXiv:2512.24618}, 2025.

\bibitem[Dong et~al.(2024{\natexlab{a}})Dong, Hong, Bei, Huang, Wang, and Huang]{dong2024clrbenchevaluatinglargelanguage}
Junnan Dong, Zijin Hong, Yuanchen Bei, Feiran Huang, Xinrun Wang, and Xiao Huang.
\newblock Clr-bench: Evaluating large language models in college-level reasoning.
\newblock \emph{arXiv preprint arXiv:2410.17558}, 2024{\natexlab{a}}.

\bibitem[Liu et~al.(2024{\natexlab{a}})Liu, Feng, Xue, Wang, Wu, Lu, Zhao, Deng, Zhang, Ruan, et~al.]{liu2024deepseek}
Aixin Liu, Bei Feng, Bing Xue, Bingxuan Wang, Bochao Wu, Chengda Lu, Chenggang Zhao, Chengqi Deng, Chenyu Zhang, Chong Ruan, et~al.
\newblock Deepseek-v3 technical report.
\newblock \emph{arXiv preprint arXiv:2412.19437}, 2024{\natexlab{a}}.

\bibitem[Bai et~al.(2023)Bai, Bai, Chu, Cui, Dang, Deng, Fan, Ge, Han, Huang, et~al.]{bai2023qwen}
Jinze Bai, Shuai Bai, Yunfei Chu, Zeyu Cui, Kai Dang, Xiaodong Deng, Yang Fan, Wenbin Ge, Yu~Han, Fei Huang, et~al.
\newblock Qwen technical report.
\newblock \emph{arXiv preprint arXiv:2309.16609}, 2023.

\bibitem[{Qwen Team}(2025{\natexlab{a}})]{bai2025qwen3vltechnicalreport}
{Qwen Team}.
\newblock Qwen3-vl technical report.
\newblock \emph{arXiv preprint arXiv:2511.21631}, 2025{\natexlab{a}}.

\bibitem[Tong et~al.(2026)Tong, Fan, Nguyen, Brown, Zhou, Qian, Zheng, Vallaeys, Han, Fergus, et~al.]{tong2026beyond}
Shengbang Tong, David Fan, John Nguyen, Ellis Brown, Gaoyue Zhou, Shengyi Qian, Boyang Zheng, Th{\'e}ophane Vallaeys, Junlin Han, Rob Fergus, et~al.
\newblock Beyond language modeling: An exploration of multimodal pretraining.
\newblock \emph{arXiv preprint arXiv:2603.03276}, 2026.

\bibitem[Wang et~al.(2025{\natexlab{a}})Wang, Gao, Gu, Pu, Cui, Wei, Liu, Jing, Ye, Shao, et~al.]{InternVL3.5_2025}
Weiyun Wang, Zhangwei Gao, Lixin Gu, Hengjun Pu, Long Cui, Xingguang Wei, Zhaoyang Liu, Linglin Jing, Shenglong Ye, Jie Shao, et~al.
\newblock Internvl3. 5: Advancing open-source multimodal models in versatility, reasoning, and efficiency.
\newblock \emph{arXiv preprint arXiv:2508.18265}, 2025{\natexlab{a}}.

\bibitem[Caffagni et~al.(2024)Caffagni, Cocchi, Barsellotti, Moratelli, Sarto, Baraldi, Cornia, and Cucchiara]{caffagni2024revolution}
Davide Caffagni, Federico Cocchi, Luca Barsellotti, Nicholas Moratelli, Sara Sarto, Lorenzo Baraldi, Marcella Cornia, and Rita Cucchiara.
\newblock The revolution of multimodal large language models: A survey.
\newblock \emph{Findings of the association for computational linguistics: ACL 2024}, pages 13590--13618, 2024.

\bibitem[Yin et~al.(2023)Yin, Fu, Zhao, Li, Sun, Xu, and Chen]{Yin_2024_survey}
Shukang Yin, Chaoyou Fu, Sirui Zhao, Ke~Li, Xing Sun, Tong Xu, and Enhong Chen.
\newblock A survey on multimodal large language models.
\newblock \emph{National Science Review}, 11, 2023.

\bibitem[Dong et~al.(2024{\natexlab{b}})Dong, Zhang, Zhou, Zha, Zheng, and Huang]{dong2024modality}
Junnan Dong, Qinggang Zhang, Huachi Zhou, Daochen Zha, Pai Zheng, and Xiao Huang.
\newblock Modality-aware integration with large language models for knowledge-based visual question answering.
\newblock In \emph{ACL}, pages 2417--2429, 2024{\natexlab{b}}.

\bibitem[Zhao et~al.(2025)Zhao, Zhang, Guo, Hu, Duan, Fu, Chng, Wang, Chen, Xu, et~al.]{zhao2025unified}
Shanshan Zhao, Xinjie Zhang, Jintao Guo, Jiakui Hu, Lunhao Duan, Minghao Fu, Yong~Xien Chng, Guo-Hua Wang, Qing-Guo Chen, Zhao Xu, et~al.
\newblock Unified multimodal understanding and generation models: Advances, challenges, and opportunities.
\newblock \emph{arXiv preprint arXiv:2505.02567}, 2025.

\bibitem[Cui et~al.(2025)Cui, Chen, Deng, Huang, Li, Liu, Liu, Luo, Wang, Wang, Wang, Wang, Zhang, Zhao, Pan, Li, Hao, Ma, Chen, Ao, Huang, Wang, and Wang]{cui2025emu35nativemultimodalmodels}
Yufeng Cui, Honghao Chen, Haoge Deng, Xu~Huang, Xinghang Li, Jirong Liu, Yang Liu, Zhuoyan Luo, Jinsheng Wang, Wenxuan Wang, Yueze Wang, Chengyuan Wang, Fan Zhang, Yingli Zhao, Ting Pan, Xianduo Li, Zecheng Hao, Wenxuan Ma, Zhuo Chen, Yulong Ao, Tiejun Huang, Zhongyuan Wang, and Xinlong Wang.
\newblock Emu3.5: Native multimodal models are world learners.
\newblock \emph{arXiv preprint arXiv:2510.26583}, 2025.

\bibitem[Zhang et~al.(2024{\natexlab{a}})Zhang, Wu, Li, Li, Ma, Liu, and Li]{zhang2024llava}
Yuanhan Zhang, Jinming Wu, Wei Li, Bo~Li, Zejun Ma, Ziwei Liu, and Chunyuan Li.
\newblock Llava-video: Video instruction tuning with synthetic data.
\newblock \emph{arXiv preprint arXiv:2410.02713}, 2024{\natexlab{a}}.

\bibitem[Lu et~al.(2024)Lu, Liu, Zhang, Wang, Dong, Liu, Sun, Ren, Li, Yang, et~al.]{lu2024deepseekvl}
Haoyu Lu, Wen Liu, Bo~Zhang, Bingxuan Wang, Kai Dong, Bo~Liu, Jingxiang Sun, Tongzheng Ren, Zhuoshu Li, Hao Yang, et~al.
\newblock Deepseek-vl: towards real-world vision-language understanding.
\newblock \emph{arXiv preprint arXiv:2403.05525}, 2024.

\bibitem[Wu et~al.(2025{\natexlab{a}})Wu, Li, Zhou, Lin, Gao, Yan, Yin, Bai, Xu, Chen, et~al.]{wu2025qwenimage}
Chenfei Wu, Jiahao Li, Jingren Zhou, Junyang Lin, Kaiyuan Gao, Kun Yan, Sheng-ming Yin, Shuai Bai, Xiao Xu, Yilei Chen, et~al.
\newblock Qwen-image technical report.
\newblock \emph{arXiv preprint arXiv:2508.02324}, 2025{\natexlab{a}}.

\bibitem[Yu et~al.(2025{\natexlab{a}})Yu, Wang, Wang, Huang, Ma, He, Cai, Chen, Huang, Zhao, et~al.]{yu2025minicpmv45cookingefficient}
Tianyu Yu, Zefan Wang, Chongyi Wang, Fuwei Huang, Wenshuo Ma, Zhihui He, Tianchi Cai, Weize Chen, Yuxiang Huang, Yuanqian Zhao, et~al.
\newblock Minicpm-v 4.5: Cooking efficient mllms via architecture, data, and training recipe.
\newblock \emph{arXiv preprint arXiv:2509.18154}, 2025{\natexlab{a}}.

\bibitem[Deshmukh et~al.(2026)Deshmukh, Chumachenko, Rintamaki, Le, Poon, Taheri, Karmanov, Liu, Seppanen, Goel, et~al.]{nvidia2026nemotron3nanoomni}
Amala~Sanjay Deshmukh, Kateryna Chumachenko, Tuomas Rintamaki, Matthieu Le, Tyler Poon, Danial~Mohseni Taheri, Ilia Karmanov, Guilin Liu, Jarno Seppanen, Arushi Goel, et~al.
\newblock Nemotron 3 nano omni: Efficient and open multimodal intelligence.
\newblock \emph{arXiv preprint arXiv:2604.24954}, 2026.

\bibitem[{Xiaomi MiMo Team}(2026)]{xiaomi2026mimov25}
{Xiaomi MiMo Team}.
\newblock Mimo-v2.5: Frontier agency and native multimodality, 2026.
\newblock Accessed: 2026-04-29.

\bibitem[{Qwen Team}(2026)]{qwen36_35b_a3b}
{Qwen Team}.
\newblock {Qwen3.6-35B-A3B}: Agentic coding power, now open to all, April 2026.
\newblock URL \url{https://qwen.ai/blog?id=qwen3.6-35b-a3b}.

\bibitem[{Google DeepMind}(2026)]{Gemma4Team2026}
{Google DeepMind}.
\newblock {Gemma 4}: Intelligence-per-parameter for advanced reasoning.
\newblock \url{https://ai.google.dev/gemma/docs/core/model\_card\_4}, 2026.

\bibitem[{Kimi Team} et~al.(2026){Kimi Team}, Bai, et~al.]{KimiK2_5_2026}
{Kimi Team}, Tongtong Bai, et~al.
\newblock {Kimi K2.5}: Visual agentic intelligence.
\newblock \emph{arXiv preprint arXiv:2602.02276}, 2026.

\bibitem[Hong et~al.(2026)Hong, Gu, Pan, Yang, Wang, Wang, Yue, Wang, Wang, Wang, et~al.]{GLM5VTurbo2026}
Wenyi Hong, Xiaotao Gu, Ziyang Pan, Zhen Yang, Yuting Wang, Yue Wang, Yuanchang Yue, Yu~Wang, Yanling Wang, Yan Wang, et~al.
\newblock Glm-5v-turbo: Toward a native foundation model for multimodal agents.
\newblock Technical report, ZhipuAI, 2026.

\bibitem[Adcock et~al.(2025)Adcock, Srivastava, Dubey, et~al.]{Adcock2026TheL4}
Aaron Adcock, Aayushi Srivastava, Abhimanyu Dubey, et~al.
\newblock The {Llama 4} herd: Architecture, training, evaluation, and deployment notes.
\newblock Technical report, Meta AI, 2025.
\newblock Includes Scout and Maverick series.

\bibitem[{Qwen Team}(2025{\natexlab{b}})]{bai2025qwen25vltechnicalreport}
{Qwen Team}.
\newblock Qwen2.5-vl technical report.
\newblock \emph{arXiv preprint arXiv:2502.13923}, 2025{\natexlab{b}}.

\bibitem[Wang et~al.(2023{\natexlab{a}})]{wang2023cogvlm}
Weihan Wang et~al.
\newblock {CogVLM}: Visual expert for pretrained language models.
\newblock \emph{arXiv preprint arXiv:2311.03079}, 2023{\natexlab{a}}.

\bibitem[Lin et~al.(2024)Lin, Ye, Zhu, Cui, Ning, Jin, and Yuan]{lin2023video}
Bin Lin, Yang Ye, Bin Zhu, Jiaxi Cui, Munan Ning, Peng Jin, and Li~Yuan.
\newblock Video-llava: Learning united visual representation by alignment before projection.
\newblock \emph{arXiv preprint arXiv:2311.10122}, 2024.

\bibitem[Chu et~al.(2023)Chu, Xu, Zhou, Yang, Zhang, Yan, Zhou, and Zhou]{chu2023qwenaudioadvancinguniversalaudio}
Yunfei Chu, Jin Xu, Xiaohuan Zhou, Qian Yang, Shiliang Zhang, Zhijie Yan, Chang Zhou, and Jingren Zhou.
\newblock Qwen-audio: Advancing universal audio understanding via unified large-scale audio-language models.
\newblock \emph{arXiv preprint arXiv:2311.07919}, 2023.

\bibitem[Cai et~al.(2026)Cai, Chen, Gao, Gong, Li, Mei, Pan, Peng, Qiu, Yao, Yu, Zhang, et~al.]{hidreamolimage}
Qi~Cai, Jingwen Chen, Chengmin Gao, Zijian Gong, Yehao Li, Tao Mei, Yingwei Pan, Yi~Peng, Zhaofan Qiu, Ting Yao, Kai Yu, Yiheng Zhang, et~al.
\newblock Hidream-o1-image: A natively unified image generative foundation model with pixel-level unified transformer.
\newblock \emph{arXiv preprint arXiv:2605.11061}, 2026.

\bibitem[Zhu et~al.(2026{\natexlab{a}})Zhu, Ye, Kang, Yao, Guo, Kuang, Han, Zhuang, Lin, and Povey]{zhu2026omnivoiceomnilingualzeroshottexttospeech}
Han Zhu, Lingxuan Ye, Wei Kang, Zengwei Yao, Liyong Guo, Fangjun Kuang, Zhifeng Han, Weiji Zhuang, Long Lin, and Daniel Povey.
\newblock Omnivoice: Towards omnilingual zero-shot text-to-speech with diffusion language models.
\newblock \emph{arXiv preprint arXiv:2604.00688}, 2026{\natexlab{a}}.

\bibitem[HaCohen et~al.(2026)HaCohen, Brazowski, Chiprut, Bitterman, Kvochko, Berkowitz, Shalem, Lifschitz, Moshe, Porat, Richardson, Shiran, Chachy, Chetboun, Finkelson, Kupchick, Zabari, Guetta, Kotler, Bibi, Gordon, Panet, Benita, Armon, Kulikov, Inger, Shiftan, Melumian, and Farbman]{LightricksLTX2_2026}
Yoav HaCohen, Benny Brazowski, Nisan Chiprut, Yaki Bitterman, Andrew Kvochko, Avishai Berkowitz, Daniel Shalem, Daphna Lifschitz, Dudu Moshe, Eitan Porat, Eitan Richardson, Guy Shiran, Itay Chachy, Jonathan Chetboun, Michael Finkelson, Michael Kupchick, Nir Zabari, Nitzan Guetta, Noa Kotler, Ofir Bibi, Ori Gordon, Poriya Panet, Roi Benita, Shahar Armon, Victor Kulikov, Yaron Inger, Yonatan Shiftan, Zeev Melumian, and Zeev Farbman.
\newblock Ltx-2: Efficient joint audio-visual foundation model.
\newblock \emph{arXiv preprint arXiv:2601.03233}, 2026.

\bibitem[{Inclusion AI}(2026{\natexlab{a}})]{ai2026mingflashomnisparseunifiedarchitecture}
{Inclusion AI}.
\newblock Ming-flash-omni: A sparse, unified architecture for multimodal perception and generation.
\newblock \emph{arXiv preprint arXiv:2510.24821}, 2026{\natexlab{a}}.

\bibitem[Cui et~al.(2026)Cui, Xu, Wang, Yu, Sun, Xu, Wang, He, Ma, Cai, et~al.]{cui2026minicpm}
Junbo Cui, Bokai Xu, Chongyi Wang, Tianyu Yu, Weiyue Sun, Yingjing Xu, Tianran Wang, Zhihui He, Wenshuo Ma, Tianchi Cai, et~al.
\newblock Minicpm-o 4.5: Towards real-time full-duplex omni-modal interaction.
\newblock \emph{arXiv preprint arXiv:2604.27393}, 2026.

\bibitem[{Kling Team}(2025)]{klingteam2025klingomnitechnicalreport}
{Kling Team}.
\newblock Kling-omni technical report.
\newblock \emph{arXiv preprint arXiv:2512.16776}, 2025.

\bibitem[Wu et~al.(2025{\natexlab{b}})Wu, Zou, Li, Huang, Yang, Tan, et~al.]{wu2025hunyuanvideo15technicalreport}
Bing Wu, Chang Zou, Changlin Li, Duojun Huang, Fang Yang, Hao Tan, et~al.
\newblock Hunyuanvideo 1.5 technical report.
\newblock \emph{arXiv preprint arXiv:2511.18870}, 2025{\natexlab{b}}.

\bibitem[Xu et~al.(2025{\natexlab{a}})Xu, Guo, Hu, Chu, Wang, He, Wang, Shi, He, Zhu, et~al.]{Qwen3Omni2025}
Jin Xu, Zhifang Guo, Hangrui Hu, Yunfei Chu, Xiong Wang, Jinzheng He, Yuxuan Wang, Xian Shi, Ting He, Xinfa Zhu, et~al.
\newblock Qwen3-omni technical report.
\newblock \emph{arXiv preprint arXiv:2509.17765}, 2025{\natexlab{a}}.

\bibitem[{Wan Team}(2025)]{wan22_2025}
{Wan Team}.
\newblock {Wan 2.2}: A high-fidelity video generation foundation model, 2025.

\bibitem[Gao et~al.(2025)Gao, Gong, Guo, Hou, Lai, Li, Li, Lian, Liao, Liu, Liu, Shi, Sun, Tian, Tian, Wang, Wang, Wang, Wang, Wang, Wu, Wu, Xia, Xiao, Zhai, Zhang, Zhang, Zhang, Zhao, Yang, and Huang]{gao2025seedream30technicalreport}
Yu~Gao, Lixue Gong, Qiushan Guo, Xiaoxia Hou, Zhichao Lai, Fanshi Li, Liang Li, Xiaochen Lian, Chao Liao, Liyang Liu, Wei Liu, Yichun Shi, Shiqi Sun, Yu~Tian, Zhi Tian, Peng Wang, Rui Wang, Xuanda Wang, Xun Wang, Ye~Wang, Guofeng Wu, Jie Wu, Xin Xia, Xuefeng Xiao, Zhonghua Zhai, Xinyu Zhang, Qi~Zhang, Yuwei Zhang, Shijia Zhao, Jianchao Yang, and Weilin Huang.
\newblock Seedream 3.0 technical report.
\newblock \emph{arXiv preprint arXiv:2504.11346}, 2025.

\bibitem[Fu et~al.(2026)Fu, Huang, Wu, Jiang, Huo, Li, Song, Ding, Guo, He, Fu, Mao, and Zhang]{fu2026lanceunifiedmultimodalmodeling}
Fengyi Fu, Mengqi Huang, Shaojin Wu, Yunsheng Jiang, Yufei Huo, Hao Li, Yinghang Song, Fei Ding, Jianzhu Guo, Qian He, Zheren Fu, Zhendong Mao, and Yongdong Zhang.
\newblock Lance: Unified multimodal modeling by multi-task synergy.
\newblock \emph{arXiv preprint arXiv:2605.18678}, 2026.

\bibitem[Shi et~al.(2026{\natexlab{a}})Shi, Zhu, Shen, Yu, Chen, Chen, Yang, Zhou, Cheng, Ma, Wu, Yan, Li, Zhang, Li, Liu, and Sun]{shi2026mamoda25enhancingunifiedmultimodal}
Yangming Shi, Shixiang Zhu, Tao Shen, Zhimiao Yu, Dengsheng Chen, Taicai Chen, Yunfei Yang, Juan Zhou, Chen Cheng, Liang Ma, Xibin Wu, Benxuan Yan, Ge~Li, Tuoyu Zhang, Dan Li, Chang Liu, and Zhenbang Sun.
\newblock Mamoda2.5: Enhancing unified multimodal model with dit-moe.
\newblock \emph{arXiv preprint arXiv:2605.02641}, 2026{\natexlab{a}}.

\bibitem[Liu et~al.(2026{\natexlab{a}})Liu, Ren, Huang, Chen, Li, Chen, Ji, He, Schult, Zeng, Xiang, Chen, Luo, Zettlemoyer, and Cong]{liu2026tuna2pixelembeddingsbeat}
Zhiheng Liu, Weiming Ren, Xiaoke Huang, Shoufa Chen, Tianhong Li, Mengzhao Chen, Yatai Ji, Sen He, Jonas Schult, Belinda Zeng, Tao Xiang, Wenhu Chen, Ping Luo, Luke Zettlemoyer, and Yuren Cong.
\newblock Tuna-2: Pixel embeddings beat vision encoders for multimodal understanding and generation.
\newblock \emph{arXiv preprint arXiv:2604.24763}, 2026{\natexlab{a}}.

\bibitem[Diao et~al.(2026)Diao, Wu, Deng, Wang, Bai, Wu, Fan, Ye, Tong, Fan, et~al.]{diao2026sensenovau1unifyingmultimodalunderstanding}
Haiwen Diao, Penghao Wu, Hanming Deng, Jiahao Wang, Shihao Bai, Silei Wu, Weichen Fan, Wenjie Ye, Wenwen Tong, Xiangyu Fan, et~al.
\newblock Sensenova-u1: Unifying multimodal understanding and generation with neo-unify architecture.
\newblock \emph{arXiv preprint arXiv:2605.12500}, 2026.

\bibitem[{Inclusion AI}(2026{\natexlab{b}})]{ai2026llada20uniunifyingmultimodalunderstanding}
{Inclusion AI}.
\newblock Llada2.0-uni: Unifying multimodal understanding and generation with diffusion large language model.
\newblock \emph{arXiv preprint arXiv:2604.20796}, 2026{\natexlab{b}}.

\bibitem[Team et~al.(2026)Team, Xiao, Wang, Li, Zhang, Peng, Yu, Yang, Yan, Sun, et~al.]{MeituanLongCat2026}
Meituan~LongCat Team, Bin Xiao, Chao Wang, Chengjiang Li, Chi Zhang, Chong Peng, Hang Yu, Hao Yang, Haonan Yan, Haoze Sun, et~al.
\newblock Longcat-next: Lexicalizing modalities as discrete tokens.
\newblock \emph{arXiv preprint arXiv:2603.27538}, 2026.

\bibitem[Xie et~al.(2025{\natexlab{a}})Xie, Yang, and Shou]{xie2025showo2improvednativeunified}
Jinheng Xie, Zhenheng Yang, and Mike~Zheng Shou.
\newblock Show-o2: Improved native unified multimodal models.
\newblock \emph{arXiv preprint arXiv:2506.15564}, 2025{\natexlab{a}}.

\bibitem[Deng et~al.(2025)Deng, Zhu, Li, Gou, Li, Wang, Zhong, Yu, Nie, Song, Shi, and Fan]{BAGEL7B2025}
Chaorui Deng, Deyao Zhu, Kunchang Li, Chenhui Gou, Feng Li, Zeyu Wang, Shu Zhong, Weihao Yu, Xiaonan Nie, Ziang Song, Guang Shi, and Haoqi Fan.
\newblock Emerging properties in unified multimodal pretraining.
\newblock \emph{arXiv preprint arXiv:2505.14683}, 2025.

\bibitem[Li et~al.(2025{\natexlab{a}})Li, Peng, Wang, Peng, Chen, Weng, Wang, Cai, Dai, and Xiong]{OneCAT3B2025}
Han Li, Xinyu Peng, Yaoming Wang, Zelin Peng, Xin Chen, Rongxiang Weng, Jingang Wang, Xunliang Cai, Wenrui Dai, and Hongkai Xiong.
\newblock Onecat: Decoder-only auto-regressive model for unified understanding and generation.
\newblock \emph{arXiv preprint arXiv:2509.03498}, 2025{\natexlab{a}}.

\bibitem[{DeepSeek-AI}(2025)]{DeepSeekJanusPro2025}
{DeepSeek-AI}.
\newblock {Janus-Pro}: Unified multimodal understanding and generation with data and model scaling.
\newblock \emph{arXiv preprint arXiv:2501.17811}, 2025.

\bibitem[Défossez et~al.(2024)Défossez, Mazaré, Orsini, Royer, Pérez, Jégou, Grave, and Zeghidour]{defossez2024moshispeechtextfoundationmodel}
Alexandre Défossez, Laurent Mazaré, Manu Orsini, Amélie Royer, Patrick Pérez, Hervé Jégou, Edouard Grave, and Neil Zeghidour.
\newblock Moshi: a speech-text foundation model for real-time dialogue.
\newblock \emph{arXiv preprint arXiv:2410.00037}, 2024.

\bibitem[Zhou et~al.(2024{\natexlab{a}})Zhou, Yu, Babu, Tirumala, Yasunaga, Shamis, Kahn, Ma, Zettlemoyer, and Levy]{zhou2024transfusion}
Chunting Zhou, Lili Yu, Arun Babu, Kushal Tirumala, Michihiro Yasunaga, Leonid Shamis, Jacob Kahn, Xuezhe Ma, Luke Zettlemoyer, and Omer Levy.
\newblock Transfusion: Predict the next token and diffuse images with one multi-modal model.
\newblock \emph{arXiv preprint arXiv:2408.11039}, 2024{\natexlab{a}}.

\bibitem[{Chameleon Team, Meta AI}(2024)]{team2024chameleon}
{Chameleon Team, Meta AI}.
\newblock Chameleon: Mixed-modal early-fusion foundation models.
\newblock \emph{arXiv preprint arXiv:2405.09818}, 2024.

\bibitem[Zhan et~al.(2025)Zhan, Dai, Ye, Zhou, Zhang, Liu, Zhang, Yuan, Zhang, Li, Yan, Fu, Gui, Sun, Jiang, and Qiu]{zhan2024anygpt}
Jun Zhan, Junqi Dai, Jiasheng Ye, Yunhua Zhou, Dong Zhang, Zhigeng Liu, Xin Zhang, Ruibin Yuan, Ge~Zhang, Linyang Li, Hang Yan, Jie Fu, Tao Gui, Tianxiang Sun, Yu-Gang Jiang, and Xipeng Qiu.
\newblock Anygpt: Unified multimodal llm with discrete sequence modeling.
\newblock \emph{arXiv preprint arXiv:2402.12226}, 2025.

\bibitem[Yu et~al.(2025{\natexlab{b}})Yu, Wang, Wang, Huang, Ma, He, Cai, Chen, Huang, Zhao, et~al.]{yu2025minicpm}
Tianyu Yu, Zefan Wang, Chongyi Wang, Fuwei Huang, Wenshuo Ma, Zhihui He, Tianchi Cai, Weize Chen, Yuxiang Huang, Yuanqian Zhao, et~al.
\newblock Minicpm-v 4.5: Cooking efficient mllms via architecture, data, and training recipe.
\newblock \emph{arXiv preprint arXiv:2509.18154}, 2025{\natexlab{b}}.

\bibitem[Zhu et~al.(2026{\natexlab{b}})Zhu, Ye, Kang, Yao, Guo, Kuang, Han, Zhuang, Lin, and Povey]{zhu2026omnivoice}
Han Zhu, Lingxuan Ye, Wei Kang, Zengwei Yao, Liyong Guo, Fangjun Kuang, Zhifeng Han, Weiji Zhuang, Long Lin, and Daniel Povey.
\newblock Omnivoice: Towards omnilingual zero-shot text-to-speech with diffusion language models.
\newblock \emph{arXiv preprint arXiv:2604.00688}, 2026{\natexlab{b}}.

\bibitem[D{\'e}fossez et~al.(2024)D{\'e}fossez, Mazar{\'e}, Orsini, Royer, P{\'e}rez, J{\'e}gou, Grave, and Zeghidour]{defossez2024moshi}
Alexandre D{\'e}fossez, Laurent Mazar{\'e}, Manu Orsini, Am{\'e}lie Royer, Patrick P{\'e}rez, Herv{\'e} J{\'e}gou, Edouard Grave, and Neil Zeghidour.
\newblock Moshi: a speech-text foundation model for real-time dialogue.
\newblock \emph{arXiv preprint arXiv:2410.00037}, 2024.

\bibitem[Dong et~al.(2024{\natexlab{c}})Dong, Zhang, Zang, Cao, Wang, Ouyang, Wei, Zhang, Duan, Cao, et~al.]{internlmxcomposer2}
Xiaoyi Dong, Pan Zhang, Yuhang Zang, Yuhang Cao, Bin Wang, Linke Ouyang, Xilin Wei, Songyang Zhang, Haodong Duan, Maosong Cao, et~al.
\newblock Internlm-xcomposer2: Mastering free-form text-image composition and comprehension in vision-language large model.
\newblock \emph{arXiv preprint arXiv:2401.16420}, 2024{\natexlab{c}}.

\bibitem[Wu et~al.(2026)Wu, Yang, Zhang, Wang, Zhu, Leng, Yang, Wang, Wang, Wang, Wang, Zhang, Wang, Zhou, Pu, Li, Zhan, Li, Bing, Song, Liu, Chen, Wang, Wang, Qi, Lu, and Wang]{wu2026visualgenerationnewera}
Keming Wu, Zuhao Yang, Kaichen Zhang, Shizun Wang, Haowei Zhu, Sicong Leng, Zhongyu Yang, Qijie Wang, Sudong Wang, Ziting Wang, Zili Wang, Hui Zhang, Haonan Wang, Hang Zhou, Yifan Pu, Xingxuan Li, Fangneng Zhan, Bo~Li, Lidong Bing, Yuxin Song, Ziwei Liu, Wenhu Chen, Jingdong Wang, Xinchao Wang, Xiaojuan Qi, Shijian Lu, and Bin Wang.
\newblock Visual generation in the new era: An evolution from atomic mapping to agentic world modeling.
\newblock \emph{arXiv preprint arXiv:2604.28185}, 2026.

\bibitem[Du et~al.(2024)Du, Chen, Zhang, Hu, Lu, Yang, Hu, Zheng, Gu, Ma, et~al.]{du2024cosyvoice}
Zhihao Du, Qian Chen, Shiliang Zhang, Kai Hu, Heng Lu, Yexin Yang, Hangrui Hu, Siqi Zheng, Yue Gu, Ziyang Ma, et~al.
\newblock Cosyvoice: A scalable multilingual zero-shot text-to-speech synthesizer based on supervised semantic tokens.
\newblock \emph{arXiv preprint arXiv:2407.05407}, 2024.

\bibitem[Gloeckle et~al.(2024)Gloeckle, Idrissi, Rozi{\`e}re, Lopez-Paz, and Synnaeve]{gloeckle2024better}
Fabian Gloeckle, Badr~Youbi Idrissi, Baptiste Rozi{\`e}re, David Lopez-Paz, and Gabriel Synnaeve.
\newblock Better \& faster large language models via multi-token prediction.
\newblock \emph{arXiv preprint arXiv:2404.19737}, 2024.

\bibitem[Zeng et~al.(2024)Zeng, Du, Liu, Wang, Jiang, Zhao, Dong, and Tang]{zeng2024glm}
Aohan Zeng, Zhengxiao Du, Mingdao Liu, Kedong Wang, Shengmin Jiang, Lei Zhao, Yuxiao Dong, and Jie Tang.
\newblock Glm-4-voice: Towards intelligent and human-like end-to-end spoken chatbot.
\newblock \emph{arXiv preprint arXiv:2412.02612}, 2024.

\bibitem[Radford et~al.(2022)Radford, Kim, Xu, Brockman, McLeavey, and Sutskever]{Radford2022RobustSR}
Alec Radford, Jong~Wook Kim, Tao Xu, Greg Brockman, Christine McLeavey, and Ilya Sutskever.
\newblock Robust speech recognition via large-scale weak supervision.
\newblock In \emph{International Conference on Machine Learning}, 2022.

\bibitem[Peebles and Xie(2023)]{peebles2023scalable}
William Peebles and Saining Xie.
\newblock Scalable diffusion models with transformers.
\newblock In \emph{Proceedings of the IEEE/CVF international conference on computer vision}, pages 4195--4205, 2023.

\bibitem[Xie et~al.(2025{\natexlab{b}})Xie, Ma, Liu, Pang, Li, Zhang, Liao, Ye, Miao, and Yan]{xie2025mini}
Zhifei Xie, Ziyang Ma, Zihang Liu, Kaiyu Pang, Hongyu Li, Jialin Zhang, Yue Liao, Deheng Ye, Chunyan Miao, and Shuicheng Yan.
\newblock Mini-omni-reasoner: Token-level thinking-in-speaking in large speech models.
\newblock \emph{arXiv preprint arXiv:2508.15827}, 2025{\natexlab{b}}.

\bibitem[Le et~al.(2025)Le, Zhu, Kalogeiton, and Samaras]{le2025gravityvideogenerationposttraining}
Minh-Quan Le, Yuanzhi Zhu, Vicky Kalogeiton, and Dimitris Samaras.
\newblock What about gravity in video generation? post-training newton's laws with verifiable rewards.
\newblock \emph{arXiv preprint arXiv:2512.00425}, 2025.

\bibitem[Zhang et~al.(2026{\natexlab{a}})Zhang, Gong, Tan, Zhang, Shen, Zhu, Li, Yao, Shen, and Zou]{zhang2026physrvgphysicsawareunifiedreinforcement}
Qiyuan Zhang, Biao Gong, Shuai Tan, Zheng Zhang, Yujun Shen, Xing Zhu, Yuyuan Li, Kelu Yao, Chunhua Shen, and Changqing Zou.
\newblock Physrvg: Physics-aware unified reinforcement learning for video generative models.
\newblock \emph{arXiv preprint arXiv:2601.11087}, 2026{\natexlab{a}}.

\bibitem[Ravi et~al.(2025)Ravi, Gabeur, Hu, Hu, Ryali, Ma, Khedr, R{\"a}dle, Rolland, Gustafson, et~al.]{ravi2025sam}
Nikhila Ravi, Valentin Gabeur, Yuan-Ting Hu, Ronghang Hu, Chaitanya Ryali, Tengyu Ma, Haitham Khedr, Roman R{\"a}dle, Chloe Rolland, Laura Gustafson, et~al.
\newblock Sam 2: Segment anything in images and videos.
\newblock In \emph{International Conference on Learning Representations}, volume 2025, pages 28085--28128, 2025.

\bibitem[Rafailov et~al.(2024)Rafailov, Sharma, Mitchell, Ermon, Manning, and Finn]{rafailov2024directpreferenceoptimizationlanguage}
Rafael Rafailov, Archit Sharma, Eric Mitchell, Stefano Ermon, Christopher~D. Manning, and Chelsea Finn.
\newblock Direct preference optimization: Your language model is secretly a reward model.
\newblock \emph{arXiv preprint arXiv:2305.18290}, 2024.

\bibitem[Ronneberger et~al.(2015)Ronneberger, Fischer, and Brox]{ronneberger2015u}
Olaf Ronneberger, Philipp Fischer, and Thomas Brox.
\newblock U-net: Convolutional networks for biomedical image segmentation.
\newblock In \emph{International Conference on Medical image computing and computer-assisted intervention}, pages 234--241. Springer, 2015.

\bibitem[Liu et~al.(2022)Liu, Mao, Wu, Feichtenhofer, Darrell, and Xie]{liu2022convnet}
Zhuang Liu, Hanzi Mao, Chao-Yuan Wu, Christoph Feichtenhofer, Trevor Darrell, and Saining Xie.
\newblock A convnet for the 2020s.
\newblock In \emph{Proceedings of the IEEE/CVF conference on computer vision and pattern recognition}, pages 11976--11986, 2022.

\bibitem[Radford et~al.(2021)Radford, Kim, Hallacy, Ramesh, Goh, Agarwal, Sastry, Askell, Mishkin, Clark, et~al.]{radford2021learning}
Alec Radford, Jong~Wook Kim, Chris Hallacy, Aditya Ramesh, Gabriel Goh, Sandhini Agarwal, Girish Sastry, Amanda Askell, Pamela Mishkin, Jack Clark, et~al.
\newblock Learning transferable visual models from natural language supervision.
\newblock In \emph{International conference on machine learning}, pages 8748--8763. PmLR, 2021.

\bibitem[Schuhmann et~al.(2022)Schuhmann, Beaumont, Vencu, Gordon, Wightman, Cherti, Coombes, Katta, Mullis, Wortsman, et~al.]{schuhmann2022laion}
Christoph Schuhmann, Romain Beaumont, Richard Vencu, Cade Gordon, Ross Wightman, Mehdi Cherti, Theo Coombes, Aarush Katta, Clayton Mullis, Mitchell Wortsman, et~al.
\newblock Laion-5b: An open large-scale dataset for training next generation image-text models.
\newblock \emph{Advances in neural information processing systems}, 35:\penalty0 25278--25294, 2022.

\bibitem[Chen et~al.(2015)Chen, Fang, Lin, Vedantam, Gupta, Doll{\'a}r, and Zitnick]{chen2015microsoft}
Xinlei Chen, Hao Fang, Tsung-Yi Lin, Ramakrishna Vedantam, Saurabh Gupta, Piotr Doll{\'a}r, and C~Lawrence Zitnick.
\newblock Microsoft coco captions: Data collection and evaluation server.
\newblock \emph{arXiv preprint arXiv:1504.00325}, 2015.

\bibitem[Sharma et~al.(2018)Sharma, Ding, Goodman, and Soricut]{sharma2018conceptual}
Piyush Sharma, Nan Ding, Sebastian Goodman, and Radu Soricut.
\newblock Conceptual captions: A cleaned, hypernymed, image alt-text dataset for automatic image captioning.
\newblock In \emph{Proceedings of ACL}, 2018.

\bibitem[Thomee et~al.(2016)Thomee, Shamma, Friedland, Elizalde, Ni, Poland, Borth, and Li]{thomee2016yfcc100m}
Bart Thomee, David~A Shamma, Gerald Friedland, Benjamin Elizalde, Karl Ni, Douglas Poland, Damian Borth, and Li-Jia Li.
\newblock Yfcc100m: The new data in multimedia research.
\newblock \emph{Communications of the ACM}, 59\penalty0 (2):\penalty0 64--73, 2016.

\bibitem[Gadre et~al.(2023)Gadre, Ilharco, Fang, Hayase, Smyrnis, Nguyen, Marten, Wortsman, Ghosh, Zhang, et~al.]{gadre2023datacomp}
Samir~Yitzhak Gadre, Gabriel Ilharco, Alex Fang, Jonathan Hayase, Georgios Smyrnis, Thao Nguyen, Ryan Marten, Mitchell Wortsman, Dhruba Ghosh, Jieyu Zhang, et~al.
\newblock Datacomp: In search of the next generation of multimodal datasets.
\newblock \emph{Advances in Neural Information Processing Systems}, 36:\penalty0 27092--27112, 2023.

\bibitem[Antol et~al.(2015)Antol, Agrawal, Lu, Mitchell, Batra, Zitnick, and Parikh]{antol2015vqa}
Stanislaw Antol, Aishwarya Agrawal, Jiasen Lu, Margaret Mitchell, Dhruv Batra, C~Lawrence Zitnick, and Devi Parikh.
\newblock Vqa: Visual question answering.
\newblock In \emph{Proceedings of the IEEE international conference on computer vision}, pages 2425--2433, 2015.

\bibitem[Hudson and Manning(2019)]{hudson2019gqa}
Drew~A Hudson and Christopher~D Manning.
\newblock Gqa: A new dataset for real-world visual reasoning and compositional question answering.
\newblock In \emph{Proceedings of the IEEE/CVF conference on computer vision and pattern recognition}, pages 6700--6709, 2019.

\bibitem[Marino et~al.(2019)Marino, Rastegari, Farhadi, and Mottaghi]{marino2019ok}
Kenneth Marino, Mohammad Rastegari, Ali Farhadi, and Roozbeh Mottaghi.
\newblock Ok-vqa: A visual question answering benchmark requiring external knowledge.
\newblock In \emph{Proceedings of the IEEE/cvf conference on computer vision and pattern recognition}, pages 3195--3204, 2019.

\bibitem[Lu et~al.(2022)Lu, Mishra, Xia, Qiu, Chang, Zhu, Tafjord, Clark, and Kalyan]{lu2022learn}
Pan Lu, Swaroop Mishra, Tanglin Xia, Liang Qiu, Kai-Wei Chang, Song-Chun Zhu, Oyvind Tafjord, Peter Clark, and Ashwin Kalyan.
\newblock Learn to explain: Multimodal reasoning via thought chains for science question answering.
\newblock \emph{Advances in neural information processing systems}, 35:\penalty0 2507--2521, 2022.

\bibitem[Liu et~al.(2023{\natexlab{a}})Liu, Li, Wu, and Lee]{Liu2023VisualIT}
Haotian Liu, Chunyuan Li, Qingyang Wu, and Yong~Jae Lee.
\newblock Visual instruction tuning.
\newblock \emph{ArXiv}, abs/2304.08485, 2023{\natexlab{a}}.

\bibitem[Dai et~al.(2023)Dai, Li, Li, Tiong, Zhao, Wang, Li, Fung, and Hoi]{dai2023instructblip}
Wenliang Dai, Junnan Li, Dongxu Li, Anthony Tiong, Junqi Zhao, Weisheng Wang, Boyang Li, Pascale~N Fung, and Steven Hoi.
\newblock Instructblip: Towards general-purpose vision-language models with instruction tuning.
\newblock \emph{Advances in neural information processing systems}, 36:\penalty0 49250--49267, 2023.

\bibitem[Zhu et~al.(2023)Zhu, Hessel, Awadalla, Gadre, Dodge, Fang, Yu, Schmidt, Wang, and Choi]{zhu2023multimodal}
Wanrong Zhu, Jack Hessel, Anas Awadalla, Samir~Yitzhak Gadre, Jesse Dodge, Alex Fang, Youngjae Yu, Ludwig Schmidt, William~Yang Wang, and Yejin Choi.
\newblock Multimodal c4: An open, billion-scale corpus of images interleaved with text.
\newblock \emph{Advances in Neural Information Processing Systems}, 36:\penalty0 8958--8974, 2023.

\bibitem[Lauren{\c{c}}on et~al.(2023)Lauren{\c{c}}on, Saulnier, Tronchon, Bekman, Singh, Lozhkov, Wang, Karamcheti, Rush, Kiela, et~al.]{laurenccon2023obelics}
Hugo Lauren{\c{c}}on, Lucile Saulnier, L{\'e}o Tronchon, Stas Bekman, Amanpreet Singh, Anton Lozhkov, Thomas Wang, Siddharth Karamcheti, Alexander Rush, Douwe Kiela, et~al.
\newblock Obelics: An open web-scale filtered dataset of interleaved image-text documents.
\newblock \emph{Advances in Neural Information Processing Systems}, 36:\penalty0 71683--71702, 2023.

\bibitem[Li et~al.(2025{\natexlab{b}})Li, Chen, Wang, Wang, Ye, Jin, Chen, He, Gao, Cui, et~al.]{li2025omnicorpus}
Qingyun Li, Zhe Chen, Weiyun Wang, Wenhai Wang, Shenglong Ye, Zhenjiang Jin, Guanzhou Chen, Yinan He, Zhangwei Gao, Erfei Cui, et~al.
\newblock Omnicorpus: A unified multimodal corpus of 10 billion-level images interleaved with text.
\newblock In \emph{International Conference on Learning Representations}, volume 2025, pages 13647--13689, 2025{\natexlab{b}}.

\bibitem[Penha et~al.(2019)Penha, Balan, and Hauff]{penha2019introducing}
Gustavo Penha, Alexandru Balan, and Claudia Hauff.
\newblock Introducing mantis: a novel multi-domain information seeking dialogues dataset.
\newblock \emph{arXiv preprint arXiv:1912.04639}, 2019.

\bibitem[Suhr et~al.(2019)Suhr, Zhou, Zhang, Zhang, Bai, and Artzi]{suhr2019corpus}
Alane Suhr, Stephanie Zhou, Ally Zhang, Iris Zhang, Huajun Bai, and Yoav Artzi.
\newblock A corpus for reasoning about natural language grounded in photographs.
\newblock In \emph{Proceedings of the 57th annual meeting of the association for computational linguistics}, pages 6418--6428, 2019.

\bibitem[Wang et~al.(2025{\natexlab{b}})Wang, Fu, Huang, Li, Liu, Liu, Ma, Xu, Zhou, Zhang, et~al.]{wang2025muirbench}
Fei Wang, Xingyu Fu, James~Y Huang, Zekun Li, Qin Liu, Xiaogeng Liu, Mingyu~Derek Ma, Nan Xu, Wenxuan Zhou, Kai Zhang, et~al.
\newblock Muirbench: A comprehensive benchmark for robust multi-image understanding.
\newblock In \emph{International Conference on Learning Representations}, volume 2025, pages 62624--62650, 2025{\natexlab{b}}.

\bibitem[Fu et~al.(2024)Fu, Hu, Li, Feng, Wang, Lin, Roth, Smith, Ma, and Krishna]{fu2024blink}
Xingyu Fu, Yushi Hu, Bangzheng Li, Yu~Feng, Haoyu Wang, Xudong Lin, Dan Roth, Noah~A Smith, Wei-Chiu Ma, and Ranjay Krishna.
\newblock Blink: Multimodal large language models can see but not perceive.
\newblock In \emph{European Conference on Computer Vision}, pages 148--166. Springer, 2024.

\bibitem[Mathew et~al.(2021)Mathew, Karatzas, and Jawahar]{mathew2021docvqa}
Minesh Mathew, Dimosthenis Karatzas, and CV~Jawahar.
\newblock Docvqa: A dataset for vqa on document images.
\newblock In \emph{Proceedings of the IEEE/CVF winter conference on applications of computer vision}, pages 2200--2209, 2021.

\bibitem[Mathew et~al.(2022)Mathew, Bagal, Tito, Karatzas, Valveny, and Jawahar]{mathew2022infographicvqa}
Minesh Mathew, Viraj Bagal, Rub{\`e}n Tito, Dimosthenis Karatzas, Ernest Valveny, and CV~Jawahar.
\newblock Infographicvqa.
\newblock In \emph{Proceedings of the IEEE/CVF Winter Conference on Applications of Computer Vision}, pages 1697--1706, 2022.

\bibitem[Masry et~al.(2022)Masry, Do, Tan, Joty, and Hoque]{masry2022chartqa}
Ahmed Masry, Xuan~Long Do, Jia~Qing Tan, Shafiq Joty, and Enamul Hoque.
\newblock Chartqa: A benchmark for question answering about charts with visual and logical reasoning.
\newblock In \emph{Findings of the association for computational linguistics: ACL 2022}, pages 2263--2279, 2022.

\bibitem[Singh et~al.(2019)Singh, Natarajan, Shah, Jiang, Chen, Batra, Parikh, and Rohrbach]{singh2019towards}
Amanpreet Singh, Vivek Natarajan, Meet Shah, Yu~Jiang, Xinlei Chen, Dhruv Batra, Devi Parikh, and Marcus Rohrbach.
\newblock Towards vqa models that can read.
\newblock In \emph{Proceedings of the IEEE/CVF conference on computer vision and pattern recognition}, pages 8317--8326, 2019.

\bibitem[Plummer et~al.(2015)Plummer, Wang, Cervantes, Caicedo, Hockenmaier, and Lazebnik]{plummer2015flickr30k}
Bryan~A Plummer, Liwei Wang, Chris~M Cervantes, Juan~C Caicedo, Julia Hockenmaier, and Svetlana Lazebnik.
\newblock Flickr30k entities: Collecting region-to-phrase correspondences for richer image-to-sentence models.
\newblock In \emph{Proceedings of the IEEE international conference on computer vision}, pages 2641--2649, 2015.

\bibitem[Kazemzadeh et~al.(2014)Kazemzadeh, Ordonez, Matten, and Berg]{kazemzadeh2014referitgame}
Sahar Kazemzadeh, Vicente Ordonez, Mark Matten, and Tamara Berg.
\newblock Referitgame: Referring to objects in photographs of natural scenes.
\newblock In \emph{Proceedings of the 2014 conference on empirical methods in natural language processing (EMNLP)}, pages 787--798, 2014.

\bibitem[Chen et~al.(2022)Chen, Li, Frintrop, and Hu]{chen2022msr}
Haoran Chen, Jianmin Li, Simone Frintrop, and Xiaolin Hu.
\newblock The msr-video to text dataset with clean annotations.
\newblock \emph{Computer Vision and Image Understanding}, 225:\penalty0 103581, 2022.

\bibitem[Krishna et~al.(2017)Krishna, Hata, Ren, Fei-Fei, and Carlos~Niebles]{krishna2017dense}
Ranjay Krishna, Kenji Hata, Frederic Ren, Li~Fei-Fei, and Juan Carlos~Niebles.
\newblock Dense-captioning events in videos.
\newblock In \emph{Proceedings of the IEEE international conference on computer vision}, pages 706--715, 2017.

\bibitem[Bain et~al.(2021)Bain, Nagrani, Varol, and Zisserman]{Bain21}
Max Bain, Arsha Nagrani, G{\"u}l Varol, and Andrew Zisserman.
\newblock Frozen in time: A joint video and image encoder for end-to-end retrieval.
\newblock In \emph{IEEE International Conference on Computer Vision}, 2021.

\bibitem[Gemmeke et~al.(2017)Gemmeke, Ellis, Freedman, Jansen, Lawrence, Moore, Plakal, and Ritter]{jort_audioset_2017}
Jort~F. Gemmeke, Daniel P.~W. Ellis, Dylan Freedman, Aren Jansen, Wade Lawrence, R.~Channing Moore, Manoj Plakal, and Marvin Ritter.
\newblock Audio set: An ontology and human-labeled dataset for audio events.
\newblock In \emph{Proc. IEEE ICASSP 2017}, New Orleans, LA, 2017.

\bibitem[Panayotov et~al.(2015)Panayotov, Chen, Povey, and Khudanpur]{panayotov2015librispeech}
Vassil Panayotov, Guoguo Chen, Daniel Povey, and Sanjeev Khudanpur.
\newblock Librispeech: an asr corpus based on public domain audio books.
\newblock In \emph{2015 IEEE international conference on acoustics, speech and signal processing (ICASSP)}, pages 5206--5210. IEEE, 2015.

\bibitem[Ardila et~al.(2020)Ardila, Branson, Davis, Kohler, Meyer, Henretty, Morais, Saunders, Tyers, and Weber]{ardila2020common}
Rosana Ardila, Megan Branson, Kelly Davis, Michael Kohler, Josh Meyer, Michael Henretty, Reuben Morais, Lindsay Saunders, Francis Tyers, and Gregor Weber.
\newblock Common voice: A massively-multilingual speech corpus.
\newblock In \emph{Proceedings of the twelfth language resources and evaluation conference}, pages 4218--4222, 2020.

\bibitem[Drossos et~al.(2020)Drossos, Lipping, and Virtanen]{drossos2020clotho}
Konstantinos Drossos, Samuel Lipping, and Tuomas Virtanen.
\newblock Clotho: An audio captioning dataset.
\newblock In \emph{ICASSP 2020-2020 IEEE International Conference on Acoustics, Speech and Signal Processing (ICASSP)}, pages 736--740. IEEE, 2020.

\bibitem[Wang et~al.(2023{\natexlab{b}})Wang, Montoya, Munechika, Yang, Hoover, and Chau]{wang2023diffusiondb}
Zijie~J Wang, Evan Montoya, David Munechika, Haoyang Yang, Benjamin Hoover, and Duen~Horng Chau.
\newblock Diffusiondb: A large-scale prompt gallery dataset for text-to-image generative models.
\newblock In \emph{Proceedings of the 61st annual meeting of the association for computational linguistics (volume 1: Long papers)}, pages 893--911, 2023{\natexlab{b}}.

\bibitem[Brooks et~al.(2022)Brooks, Holynski, and Efros]{brooks2022instructpix2pix}
Tim Brooks, Aleksander Holynski, and Alexei~A Efros.
\newblock Instructpix2pix: Learning to follow image editing instructions.
\newblock \emph{arXiv preprint arXiv:2211.09800}, 2022.

\bibitem[Zhang et~al.(2023{\natexlab{a}})Zhang, Mo, Chen, Sun, and Su]{Zhang2023MagicBrush}
Kai Zhang, Lingbo Mo, Wenhu Chen, Huan Sun, and Yu~Su.
\newblock Magicbrush: A manually annotated dataset for instruction-guided image editing.
\newblock In \emph{Advances in Neural Information Processing Systems}, 2023{\natexlab{a}}.

\bibitem[Hui et~al.(2024)Hui, Yang, Zhao, Shi, Wang, Wang, Zhou, and Xie]{hui2024hq}
Mude Hui, Siwei Yang, Bingchen Zhao, Yichun Shi, Heng Wang, Peng Wang, Yuyin Zhou, and Cihang Xie.
\newblock Hq-edit: A high-quality dataset for instruction-based image editing.
\newblock \emph{arXiv preprint arXiv:2404.09990}, 2024.

\bibitem[Zhao et~al.(2024{\natexlab{a}})Zhao, Ma, Chen, Si, Wu, An, Yu, Zhang, Li, and Chang]{zhao2024ultraeditinstructionbasedfinegrainedimage}
Haozhe Zhao, Xiaojian Ma, Liang Chen, Shuzheng Si, Rujie Wu, Kaikai An, Peiyu Yu, Minjia Zhang, Qing Li, and Baobao Chang.
\newblock Ultraedit: Instruction-based fine-grained image editing at scale.
\newblock \emph{arXiv preprint arXiv:2407.05282}, 2024{\natexlab{a}}.

\bibitem[Zhang et~al.(2023{\natexlab{b}})Zhang, Rao, and Agrawala]{zhang2023adding}
Lvmin Zhang, Anyi Rao, and Maneesh Agrawala.
\newblock Adding conditional control to text-to-image diffusion models.
\newblock In \emph{Proceedings of the IEEE/CVF international conference on computer vision}, pages 3836--3847, 2023{\natexlab{b}}.

\bibitem[Li et~al.(2023{\natexlab{a}})Li, Liu, Wu, Mu, Yang, Gao, Li, and Lee]{li2023gligen}
Yuheng Li, Haotian Liu, Qingyang Wu, Fangzhou Mu, Jianwei Yang, Jianfeng Gao, Chunyuan Li, and Yong~Jae Lee.
\newblock Gligen: Open-set grounded text-to-image generation.
\newblock In \emph{Proceedings of the IEEE/CVF conference on computer vision and pattern recognition}, pages 22511--22521, 2023{\natexlab{a}}.

\bibitem[Mou et~al.(2024)Mou, Wang, Xie, Wu, Zhang, Qi, and Shan]{mou2024t2i}
Chong Mou, Xintao Wang, Liangbin Xie, Yanze Wu, Jian Zhang, Zhongang Qi, and Ying Shan.
\newblock T2i-adapter: Learning adapters to dig out more controllable ability for text-to-image diffusion models.
\newblock In \emph{Proceedings of the AAAI conference on artificial intelligence}, volume~38, pages 4296--4304, 2024.

\bibitem[Kim et~al.(2021)Kim, Kong, and Son]{kim2021conditional}
Jaehyeon Kim, Jungil Kong, and Juhee Son.
\newblock Conditional variational autoencoder with adversarial learning for end-to-end text-to-speech.
\newblock In \emph{International conference on machine learning}, pages 5530--5540. PMLR, 2021.

\bibitem[An et~al.(2023)An, Yang, Li, Wang, Lin, Liu, Wang, and Luo]{an2023openleaf}
Jie An, Zhengyuan Yang, Linjie Li, Jianfeng Wang, Kevin Lin, Zicheng Liu, Lijuan Wang, and Jiebo Luo.
\newblock Openleaf: Open-domain interleaved image-text generation and evaluation.
\newblock \emph{arXiv preprint arXiv:2310.07749}, 2023.

\bibitem[Chen et~al.(2025{\natexlab{a}})Chen, Li, Yang, Wen, Yang, Gao, Wu, and Chen]{chen2025comm}
Wei Chen, Lin Li, Yongqi Yang, Bin Wen, Fan Yang, Tingting Gao, Yu~Wu, and Long Chen.
\newblock Comm: A coherent interleaved image-text dataset for multimodal understanding and generation.
\newblock In \emph{Proceedings of the Computer Vision and Pattern Recognition Conference}, pages 8073--8082, 2025{\natexlab{a}}.

\bibitem[Ma et~al.(2025{\natexlab{a}})Ma, Liang, Li, Zhang, and Li]{ma2025intersyn}
Yiyi Ma, Yuanzhi Liang, Xiu Li, Chi Zhang, and Xuelong Li.
\newblock Intersyn: Interleaved learning for dynamic motion synthesis in the wild.
\newblock In \emph{Proceedings of the IEEE/CVF International Conference on Computer Vision}, pages 12832--12841, 2025{\natexlab{a}}.

\bibitem[Chen et~al.(2024{\natexlab{a}})Chen, Siarohin, Menapace, Deyneka, Chao, Jeon, Fang, Lee, Ren, Yang, and Tulyakov]{chen2024panda70m}
Tsai-Shien Chen, Aliaksandr Siarohin, Willi Menapace, Ekaterina Deyneka, Hsiang-wei Chao, Byung~Eun Jeon, Yuwei Fang, Hsin-Ying Lee, Jian Ren, Ming-Hsuan Yang, and Sergey Tulyakov.
\newblock Panda-70m: Captioning 70m videos with multiple cross-modality teachers.
\newblock In \emph{Proceedings of the IEEE/CVF Conference on Computer Vision and Pattern Recognition}, 2024{\natexlab{a}}.

\bibitem[Nan et~al.(2024)Nan, Xie, Zhou, Fan, Yang, Chen, Li, Yang, and Tai]{nan2024openvid}
Kepan Nan, Rui Xie, Penghao Zhou, Tiehan Fan, Zhenheng Yang, Zhijie Chen, Xiang Li, Jian Yang, and Ying Tai.
\newblock Openvid-1m: A large-scale high-quality dataset for text-to-video generation.
\newblock \emph{arXiv preprint arXiv:2407.02371}, 2024.

\bibitem[Tan et~al.(2024)Tan, Yang, Qin, and Li]{tan2024vidgen}
Zhiyu Tan, Xiaomeng Yang, Luozheng Qin, and Hao Li.
\newblock Vidgen-1m: A large-scale dataset for text-to-video generation.
\newblock \emph{arXiv preprint arXiv:2408.02629}, 2024.

\bibitem[Zen et~al.(2019)Zen, Dang, Clark, Zhang, Weiss, Jia, Chen, and Wu]{zen2019libritts}
Heiga Zen, Viet Dang, Rob Clark, Yu~Zhang, Ron~J Weiss, Ye~Jia, Zhifeng Chen, and Yonghui Wu.
\newblock Libritts: A corpus derived from librispeech for text-to-speech.
\newblock \emph{arXiv preprint arXiv:1904.02882}, 2019.

\bibitem[Yamagishi et~al.(2019)Yamagishi, Veaux, and MacDonald]{yamagishi2019cstr}
Junichi Yamagishi, Christophe Veaux, and Kirsten MacDonald.
\newblock Cstr vctk corpus: English multi-speaker corpus for cstr voice cloning toolkit (version 0.92).
\newblock \emph{The Rainbow Passage which the speakers read out can be found in the International Dialects of English Archive:(http://web. ku. edu/\~{} idea/readings/rainbow. htm).}, 2019.

\bibitem[Chen et~al.(2021)Chen, Chai, Wang, Du, Zhang, Weng, Su, Povey, Trmal, Zhang, et~al.]{chen2021gigaspeech}
Guoguo Chen, Shuzhou Chai, Guanbo Wang, Jiayu Du, Wei-Qiang Zhang, Chao Weng, Dan Su, Daniel Povey, Jan Trmal, Junbo Zhang, et~al.
\newblock Gigaspeech: An evolving, multi-domain asr corpus with 10,000 hours of transcribed audio.
\newblock \emph{arXiv preprint arXiv:2106.06909}, 2021.

\bibitem[He et~al.(2024{\natexlab{a}})He, Shang, Wang, Li, Gu, Hua, Liu, Yang, Li, Shi, et~al.]{he2024emilia}
Haorui He, Zengqiang Shang, Chaoren Wang, Xuyuan Li, Yicheng Gu, Hua Hua, Liwei Liu, Chen Yang, Jiaqi Li, Peiyang Shi, et~al.
\newblock Emilia: An extensive, multilingual, and diverse speech dataset for large-scale speech generation.
\newblock In \emph{2024 IEEE Spoken Language Technology Workshop (SLT)}, pages 885--890. IEEE, 2024{\natexlab{a}}.

\bibitem[Kim et~al.(2019)Kim, Kim, Lee, and Kim]{kim-NAACL-HLT-2019}
Chris~Dongjoo Kim, Byeongchang Kim, Hyunmin Lee, and Gunhee Kim.
\newblock {AudioCaps: Generating Captions for Audios in The Wild}.
\newblock In \emph{NAACL-HLT}, 2019.

\bibitem[Mei et~al.(2024)Mei, Meng, Liu, Kong, Ko, Zhao, Plumbley, Zou, and Wang]{mei2023wavcaps}
Xinhao Mei, Chutong Meng, Haohe Liu, Qiuqiang Kong, Tom Ko, Chengqi Zhao, Mark~D. Plumbley, Yuexian Zou, and Wenwu Wang.
\newblock Wav{C}aps: A {ChatGPT}-assisted weakly-labelled audio captioning dataset for audio-language multimodal research.
\newblock \emph{IEEE/ACM Transactions on Audio, Speech, and Language Processing}, pages 1--15, 2024.

\bibitem[Agostinelli et~al.(2023)Agostinelli, Denk, Borsos, Engel, Verzetti, Caillon, Huang, Jansen, Roberts, Tagliasacchi, et~al.]{agostinelli2023musiclm}
Andrea Agostinelli, Timo~I Denk, Zal{\'a}n Borsos, Jesse Engel, Mauro Verzetti, Antoine Caillon, Qingqing Huang, Aren Jansen, Adam Roberts, Marco Tagliasacchi, et~al.
\newblock Musiclm: Generating music from text.
\newblock \emph{arXiv preprint arXiv:2301.11325}, 2023.

\bibitem[Yao et~al.(2022)Yao, Chen, Yang, and Narasimhan]{yao2022webshop}
Shunyu Yao, Howard Chen, John Yang, and Karthik Narasimhan.
\newblock Webshop: Towards scalable real-world web interaction with grounded language agents.
\newblock \emph{Advances in Neural Information Processing Systems}, 35:\penalty0 20744--20757, 2022.

\bibitem[Deng et~al.(2023)Deng, Gu, Zheng, Chen, Stevens, Wang, Sun, and Su]{deng2023mind2web}
Xiang Deng, Yu~Gu, Boyuan Zheng, Shijie Chen, Samuel Stevens, Boshi Wang, Huan Sun, and Yu~Su.
\newblock Mind2web: Towards a generalist agent for the web.
\newblock \emph{arXiv preprint arXiv:2306.06070}, 2023.

\bibitem[Zhou et~al.(2024{\natexlab{b}})Zhou, Xu, Zhu, Zhou, Lo, Sridhar, Cheng, Bisk, Fried, Alon, et~al.]{zhou2024webarena}
Shuyan Zhou, Frank~F Xu, Hao Zhu, Xuhui Zhou, Robert Lo, Abishek Sridhar, Xianyi Cheng, Yonatan Bisk, Daniel Fried, Uri Alon, et~al.
\newblock Webarena: A realistic web environment for building autonomous agents.
\newblock \emph{ICLR}, 2024{\natexlab{b}}.

\bibitem[Koh et~al.(2024)Koh, Lo, Jang, Duvvur, Lim, Huang, Neubig, Zhou, Salakhutdinov, and Fried]{koh2024visualwebarena}
Jing~Yu Koh, Robert Lo, Lawrence Jang, Vikram Duvvur, Ming~Chong Lim, Po-Yu Huang, Graham Neubig, Shuyan Zhou, Ruslan Salakhutdinov, and Daniel Fried.
\newblock Visualwebarena: Evaluating multimodal agents on realistic visual web tasks.
\newblock \emph{arXiv preprint arXiv:2401.13649}, 2024.

\bibitem[Lù et~al.(2024)Lù, Kasner, and Reddy]{lù2024weblinx}
Xing~Han Lù, Zdeněk Kasner, and Siva Reddy.
\newblock Weblinx: Real-world website navigation with multi-turn dialogue.
\newblock \emph{arXiv preprint arXiv:2402.05930}, 2024.

\bibitem[He et~al.(2024{\natexlab{b}})He, Yao, Ma, Yu, Dai, Zhang, Lan, and Yu]{he2024webvoyager}
Hongliang He, Wenlin Yao, Kaixin Ma, Wenhao Yu, Yong Dai, Hongming Zhang, Zhenzhong Lan, and Dong Yu.
\newblock Webvoyager: Building an end-to-end web agent with large multimodal models.
\newblock \emph{arXiv preprint arXiv:2401.13919}, 2024{\natexlab{b}}.

\bibitem[Rawles et~al.(2023)Rawles, Li, Rodriguez, Riva, and Lillicrap]{rawles2023androidinthewild}
Christopher Rawles, Alice Li, Daniel Rodriguez, Oriana Riva, and Timothy Lillicrap.
\newblock Androidinthewild: A large-scale dataset for android device control.
\newblock \emph{Advances in Neural Information Processing Systems}, 36:\penalty0 59708--59728, 2023.

\bibitem[Deka et~al.(2017)Deka, Huang, Franzen, Hibschman, Afergan, Li, Nichols, and Kumar]{deka2017rico}
Biplab Deka, Zifeng Huang, Chad Franzen, Joshua Hibschman, Daniel Afergan, Yang Li, Jeffrey Nichols, and Ranjitha Kumar.
\newblock Rico: A mobile app dataset for building data-driven design applications.
\newblock In \emph{Proceedings of the 30th annual ACM symposium on user interface software and technology}, pages 845--854, 2017.

\bibitem[Baechler et~al.(2024)Baechler, Sunkara, Wang, Zubach, Mansoor, Etter, Cărbune, Lin, Chen, and Sharma]{baechler2024screenai}
Gilles Baechler, Srinivas Sunkara, Maria Wang, Fedir Zubach, Hassan Mansoor, Vincent Etter, Victor Cărbune, Jason Lin, Jindong Chen, and Abhanshu Sharma.
\newblock Screenai: A vision-language model for ui and infographics understanding.
\newblock \emph{arXiv preprint arXiv:2402.04615}, 2024.

\bibitem[Cheng et~al.(2024)Cheng, Sun, Chu, Xu, YanTao, Zhang, and Wu]{cheng2024seeclick}
Kanzhi Cheng, Qiushi Sun, Yougang Chu, Fangzhi Xu, Li~YanTao, Jianbing Zhang, and Zhiyong Wu.
\newblock {S}ee{C}lick: Harnessing {GUI} grounding for advanced visual {GUI} agents.
\newblock In \emph{Proceedings of the 62nd Annual Meeting of the Association for Computational Linguistics (Volume 1: Long Papers)}, pages 9313--9332, Bangkok, Thailand, August 2024. Association for Computational Linguistics.

\bibitem[Xie et~al.(2024{\natexlab{a}})Xie, Zhang, Chen, Li, Zhao, Cao, Hua, Cheng, Shin, Lei, Liu, Xu, Zhou, Savarese, Xiong, Zhong, and Yu]{OSWorld}
Tianbao Xie, Danyang Zhang, Jixuan Chen, Xiaochuan Li, Siheng Zhao, Ruisheng Cao, Toh~Jing Hua, Zhoujun Cheng, Dongchan Shin, Fangyu Lei, Yitao Liu, Yiheng Xu, Shuyan Zhou, Silvio Savarese, Caiming Xiong, Victor Zhong, and Tao Yu.
\newblock Osworld: Benchmarking multimodal agents for open-ended tasks in real computer environments.
\newblock \emph{arXiv preprint arXiv:2404.07972}, 2024{\natexlab{a}}.

\bibitem[Bonatti et~al.(2024)Bonatti, Zhao, Bonacci, Dupont, Abdali, Li, Lu, Wagle, Koishida, Bucker, et~al.]{bonatti2024windows}
Rogerio Bonatti, Dan Zhao, Francesco Bonacci, Dillon Dupont, Sara Abdali, Yinheng Li, Yadong Lu, Justin Wagle, Kazuhito Koishida, Arthur Bucker, et~al.
\newblock Windows agent arena: Evaluating multi-modal os agents at scale.
\newblock \emph{arXiv preprint arXiv:2409.08264}, 2024.

\bibitem[Shridhar et~al.(2021)Shridhar, Yuan, C\^ot\'e, Bisk, Trischler, and Hausknecht]{ALFWorld20}
Mohit Shridhar, Xingdi Yuan, Marc-Alexandre C\^ot\'e, Yonatan Bisk, Adam Trischler, and Matthew Hausknecht.
\newblock {ALFWorld: Aligning Text and Embodied Environments for Interactive Learning}.
\newblock In \emph{Proceedings of the International Conference on Learning Representations (ICLR)}, 2021.

\bibitem[Walke et~al.(2023)Walke, Black, Lee, Kim, Du, Zheng, Zhao, Hansen-Estruch, Vuong, He, Myers, Fang, Finn, and Levine]{walke2023bridgedata}
Homer Walke, Kevin Black, Abraham Lee, Moo~Jin Kim, Max Du, Chongyi Zheng, Tony Zhao, Philippe Hansen-Estruch, Quan Vuong, Andre He, Vivek Myers, Kuan Fang, Chelsea Finn, and Sergey Levine.
\newblock Bridgedata v2: A dataset for robot learning at scale.
\newblock In \emph{Conference on Robot Learning (CoRL)}, 2023.

\bibitem[O’Neill et~al.(2024)O’Neill, Rehman, Maddukuri, Gupta, Padalkar, Lee, Pooley, Gupta, Mandlekar, Jain, et~al.]{open_x_embodiment_rt_x_2023}
Abby O’Neill, Abdul Rehman, Abhiram Maddukuri, Abhishek Gupta, Abhishek Padalkar, Abraham Lee, Acorn Pooley, Agrim Gupta, Ajay Mandlekar, Ajinkya Jain, et~al.
\newblock Open x-embodiment: Robotic learning datasets and rt-x models: Open x-embodiment collaboration 0.
\newblock In \emph{2024 IEEE International Conference on Robotics and Automation (ICRA)}, pages 6892--6903. IEEE, 2024.

\bibitem[Yang et~al.(2025{\natexlab{a}})Yang, Tan, Wu, Zheng, Peng, Liang, Gu, Cai, Ye, Jang, et~al.]{yang2025magma}
Jianwei Yang, Reuben Tan, Qianhui Wu, Ruijie Zheng, Baolin Peng, Yongyuan Liang, Yu~Gu, Mu~Cai, Seonghyeon Ye, Joel Jang, et~al.
\newblock Magma: A foundation model for multimodal ai agents.
\newblock In \emph{Proceedings of the computer vision and pattern recognition conference}, pages 14203--14214, 2025{\natexlab{a}}.

\bibitem[Sun et~al.(2023{\natexlab{a}})Sun, Shen, Cao, Liu, Li, Shen, Gan, Gui, Wang, Yang, Keutzer, and Darrell]{2023llavarlhf}
Zhiqing Sun, Sheng Shen, Shengcao Cao, Haotian Liu, Chunyuan Li, Yikang Shen, Chuang Gan, Liang-Yan Gui, Yu-Xiong Wang, Yiming Yang, Kurt Keutzer, and Trevor Darrell.
\newblock Aligning large multimodal models with factually augmented rlhf.
\newblock \emph{arXiv:2309.14525}, 2023{\natexlab{a}}.

\bibitem[Yu et~al.(2024{\natexlab{a}})Yu, Yao, Zhang, He, Han, Cui, Hu, Liu, Zheng, Sun, and Chua]{yu2024rlhfvtrustworthymllmsbehavior}
Tianyu Yu, Yuan Yao, Haoye Zhang, Taiwen He, Yifeng Han, Ganqu Cui, Jinyi Hu, Zhiyuan Liu, Hai-Tao Zheng, Maosong Sun, and Tat-Seng Chua.
\newblock Rlhf-v: Towards trustworthy mllms via behavior alignment from fine-grained correctional human feedback.
\newblock \emph{arXiv preprint arXiv:2312.00849}, 2024{\natexlab{a}}.

\bibitem[Li et~al.(2024{\natexlab{a}})Li, Xie, Li, Chen, Wang, Chen, Yang, Wang, Kong, and Liu]{li2024vlfeedback}
Lei Li, Zhihui Xie, Mukai Li, Shunian Chen, Peiyi Wang, Liang Chen, Yazheng Yang, Benyou Wang, Lingpeng Kong, and Qi~Liu.
\newblock Vlfeedback: A large-scale ai feedback dataset for large vision-language models alignment.
\newblock In \emph{Proceedings of the 2024 Conference on Empirical Methods in Natural Language Processing}, pages 6227--6246, 2024{\natexlab{a}}.

\bibitem[Yu et~al.(2024{\natexlab{b}})Yu, Zhang, Li, Xu, Yao, Chen, Lu, Cui, Dang, He, Feng, Song, Zheng, Liu, Chua, and Sun]{yu2024rlaifv}
Tianyu Yu, Haoye Zhang, Qiming Li, Qixin Xu, Yuan Yao, Da~Chen, Xiaoman Lu, Ganqu Cui, Yunkai Dang, Taiwen He, Xiaocheng Feng, Jun Song, Bo~Zheng, Zhiyuan Liu, Tat-Seng Chua, and Maosong Sun.
\newblock Rlaif-v: Open-source ai feedback leads to super gpt-4v trustworthiness.
\newblock \emph{arXiv preprint arXiv:2405.17220}, 2024{\natexlab{b}}.

\bibitem[Zhang et~al.(2025{\natexlab{a}})Zhang, Yu, Tian, Fu, Li, Zeng, Xie, Shi, Zhang, Wu, et~al.]{zhang2025mm}
Yi-Fan Zhang, Tao Yu, Haochen Tian, Chaoyou Fu, Peiyan Li, Jianshu Zeng, Wulin Xie, Yang Shi, Huanyu Zhang, Junkang Wu, et~al.
\newblock Mm-rlhf: The next step forward in multimodal llm alignment.
\newblock \emph{arXiv preprint arXiv:2502.10391}, 2025{\natexlab{a}}.

\bibitem[Xie et~al.(2024{\natexlab{b}})Xie, Li, Xu, and Kan]{xie2024v}
Yuxi Xie, Guanzhen Li, Xiao Xu, and Min-Yen Kan.
\newblock V-dpo: Mitigating hallucination in large vision language models via vision-guided direct preference optimization.
\newblock In \emph{Findings of the Association for Computational Linguistics: EMNLP 2024}, pages 13258--13273, 2024{\natexlab{b}}.

\bibitem[Zhang et~al.(2024{\natexlab{b}})Zhang, Chen, Zheng, Gao, Zheng, Fu, Yin, Jin, Qiao, Huang, Zhao, Gui, and Shao]{zhang2024spavl}
Yongting Zhang, Lu~Chen, Guodong Zheng, Yifeng Gao, Rui Zheng, Jinlan Fu, Zhenfei Yin, Senjie Jin, Yu~Qiao, Xuanjing Huang, Feng Zhao, Tao Gui, and Jing Shao.
\newblock Spa-vl: A comprehensive safety preference alignment dataset for vision language model.
\newblock \emph{arXiv preprint arXiv:2406.12030}, 2024{\natexlab{b}}.

\bibitem[Ji et~al.(2026)Ji, Chen, Pan, Zhu, Li, Hong, Chen, Zhou, Wang, Dai, Chan, Han, Guo, and Yang]{ji2026safe}
Jiaming Ji, Xinyu Chen, Rui Pan, Han Zhu, Jiahao Li, Donghai Hong, Boyuan Chen, Jiayi Zhou, Kaile Wang, Juntao Dai, Chi-Min Chan, Sirui Han, Yike Guo, and Yaodong Yang.
\newblock Safe {RLHF}-v: Safe reinforcement learning from multi-modal human feedback.
\newblock In \emph{The Thirty-ninth Annual Conference on Neural Information Processing Systems}, 2026.

\bibitem[Xu et~al.(2023)Xu, Liu, Wu, Tong, Li, Ding, Tang, and Dong]{xu2023imagereward}
Jiazheng Xu, Xiao Liu, Yuchen Wu, Yuxuan Tong, Qinkai Li, Ming Ding, Jie Tang, and Yuxiao Dong.
\newblock Imagereward: Learning and evaluating human preferences for text-to-image generation.
\newblock \emph{Advances in Neural Information Processing Systems}, 36:\penalty0 15903--15935, 2023.

\bibitem[Kirstain et~al.(2023)Kirstain, Polyak, Singer, Matiana, Penna, and Levy]{kirstain2023pick}
Yuval Kirstain, Adam Polyak, Uriel Singer, Shahbuland Matiana, Joe Penna, and Omer Levy.
\newblock Pick-a-pic: An open dataset of user preferences for text-to-image generation.
\newblock \emph{Advances in neural information processing systems}, 36:\penalty0 36652--36663, 2023.

\bibitem[Wu et~al.(2023)Wu, Hao, Sun, Chen, Zhu, Zhao, and Li]{wu2023human}
Xiaoshi Wu, Yiming Hao, Keqiang Sun, Yixiong Chen, Feng Zhu, Rui Zhao, and Hongsheng Li.
\newblock Human preference score v2: A solid benchmark for evaluating human preferences of text-to-image synthesis.
\newblock \emph{arXiv preprint arXiv:2306.09341}, 2023.

\bibitem[Huang et~al.(2024{\natexlab{a}})Huang, He, Yu, Zhang, Si, Jiang, Zhang, Wu, Jin, Chanpaisit, Wang, Chen, Wang, Lin, Qiao, and Liu]{huang2023vbench}
Ziqi Huang, Yinan He, Jiashuo Yu, Fan Zhang, Chenyang Si, Yuming Jiang, Yuanhan Zhang, Tianxing Wu, Qingyang Jin, Nattapol Chanpaisit, Yaohui Wang, Xinyuan Chen, Limin Wang, Dahua Lin, Yu~Qiao, and Ziwei Liu.
\newblock {VBench}: Comprehensive benchmark suite for video generative models.
\newblock In \emph{Proceedings of the IEEE/CVF Conference on Computer Vision and Pattern Recognition}, 2024{\natexlab{a}}.

\bibitem[Huang et~al.(2024{\natexlab{b}})Huang, Zhang, Xu, He, Yu, Dong, Ma, Chanpaisit, Si, Jiang, Wang, Chen, Chen, Wang, Lin, Qiao, and Liu]{huang2024vbench++}
Ziqi Huang, Fan Zhang, Xiaojie Xu, Yinan He, Jiashuo Yu, Ziyue Dong, Qianli Ma, Nattapol Chanpaisit, Chenyang Si, Yuming Jiang, Yaohui Wang, Xinyuan Chen, Ying-Cong Chen, Limin Wang, Dahua Lin, Yu~Qiao, and Ziwei Liu.
\newblock Vbench++: Comprehensive and versatile benchmark suite for video generative models.
\newblock \emph{arXiv preprint arXiv:2411.13503}, 2024{\natexlab{b}}.

\bibitem[Gurari et~al.(2018)Gurari, Li, Stangl, Guo, Lin, Grauman, Luo, and Bigham]{gurari2018vizwiz}
Danna Gurari, Qing Li, Abigale~J Stangl, Anhong Guo, Chi Lin, Kristen Grauman, Jiebo Luo, and Jeffrey~P Bigham.
\newblock Vizwiz grand challenge: Answering visual questions from blind people.
\newblock In \emph{Proceedings of the IEEE conference on computer vision and pattern recognition}, pages 3608--3617, 2018.

\bibitem[Alayrac et~al.(2022)Alayrac, Donahue, Luc, Miech, Barr, Hasson, Lenc, Mensch, Millican, Reynolds, et~al.]{alayrac2022flamingo}
Jean-Baptiste Alayrac, Jeff Donahue, Pauline Luc, Antoine Miech, Iain Barr, Yana Hasson, Karel Lenc, Arthur Mensch, Katherine Millican, Malcolm Reynolds, et~al.
\newblock Flamingo: a visual language model for few-shot learning.
\newblock \emph{Advances in neural information processing systems}, 35:\penalty0 23716--23736, 2022.

\bibitem[Kahou et~al.(2017)Kahou, Michalski, Atkinson, K{\'a}d{\'a}r, Trischler, and Bengio]{kahou2017figureqa}
Samira~Ebrahimi Kahou, Vincent Michalski, Adam Atkinson, {\'A}kos K{\'a}d{\'a}r, Adam Trischler, and Yoshua Bengio.
\newblock Figureqa: An annotated figure dataset for visual reasoning.
\newblock \emph{arXiv preprint arXiv:1710.07300}, 2017.

\bibitem[Methani et~al.(2020)Methani, Ganguly, Khapra, and Kumar]{methani2020plotqa}
Nitesh Methani, Pritha Ganguly, Mitesh~M Khapra, and Pratyush Kumar.
\newblock Plotqa: Reasoning over scientific plots.
\newblock In \emph{Proceedings of the ieee/cvf winter conference on applications of computer vision}, pages 1527--1536, 2020.

\bibitem[Kafle et~al.(2018)Kafle, Price, Cohen, and Kanan]{kafle2018dvqa}
Kushal Kafle, Brian Price, Scott Cohen, and Christopher Kanan.
\newblock Dvqa: Understanding data visualizations via question answering.
\newblock In \emph{Proceedings of the IEEE conference on computer vision and pattern recognition}, pages 5648--5656, 2018.

\bibitem[Peng et~al.(2024)Peng, Wang, Dong, Hao, Huang, Ma, Ye, and Wei]{peng2024grounding}
Zhiliang Peng, Wenhui Wang, Li~Dong, Yaru Hao, Shaohan Huang, Shuming Ma, Qixiang Ye, and Furu Wei.
\newblock Grounding multimodal large language models to the world.
\newblock In \emph{International Conference on Learning Representations}, volume 2024, pages 51575--51598, 2024.

\bibitem[Rasheed et~al.(2024)Rasheed, Maaz, Shaji, Shaker, Khan, Cholakkal, Anwer, Xing, Yang, and Khan]{rasheed2024glamm}
Hanoona Rasheed, Muhammad Maaz, Sahal Shaji, Abdelrahman Shaker, Salman Khan, Hisham Cholakkal, Rao~M Anwer, Eric Xing, Ming-Hsuan Yang, and Fahad~S Khan.
\newblock Glamm: Pixel grounding large multimodal model.
\newblock In \emph{Proceedings of the IEEE/CVF Conference on Computer Vision and Pattern Recognition}, pages 13009--13018, 2024.

\bibitem[Miech et~al.(2019)Miech, Zhukov, Alayrac, Tapaswi, Laptev, and Sivic]{miech2019howto100m}
Antoine Miech, Dimitri Zhukov, Jean-Baptiste Alayrac, Makarand Tapaswi, Ivan Laptev, and Josef Sivic.
\newblock Howto100m: Learning a text-video embedding by watching hundred million narrated video clips.
\newblock In \emph{Proceedings of the IEEE/CVF international conference on computer vision}, pages 2630--2640, 2019.

\bibitem[Muhammad~Maaz and Khan(2023)]{Maaz2023VideoChatGPT}
Salman~Khan Muhammad~Maaz, Hanoona~Rasheed and Fahad Khan.
\newblock Video-chatgpt: Towards detailed video understanding via large vision and language models.
\newblock \emph{ArXiv 2306.05424}, 2023.

\bibitem[Fonseca et~al.(2021)Fonseca, Favory, Pons, Font, and Serra]{fonseca2021fsd50k}
Eduardo Fonseca, Xavier Favory, Jordi Pons, Frederic Font, and Xavier Serra.
\newblock Fsd50k: an open dataset of human-labeled sound events.
\newblock \emph{IEEE/ACM Transactions on Audio, Speech, and Language Processing}, 30:\penalty0 829--852, 2021.

\bibitem[Tang et~al.(2024)Tang, Yu, Sun, Chen, Tan, Li, Lu, Ma, and Zhang]{tang2024salmonn}
Changli Tang, Wenyi Yu, Guangzhi Sun, Xianzhao Chen, Tian Tan, Wei Li, Lu~Lu, Zejun Ma, and Chao Zhang.
\newblock Salmonn: Towards generic hearing abilities for large language models.
\newblock In \emph{International Conference on Learning Representations}, volume 2024, pages 16607--16629, 2024.

\bibitem[Chu et~al.(2024)Chu, Xu, Yang, Wei, Wei, Guo, Leng, Lv, He, Lin, et~al.]{chu2024qwen2}
Yunfei Chu, Jin Xu, Qian Yang, Haojie Wei, Xipin Wei, Zhifang Guo, Yichong Leng, Yuanjun Lv, Jinzheng He, Junyang Lin, et~al.
\newblock Qwen2-audio technical report.
\newblock \emph{arXiv preprint arXiv:2407.10759}, 2024.

\bibitem[Sun et~al.(2023{\natexlab{b}})Sun, Pan, Ge, Li, Duan, Wu, Zhang, Zhou, Qin, Wang, et~al.]{sun2023journeydb}
Keqiang Sun, Junting Pan, Yuying Ge, Hao Li, Haodong Duan, Xiaoshi Wu, Renrui Zhang, Aojun Zhou, Zipeng Qin, Yi~Wang, et~al.
\newblock Journeydb: A benchmark for generative image understanding.
\newblock \emph{Advances in neural information processing systems}, 36:\penalty0 49659--49678, 2023{\natexlab{b}}.

\bibitem[Koizumi et~al.(2023)Koizumi, Zen, Karita, Ding, Yatabe, Morioka, Bacchiani, Zhang, Han, and Bapna]{koizumi2023libritts}
Yuma Koizumi, Heiga Zen, Shigeki Karita, Yifan Ding, Kohei Yatabe, Nobuyuki Morioka, Michiel Bacchiani, Yu~Zhang, Wei Han, and Ankur Bapna.
\newblock Libritts-r: A restored multi-speaker text-to-speech corpus.
\newblock \emph{arXiv preprint arXiv:2305.18802}, 2023.

\bibitem[Liu et~al.(2023{\natexlab{b}})Liu, Chen, Yuan, Mei, Liu, Mandic, Wang, and Plumbley]{liu2023audioldm}
Haohe Liu, Zehua Chen, Yi~Yuan, Xinhao Mei, Xubo Liu, Danilo Mandic, Wenwu Wang, and Mark~D Plumbley.
\newblock {AudioLDM}: Text-to-audio generation with latent diffusion models.
\newblock \emph{Proceedings of the International Conference on Machine Learning}, pages 21450--21474, 2023{\natexlab{b}}.

\bibitem[Doh et~al.(2023)Doh, Choi, Lee, and Nam]{doh2023lp}
SeungHeon Doh, Keunwoo Choi, Jongpil Lee, and Juhan Nam.
\newblock Lp-musiccaps: Llm-based pseudo music captioning.
\newblock \emph{arXiv preprint arXiv:2307.16372}, 2023.

\bibitem[Hong et~al.(2023)Hong, Wang, Lv, Xu, Yu, Ji, Wang, Wang, Dong, Ding, and Tang]{hong2023cogagent}
Wenyi Hong, Weihan Wang, Qingsong Lv, Jiazheng Xu, Wenmeng Yu, Junhui Ji, Yan Wang, Zihan Wang, Yuxiao Dong, Ming Ding, and Jie Tang.
\newblock Cogagent: A visual language model for gui agents.
\newblock \emph{arXiv preprint arXiv:2312.08914}, 2023.

\bibitem[Kapoor et~al.(2024)Kapoor, Butala, Russak, Koh, Kamble, Alshikh, and Salakhutdinov]{kapoor2024omniact}
Raghav Kapoor, Yash~Parag Butala, Melisa Russak, Jing~Yu Koh, Kiran Kamble, Waseem Alshikh, and Ruslan Salakhutdinov.
\newblock Omniact: A dataset and benchmark for enabling multimodal generalist autonomous agents for desktop and web.
\newblock \emph{arXiv preprint arXiv:2402.17553}, 2024.

\bibitem[Brohan et~al.(2022)Brohan, Brown, Carbajal, Chebotar, Dabis, Finn, Gopalakrishnan, Hausman, Herzog, Hsu, et~al.]{rt12022arxiv}
Anthony Brohan, Noah Brown, Justice Carbajal, Yevgen Chebotar, Joseph Dabis, Chelsea Finn, Keerthana Gopalakrishnan, Karol Hausman, Alex Herzog, Jasmine Hsu, et~al.
\newblock Rt-1: Robotics transformer for real-world control at scale.
\newblock In \emph{arXiv preprint arXiv:2212.06817}, 2022.

\bibitem[Li et~al.(2023{\natexlab{b}})Li, Xie, Li, Chen, Wang, Chen, Yang, Wang, and Kong]{2023vlfeedback}
Lei Li, Zhihui Xie, Mukai Li, Shunian Chen, Peiyi Wang, Liang Chen, Yazheng Yang, Benyou Wang, and Lingpeng Kong.
\newblock Silkie: Preference distillation for large visual language models.
\newblock \emph{arXiv:2312.10665}, 2023{\natexlab{b}}.

\bibitem[Ouali et~al.(2024)Ouali, Bulat, Martinez, and Tzimiropoulos]{ouali2024clip}
Yassine Ouali, Adrian Bulat, Brais Martinez, and Georgios Tzimiropoulos.
\newblock Clip-dpo: Vision-language models as a source of preference for fixing hallucinations in lvlms.
\newblock In \emph{European Conference on Computer Vision}, pages 395--413. Springer, 2024.

\bibitem[Wallace et~al.(2024)Wallace, Dang, Rafailov, Zhou, Lou, Purushwalkam, Ermon, Xiong, Joty, and Naik]{wallace2024diffusion}
Bram Wallace, Meihua Dang, Rafael Rafailov, Linqi Zhou, Aaron Lou, Senthil Purushwalkam, Stefano Ermon, Caiming Xiong, Shafiq Joty, and Nikhil Naik.
\newblock Diffusion model alignment using direct preference optimization.
\newblock In \emph{Proceedings of the IEEE/CVF Conference on Computer Vision and Pattern Recognition}, pages 8228--8238, 2024.

\bibitem[Black et~al.(2024)Black, Janner, Du, Kostrikov, and Levine]{black2024training}
Kevin Black, Michael Janner, Yilun Du, Ilya Kostrikov, and Sergey Levine.
\newblock Training diffusion models with reinforcement learning.
\newblock In \emph{International Conference on Learning Representations}, volume 2024, pages 4965--4987, 2024.

\bibitem[Fan et~al.(2023)Fan, Watkins, Du, Liu, Ryu, Boutilier, Abbeel, Ghavamzadeh, Lee, and Lee]{fan2023dpok}
Ying Fan, Olivia Watkins, Yuqing Du, Hao Liu, Moonkyung Ryu, Craig Boutilier, Pieter Abbeel, Mohammad Ghavamzadeh, Kangwook Lee, and Kimin Lee.
\newblock Dpok: Reinforcement learning for fine-tuning text-to-image diffusion models.
\newblock \emph{Advances in Neural Information Processing Systems}, 36:\penalty0 79858--79885, 2023.

\bibitem[Yu and Yu(2026)]{yu2026advancing}
Yifeng Yu and Lu~Yu.
\newblock Advancing wasserstein convergence analysis of score-based models: Insights from discretization and second-order acceleration.
\newblock \emph{Advances in Neural Information Processing Systems}, 38:\penalty0 138411--138465, 2026.

\bibitem[Zheng et~al.(2025)Zheng, Huang, Liu, Zou, He, Zhang, Gu, Zhang, He, Zheng, et~al.]{zheng2025vbench}
Dian Zheng, Ziqi Huang, Hongbo Liu, Kai Zou, Yinan He, Fan Zhang, Lulu Gu, Yuanhan Zhang, Jingwen He, Wei-Shi Zheng, et~al.
\newblock Vbench-2.0: Advancing video generation benchmark suite for intrinsic faithfulness.
\newblock \emph{arXiv preprint arXiv:2503.21755}, 2025.

\bibitem[AI et~al.(2025)AI, Gong, Zou, Zheng, Zhou, Yan, Jin, Shen, Zheng, Wang, et~al.]{ai2025ming}
Inclusion AI, Biao Gong, Cheng Zou, Chuanyang Zheng, Chunluan Zhou, Canxiang Yan, Chunxiang Jin, Chunjie Shen, Dandan Zheng, Fudong Wang, et~al.
\newblock Ming-omni: A unified multimodal model for perception and generation.
\newblock \emph{arXiv preprint arXiv:2506.09344}, 2025.

\bibitem[Xie et~al.(2026)Xie, Xiao, Liu, Huang, Zheng, and Huang]{xie2026emergentbridge}
Jincheng Xie, Xingchen Xiao, Runheng Liu, Zhongyi Huang, Yu~Zheng, and Heyan Huang.
\newblock Emergentbridge: Improving zero-shot cross-modal transfer in unified multimodal embedding models.
\newblock \emph{arXiv preprint arXiv:2604.11043}, 2026.

\bibitem[Yu et~al.(2024{\natexlab{c}})Yu, Lezama, Gundavarapu, Versari, Sohn, Minnen, Cheng, Gupta, Gu, Hauptmann, et~al.]{Yu2023LanguageMB}
Lijun Yu, Jos{\'e} Lezama, Nitesh~Bharadwaj Gundavarapu, Luca Versari, Kihyuk Sohn, David Minnen, Yong Cheng, Agrim Gupta, Xiuye Gu, Alexander~G Hauptmann, et~al.
\newblock Language model beats diffusion-tokenizer is key to visual generation.
\newblock In \emph{International Conference on Learning Representations}, volume 2024, pages 765--783, 2024{\natexlab{c}}.

\bibitem[Gafni et~al.(2022)Gafni, Polyak, Ashual, Sheynin, Parikh, and Taigman]{Gafni2022MakeASceneST}
Oran Gafni, Adam Polyak, Oron Ashual, Shelly Sheynin, Devi Parikh, and Yaniv Taigman.
\newblock Make-a-scene: Scene-based text-to-image generation with human priors.
\newblock \emph{ArXiv}, abs/2203.13131, 2022.

\bibitem[D'efossez et~al.(2022)D'efossez, Copet, Synnaeve, and Adi]{Defossez2022HighFN}
Alexandre D'efossez, Jade Copet, Gabriel Synnaeve, and Yossi Adi.
\newblock High fidelity neural audio compression.
\newblock \emph{ArXiv}, abs/2210.13438, 2022.

\bibitem[Wu et~al.(2024{\natexlab{a}})Wu, Zhu, Liu, Zhao, Zhai, Cao, and Zha]{Wu2024ImprovedVV}
Ping Wu, Kai Zhu, Yu~Liu, Liming Zhao, Wei Zhai, Yang Cao, and Zhengjun Zha.
\newblock Improved video vae for latent video diffusion model.
\newblock \emph{2025 IEEE/CVF Conference on Computer Vision and Pattern Recognition (CVPR)}, pages 18124--18133, 2024{\natexlab{a}}.

\bibitem[Peng et~al.(2025)Peng, Zheng, Shen, Young, Guo, Wang, Xu, Liu, Jiang, Li, Wang, Ye, Ren, Ma, Liang, Lian, Wu, Zhong, Li, Gong, Lei, Cheng, Zhang, Li, Zhang, Hu, Huang, Wang, Zhao, Wang, Wei, and You]{Peng2025OpenSora2T}
Xiangyu Peng, Zangwei Zheng, Chenhui Shen, Tom Young, Xinying Guo, Binluo Wang, Hang Xu, Hongxin Liu, Mingyang Jiang, Wenjun Li, Yuhui Wang, Anbang Ye, Guanjun Ren, Qianran Ma, Wanying Liang, Xiangru Lian, Xiwen Wu, Yu~Zhong, Zhuangyan Li, Chaoyu Gong, Guojun Lei, Lei Cheng, Liming Zhang, Minghao Li, Ruijie Zhang, Silan Hu, Shijie Huang, Xiaokang Wang, Yuanheng Zhao, Yuqi Wang, Ziang Wei, and Yang You.
\newblock Open-sora 2.0: Training a commercial-level video generation model in \$200k.
\newblock \emph{ArXiv}, abs/2503.09642, 2025.

\bibitem[Schulman et~al.(2017)Schulman, Wolski, Dhariwal, Radford, and Klimov]{schulman2017proximalpolicyoptimizationalgorithms}
John Schulman, Filip Wolski, Prafulla Dhariwal, Alec Radford, and Oleg Klimov.
\newblock Proximal policy optimization algorithms.
\newblock \emph{arXiv preprint arXiv:1707.06347}, 2017.

\bibitem[Shao et~al.(2024)Shao, Wang, Zhu, Xu, Song, Bi, Zhang, Zhang, Li, Wu, et~al.]{shao2024deepseekmath}
Zhihong Shao, Peiyi Wang, Qihao Zhu, Runxin Xu, Junxiao Song, Xiao Bi, Haowei Zhang, Mingchuan Zhang, YK~Li, Yang Wu, et~al.
\newblock Deepseekmath: Pushing the limits of mathematical reasoning in open language models.
\newblock \emph{arXiv preprint arXiv:2402.03300}, 2024.

\bibitem[Xu et~al.(2024{\natexlab{a}})Xu, Fu, Gao, Ye, Liu, Mei, Wang, Yu, and Wu]{xu2024dposuperiorppollm}
Shusheng Xu, Wei Fu, Jiaxuan Gao, Wenjie Ye, Weilin Liu, Zhiyu Mei, Guangju Wang, Chao Yu, and Yi~Wu.
\newblock Is dpo superior to ppo for llm alignment? a comprehensive study.
\newblock \emph{arXiv preprint arXiv:2404.10719}, 2024{\natexlab{a}}.

\bibitem[Song et~al.(2024)Song, Swamy, Singh, Bagnell, and Sun]{song2024importanceonlinedataunderstanding}
Yuda Song, Gokul Swamy, Aarti Singh, J.~Andrew Bagnell, and Wen Sun.
\newblock The importance of online data: Understanding preference fine-tuning via coverage.
\newblock \emph{arXiv preprint arXiv:2406.01462}, 2024.

\bibitem[Li et~al.(2025{\natexlab{c}})Li, Wen, Lou, Ji, Lu, Han, Zhang, and Sun]{li2025devildetailstacklingunimodal}
Zichao Li, Xueru Wen, Jie Lou, Yuqiu Ji, Yaojie Lu, Xianpei Han, Debing Zhang, and Le~Sun.
\newblock The devil is in the details: Tackling unimodal spurious correlations for generalizable multimodal reward models.
\newblock \emph{arXiv preprint arXiv:2503.03122}, 2025{\natexlab{c}}.

\bibitem[Zhang et~al.(2025{\natexlab{b}})Zhang, Yan, Zheng, Zou, Dai, and Hu]{zhang2025gmprmgenerativemultimodalprocess}
Jianghangfan Zhang, Yibo Yan, Kening Zheng, Xin Zou, Song Dai, and Xuming Hu.
\newblock Gm-prm: A generative multimodal process reward model for multimodal mathematical reasoning.
\newblock \emph{arXiv preprint arXiv:2508.04088}, 2025{\natexlab{b}}.

\bibitem[Jiang et~al.(2025{\natexlab{a}})Jiang, Guo, Zhang, Zong, Li, Zhuo, Yan, Heng, and Li]{jiang2025t2ir1reinforcingimagegeneration}
Dongzhi Jiang, Ziyu Guo, Renrui Zhang, Zhuofan Zong, Hao Li, Le~Zhuo, Shilin Yan, Pheng-Ann Heng, and Hongsheng Li.
\newblock T2i-r1: Reinforcing image generation with collaborative semantic-level and token-level cot.
\newblock \emph{arXiv preprint arXiv:2505.00703}, 2025{\natexlab{a}}.

\bibitem[Liu et~al.(2025{\natexlab{a}})Liu, Liu, Liang, Li, Liu, Wang, Wan, Zhang, and Ouyang]{liu2025flowgrpotrainingflowmatching}
Jie Liu, Gongye Liu, Jiajun Liang, Yangguang Li, Jiaheng Liu, Xintao Wang, Pengfei Wan, Di~Zhang, and Wanli Ouyang.
\newblock Flow-grpo: Training flow matching models via online rl.
\newblock \emph{arXiv preprint arXiv:2505.05470}, 2025{\natexlab{a}}.

\bibitem[Wang et~al.(2024{\natexlab{a}})Wang, Zhou, Huang, Xu, Zhang, Poon, and Chen]{wang2024mdpoconditionalpreferenceoptimization}
Fei Wang, Wenxuan Zhou, James~Y. Huang, Nan Xu, Sheng Zhang, Hoifung Poon, and Muhao Chen.
\newblock mdpo: Conditional preference optimization for multimodal large language models.
\newblock \emph{arXiv preprint arXiv:2406.11839}, 2024{\natexlab{a}}.

\bibitem[Rao and Rachuri(2026)]{rao2026understandinggenerationfightdiagnostic}
Abinav Rao and Sujan Rachuri.
\newblock Do understanding and generation fight? a diagnostic study of dpo for unified multimodal models.
\newblock \emph{arXiv preprint arXiv:2603.17044}, 2026.

\bibitem[Mao et~al.(2025)Mao, Yang, and Shou]{mao2025unirlselfimprovingunifiedmultimodal}
Weijia Mao, Zhenheng Yang, and Mike~Zheng Shou.
\newblock Unirl: Self-improving unified multimodal models via supervised and reinforcement learning.
\newblock \emph{arXiv preprint arXiv:2505.23380}, 2025.

\bibitem[Sun et~al.(2023{\natexlab{c}})Sun, Shen, Cao, Liu, Li, Shen, Gan, Gui, Wang, Yang, Keutzer, and Darrell]{sun2023aligninglargemultimodalmodels}
Zhiqing Sun, Sheng Shen, Shengcao Cao, Haotian Liu, Chunyuan Li, Yikang Shen, Chuang Gan, Liang-Yan Gui, Yu-Xiong Wang, Yiming Yang, Kurt Keutzer, and Trevor Darrell.
\newblock Aligning large multimodal models with factually augmented rlhf.
\newblock \emph{arXiv preprint arXiv:2309.14525}, 2023{\natexlab{c}}.

\bibitem[Luo et~al.(2025)Luo, Zheng, Wang, Ni, Lin, Jiang, Yu, Shi, Wang, Chu, Zeng, and Yang]{luo2025unlockingmultimodalmathematicalreasoning}
Ruilin Luo, Zhuofan Zheng, Yifan Wang, Xinzhe Ni, Zicheng Lin, Songtao Jiang, Yiyao Yu, Chufan Shi, Lei Wang, Ruihang Chu, Jin Zeng, and Yujiu Yang.
\newblock Unlocking multimodal mathematical reasoning via process reward model.
\newblock \emph{arXiv preprint arXiv:2501.04686}, 2025.

\bibitem[Team et~al.(2024)Team, Georgiev, Lei, Burnell, Bai, Gulati, Tanzer, Vincent, Pan, Wang, et~al.]{gemini2024gemini15}
Gemini Team, Petko Georgiev, Ving~Ian Lei, Ryan Burnell, Libin Bai, Anmol Gulati, Garrett Tanzer, Damien Vincent, Zhufeng Pan, Shibo Wang, et~al.
\newblock Gemini 1.5: Unlocking multimodal understanding across millions of tokens of context.
\newblock \emph{arXiv preprint arXiv:2403.05530}, 2024.

\bibitem[{Google DeepMind}(2025{\natexlab{a}})]{gemini2025gemini25}
{Google DeepMind}.
\newblock {Gemini 2.5}: Pushing the frontier with advanced reasoning, multimodality, long context, and next generation agentic capabilities.
\newblock \emph{arXiv preprint arXiv:2507.06261}, 2025{\natexlab{a}}.

\bibitem[Shao et~al.(2026)Shao, Tao, Zhang, Feng, Cai, Shang, You, Qin, Sui, and Wang]{shao2026surveytokencompressionefficient}
Kele Shao, Keda Tao, Kejia Zhang, Sicheng Feng, Mu~Cai, Yuzhang Shang, Haoxuan You, Can Qin, Yang Sui, and Huan Wang.
\newblock A survey of token compression for efficient multimodal large language models.
\newblock \emph{arXiv preprint arXiv:2507.20198}, 2026.

\bibitem[{Google DeepMind}(2025{\natexlab{b}})]{gemma2025gemma3}
{Google DeepMind}.
\newblock {Gemma 3} technical report.
\newblock \emph{arXiv preprint arXiv:2503.19786}, 2025{\natexlab{b}}.

\bibitem[Yang et~al.(2025{\natexlab{b}})Yang, Chen, Tian, Wang, Li, Yu, and Jia]{visionzip2025}
Senqiao Yang, Yukang Chen, Zhuotao Tian, Chengyao Wang, Jingyao Li, Bei Yu, and Jiaya Jia.
\newblock {VisionZip}: Longer is better but not necessary in vision language models.
\newblock \emph{arXiv preprint arXiv:2412.04467}, 2025{\natexlab{b}}.

\bibitem[Zhang et~al.(2025{\natexlab{c}})Zhang, Fan, Ma, Zheng, Huang, Cheng, Gudovskiy, Okuno, Nakata, Keutzer, and Zhang]{sparsevlm2025}
Yuan Zhang, Chun-Kai Fan, Junpeng Ma, Wenzhao Zheng, Tao Huang, Kuan Cheng, Denis Gudovskiy, Tomoyuki Okuno, Yohei Nakata, Kurt Keutzer, and Shanghang Zhang.
\newblock {SparseVLM}: Visual token sparsification for efficient vision-language model inference.
\newblock \emph{arXiv preprint arXiv:2410.04417}, 2025{\natexlab{c}}.

\bibitem[Ye et~al.(2025)Ye, Wu, Lin, and Zhou]{fitprune2025}
Weihao Ye, Qiong Wu, Wenhao Lin, and Yiyi Zhou.
\newblock Fit and prune: Fast and training-free visual token pruning for multi-modal large language models.
\newblock \emph{arXiv preprint arXiv:2409.10197}, 2025.

\bibitem[Shang et~al.(2024)Shang, Cai, Xu, Lee, and Yan]{shang2024llavaprumerge}
Yuzhang Shang, Mu~Cai, Bingxin Xu, Yong~Jae Lee, and Yan Yan.
\newblock {LLaVA-PruMerge}: Adaptive token reduction for efficient large multimodal models.
\newblock \emph{arXiv preprint arXiv:2403.15388}, 2024.

\bibitem[Zhu et~al.(2025)Zhu, Zhu, Lu, Yan, Li, Liu, Fu, and Zha]{visionselector2025}
Jiaying Zhu, Yurui Zhu, Xin Lu, Wenrui Yan, Dong Li, Kunlin Liu, Xueyang Fu, and Zheng-Jun Zha.
\newblock {VisionSelector}: End-to-end learnable visual token compression for efficient multimodal {LLMs}.
\newblock \emph{arXiv preprint arXiv:2510.16598}, 2025.

\bibitem[Liu et~al.(2025{\natexlab{b}})Liu, Niu, Chen, Zhou, and Meng]{laco2025}
Juntao Liu, Liqiang Niu, Wenchao Chen, Jie Zhou, and Fandong Meng.
\newblock {LaCo}: Efficient layer-wise compression of visual tokens for multimodal large language models.
\newblock \emph{arXiv preprint arXiv:2507.02279}, 2025{\natexlab{b}}.

\bibitem[Wang et~al.(2024{\natexlab{b}})Wang, Bai, Tan, Wang, Fan, Bai, Chen, Liu, Wang, Ge, et~al.]{wang2024qwen2}
Peng Wang, Shuai Bai, Sinan Tan, Shijie Wang, Zhihao Fan, Jinze Bai, Keqin Chen, Xuejing Liu, Jialin Wang, Wenbin Ge, et~al.
\newblock Qwen2-vl: Enhancing vision-language model's perception of the world at any resolution.
\newblock \emph{arXiv preprint arXiv:2409.12191}, 2024{\natexlab{b}}.

\bibitem[Xu et~al.(2024{\natexlab{b}})Xu, Yao, Guo, Cui, Ni, Ge, Chua, Liu, Sun, and Huang]{llavauhd2024}
Ruyi Xu, Yuan Yao, Zonghao Guo, Junbo Cui, Zanlin Ni, Chunjiang Ge, Tat-Seng Chua, Zhiyuan Liu, Maosong Sun, and Gao Huang.
\newblock {LLaVA-UHD}: an {LMM} perceiving any aspect ratio and high-resolution images.
\newblock \emph{arXiv preprint arXiv:2403.11703}, 2024{\natexlab{b}}.

\bibitem[Li et~al.(2024{\natexlab{b}})Li, Zhang, Guo, Zhang, Li, Zhang, Zhang, Li, Liu, and Li]{llavaonevision2024}
Bo~Li, Yuanhan Zhang, Dong Guo, Renrui Zhang, Feng Li, Hao Zhang, Kaichen Zhang, Yanwei Li, Ziwei Liu, and Chunyuan Li.
\newblock {LLaVA-OneVision}: Easy visual task transfer.
\newblock \emph{arXiv preprint arXiv:2408.03326}, 2024{\natexlab{b}}.

\bibitem[Liu et~al.(2024{\natexlab{b}})Liu, Dong, Liu, Hu, Lu, and Rao]{oryx2024}
Zuyan Liu, Yuhao Dong, Ziwei Liu, Winston Hu, Jiwen Lu, and Yongming Rao.
\newblock {Oryx MLLM}: On-demand spatial-temporal understanding at arbitrary resolution.
\newblock \emph{arXiv preprint arXiv:2409.12961}, 2024{\natexlab{b}}.

\bibitem[Chen et~al.(2024{\natexlab{b}})Chen, Wang, Cao, Liu, Gao, Cui, Zhu, Ye, Tian, Liu, et~al.]{internvl25_2024}
Zhe Chen, Weiyun Wang, Yue Cao, Yangzhou Liu, Zhangwei Gao, Erfei Cui, Jinguo Zhu, Shenglong Ye, Hao Tian, Zhaoyang Liu, et~al.
\newblock Expanding performance boundaries of open-source multimodal models with model, data, and test-time scaling.
\newblock \emph{arXiv preprint arXiv:2412.05271}, 2024{\natexlab{b}}.

\bibitem[Shi et~al.(2026{\natexlab{b}})Shi, Pei, Wen, Dong, and Xu]{shi2026q}
Yuheng Shi, Xiaohuan Pei, Linfeng Wen, Minjing Dong, and Chang Xu.
\newblock Q-zoom: Query-aware adaptive perception for efficient multimodal large language models.
\newblock \emph{arXiv preprint arXiv:2604.06912}, 2026{\natexlab{b}}.

\bibitem[Zhao et~al.(2026)Zhao, Zhang, Guo, Hu, Duan, Fu, Chng, Wang, Chen, Xu, Luo, and Zhang]{zhao2026unifiedmultimodalunderstandinggeneration}
Shanshan Zhao, Xinjie Zhang, Jintao Guo, Jiakui Hu, Lunhao Duan, Minghao Fu, Yong~Xien Chng, Guo-Hua Wang, Qing-Guo Chen, Zhao Xu, Weihua Luo, and Kaifu Zhang.
\newblock Unified multimodal understanding and generation models: Advances, challenges, and opportunities.
\newblock \emph{arXiv preprint arXiv:2505.02567}, 2026.

\bibitem[Zhang et~al.(2024{\natexlab{c}})Zhang, Zhong, Jiang, Hu, Sun, Ge, Zhu, Jiang, and Jin]{10.1145/3718958.3750472}
Zili Zhang, Yinmin Zhong, Yimin Jiang, Hanpeng Hu, Jian‐Yuan Sun, Zheng Ge, Yibo Zhu, Daxin Jiang, and Xin Jin.
\newblock Disttrain: Addressing model and data heterogeneity with disaggregated training for multimodal large language models.
\newblock \emph{Proceedings of the ACM SIGCOMM 2025 Conference}, 2024{\natexlab{c}}.

\bibitem[Xu et~al.(2022)Xu, Zhu, and Clifton]{10123038}
Peng Xu, Xiatian Zhu, and David~A. Clifton.
\newblock Multimodal learning with transformers: A survey.
\newblock \emph{IEEE Transactions on Pattern Analysis and Machine Intelligence}, 45:\penalty0 12113--12132, 2022.

\bibitem[Liang et~al.(2025{\natexlab{a}})Liang, Tian, Yin, Yua, An-Hou, Ming, Song, Wang, Bi, and Liu]{liang2025comprehensivesurveyguidemultimodal}
Chia~Xin Liang, Pu~Tian, Caitlyn~Heqi Yin, Yao Yua, Wei An-Hou, Li~Ming, Xinyuan Song, Tianyang Wang, Ziqian Bi, and Ming Liu.
\newblock A comprehensive survey and guide to multimodal large language models in vision-language tasks.
\newblock \emph{arXiv preprint arXiv:2411.06284}, 2025{\natexlab{a}}.

\bibitem[Kong et~al.(2026)Kong, Li, Zeng, Xin, Messica, Lin, Zhao, Kellis, Tang, and Zitnik]{kong2026tokenreductionefficiencygenerative}
Zhenglun Kong, Yize Li, Fanhu Zeng, Lei Xin, Shvat Messica, Xue Lin, Pu~Zhao, Manolis Kellis, Hao Tang, and Marinka Zitnik.
\newblock Token reduction should go beyond efficiency in generative models -- from vision, language to multimodality.
\newblock \emph{arXiv preprint arXiv:2505.18227}, 2026.

\bibitem[Li et~al.(2025{\natexlab{d}})Li, Fu, Liu, Cao, Ji, Yang, King, and Yang]{11455337}
Jindong Li, Yali Fu, Jiahong Liu, Linxiao Cao, Wei Ji, Menglin Yang, Irwin King, and Mingxue Yang.
\newblock Discrete tokenization for multimodal llms: A comprehensive survey.
\newblock \emph{ArXiv}, abs/2507.22920, 2025{\natexlab{d}}.

\bibitem[Mousavi et~al.(2025)Mousavi, Maimon, Moumen, Petermann, Shi, Wu, Yang, Kuznetsova, Ploujnikov, Marxer, Ramabhadran, Elizalde, Lugosch, Li, Subakan, Woodland, Kim, yi~Lee, Watanabe, Adi, and Ravanelli]{mousavi2025discreteaudiotokenssurvey}
Pooneh Mousavi, Gallil Maimon, Adel Moumen, Darius Petermann, Jiatong Shi, Haibin Wu, Haici Yang, Anastasia Kuznetsova, Artem Ploujnikov, Ricard Marxer, Bhuvana Ramabhadran, Benjamin Elizalde, Loren Lugosch, Jinyu Li, Cem Subakan, Phil Woodland, Minje Kim, Hung yi~Lee, Shinji Watanabe, Yossi Adi, and Mirco Ravanelli.
\newblock Discrete audio tokens: More than a survey!
\newblock \emph{arXiv preprint arXiv:2506.10274}, 2025.

\bibitem[Cai et~al.(2024)Cai, Jiang, Wang, Tang, Kim, and Huang]{10937907}
Weilin Cai, Juyong Jiang, Fan Wang, Jing Tang, Sunghun Kim, and Jiayi Huang.
\newblock A survey on mixture of experts in large language models.
\newblock \emph{IEEE Transactions on Knowledge and Data Engineering}, 37:\penalty0 3896--3915, 2024.

\bibitem[Dai et~al.(2024)Dai, Deng, Zhao, Xu, Gao, Chen, Li, Zeng, Yu, Wu, Xie, Li, Huang, Luo, Ruan, Sui, and Liang]{dai-etal-2024-deepseekmoe}
Damai Dai, Chengqi Deng, Chenggang Zhao, Runxin Xu, Huazuo Gao, Deli Chen, Jiashi Li, Wangding Zeng, Xingkai Yu, Yu~Wu, Zhenda Xie, Y.~K. Li, Panpan Huang, Fuli Luo, Chong Ruan, Zhifang Sui, and Wenfeng Liang.
\newblock Deepseekmoe: Towards ultimate expert specialization in mixture-of-experts language models.
\newblock In \emph{Annual Meeting of the Association for Computational Linguistics}, 2024.

\bibitem[Yin et~al.(2025)Yin, Zhang, Zhang, Freeman, Durand, Shechtman, and Huang]{Yin_2025_CVPR}
Tianwei Yin, Qiang Zhang, Richard Zhang, William~T. Freeman, Fredo Durand, Eli Shechtman, and Xun Huang.
\newblock From slow bidirectional to fast autoregressive video diffusion models.
\newblock In \emph{Proceedings of the IEEE/CVF Conference on Computer Vision and Pattern Recognition (CVPR)}, pages 22963--22974, June 2025.

\bibitem[Chen et~al.(2024{\natexlab{c}})Chen, Wang, Ren, Li, Zhao, Li, Cai, Guo, Zhang, Xiong, Zhang, Wu, Dong, Zhang, Yang, Meng, Hu, Chen, Lin, Bai, Vlachos, Tan, Zhang, Xiao, Yee, Liu, and Chang]{chen2024tokenpredictionmultimodalintelligence}
Liang Chen, Zekun Wang, Shuhuai Ren, Lei Li, Haozhe Zhao, Yunshui Li, Zefan Cai, Hongcheng Guo, Lei Zhang, Yizhe Xiong, Yichi Zhang, Ruoyu Wu, Qingxiu Dong, Ge~Zhang, Jian Yang, Lingwei Meng, Shujie Hu, Yulong Chen, Junyang Lin, Shuai Bai, Andreas Vlachos, Xu~Tan, Minjia Zhang, Wen Xiao, Aaron Yee, Tianyu Liu, and Baobao Chang.
\newblock Next token prediction towards multimodal intelligence: A comprehensive survey.
\newblock \emph{arXiv preprint arXiv:2412.18619}, 2024{\natexlab{c}}.

\bibitem[Dao et~al.(2022)Dao, Fu, Ermon, Rudra, and Ré]{dao2022flashattentionfastmemoryefficientexact}
Tri Dao, Daniel~Y. Fu, Stefano Ermon, Atri Rudra, and Christopher Ré.
\newblock Flashattention: Fast and memory-efficient exact attention with io-awareness.
\newblock \emph{arXiv preprint arXiv:2205.14135}, 2022.

\bibitem[Dao(2024)]{dao2023flashattention2}
Tri Dao.
\newblock Flash{A}ttention-2: Faster attention with better parallelism and work partitioning.
\newblock In \emph{International Conference on Learning Representations (ICLR)}, 2024.

\bibitem[Dong et~al.(2024{\natexlab{d}})Dong, Feng, Guessous, Liang, and He]{dong2024flexattentionprogrammingmodel}
Juechu Dong, Boyuan Feng, Driss Guessous, Yanbo Liang, and Horace He.
\newblock Flex attention: A programming model for generating optimized attention kernels.
\newblock \emph{arXiv preprint arXiv:2412.05496}, 2024{\natexlab{d}}.

\bibitem[Liu et~al.(2025{\natexlab{c}})Liu, Yue, Zhang, Sun, Zhou, Zeng, Tang, and Zhou]{liu2025efficienttrainingdiffusionmixtureofexperts}
Yahui Liu, Yang Yue, Jingyuan Zhang, Chenxi Sun, Yang Zhou, Wencong Zeng, Ruiming Tang, and Guorui Zhou.
\newblock Efficient training of diffusion mixture-of-experts models: A practical recipe.
\newblock \emph{arXiv preprint arXiv:2512.01252}, 2025{\natexlab{c}}.

\bibitem[Wang et~al.(2025{\natexlab{c}})Wang, Zeng, Xiao, Wu, Yang, Zheng, Chen, Bian, Yu, and Wang]{wang2025flashmaskefficientrichmask}
Guoxia Wang, Jinle Zeng, Xiyuan Xiao, Siming Wu, Jiabin Yang, Lujing Zheng, Zeyu Chen, Jiang Bian, Dianhai Yu, and Haifeng Wang.
\newblock Flashmask: Efficient and rich mask extension of flashattention.
\newblock \emph{arXiv preprint arXiv:2410.01359}, 2025{\natexlab{c}}.

\bibitem[Hu et~al.(2026)Hu, Zhang, Huang, Yi, Su, Weng, Xue, Ma, Yang, and Tao]{hu2026evolutionvideogenerativefoundations}
Teng Hu, Jiangning Zhang, Hongrui Huang, Ran Yi, Zihan Su, Jieyu Weng, Zhucun Xue, Lizhuang Ma, Ming-Hsuan Yang, and Dacheng Tao.
\newblock Evolution of video generative foundations.
\newblock \emph{arXiv preprint arXiv:2604.06339}, 2026.

\bibitem[Fang et~al.(2025)Fang, Zhou, Guo, Zhang, and Feng]{fang2025llama}
Qingkai Fang, Yan Zhou, Shoutao Guo, Shaolei Zhang, and Yang Feng.
\newblock Llama-omni 2: Llm-based real-time spoken chatbot with autoregressive streaming speech synthesis.
\newblock In \emph{Proceedings of the 63rd Annual Meeting of the Association for Computational Linguistics (Volume 1: Long Papers)}, pages 18617--18629, 2025.

\bibitem[Xu et~al.(2025{\natexlab{b}})Xu, Guo, Hu, Chu, Wang, He, Wang, Shi, He, Zhu, et~al.]{xu2025qwen3}
Jin Xu, Zhifang Guo, Hangrui Hu, Yunfei Chu, Xiong Wang, Jinzheng He, Yuxuan Wang, Xian Shi, Ting He, Xinfa Zhu, et~al.
\newblock Qwen3-omni technical report.
\newblock \emph{arXiv preprint arXiv:2509.17765}, 2025{\natexlab{b}}.

\bibitem[Cheng et~al.(2026)Cheng, Yuan, Li, You, Wang, Nie, Zhang, and Li]{cheng2026ar}
Dongjie Cheng, Ruifeng Yuan, Yongqi Li, Runyang You, Wenjie Wang, Liqiang Nie, Lei Zhang, and Wenjie Li.
\newblock Ar-omni: A unified autoregressive model for any-to-any generation.
\newblock \emph{arXiv preprint arXiv:2601.17761}, 2026.

\bibitem[Bruce et~al.(2024)Bruce, Dennis, Edwards, Parker-Holder, Shi, Hughes, Lai, Mavalankar, Steigerwald, Apps, et~al.]{bruce2024genie}
Jake Bruce, Michael~D Dennis, Ashley Edwards, Jack Parker-Holder, Yuge Shi, Edward Hughes, Matthew Lai, Aditi Mavalankar, Richie Steigerwald, Chris Apps, et~al.
\newblock Genie: Generative interactive environments.
\newblock In \emph{Forty-first International Conference on Machine Learning}, 2024.

\bibitem[Pang et~al.(2026)Pang, Jin, Yang, Zhu, Lin, Feng, Tang, Chen, Tay, Lim, et~al.]{pang2026next}
Yatian Pang, Peng Jin, Shuo Yang, Bin Zhu, Bin Lin, Chaoran Feng, Zhenyu Tang, Liuhan Chen, Francis~EH Tay, Ser-Nam Lim, et~al.
\newblock Next patch prediction for autoregressive visual generation.
\newblock In \emph{Proceedings of the AAAI Conference on Artificial Intelligence}, volume~40, pages 8260--8268, 2026.

\bibitem[Ren et~al.(2025)Ren, Ma, Sun, and Wei]{ren2025next}
Shuhuai Ren, Shuming Ma, Xu~Sun, and Furu Wei.
\newblock Next block prediction: Video generation via semi-autoregressive modeling.
\newblock \emph{arXiv preprint arXiv:2502.07737}, 2025.

\bibitem[Lan et~al.(2024)Lan, Niu, Meng, Li, Zhou, and Su]{lan2024avg}
Zhibin Lan, Liqiang Niu, Fandong Meng, Wenbo Li, Jie Zhou, and Jinsong Su.
\newblock Avg-llava: An efficient large multimodal model with adaptive visual granularity.
\newblock \emph{arXiv preprint arXiv:2410.02745}, 2024.

\bibitem[Lin et~al.(2025{\natexlab{a}})Lin, Liu, Yang, Tao, and Ye]{lin2025adaptvision}
Zichuan Lin, Yicheng Liu, Yang Yang, Lvfang Tao, and Deheng Ye.
\newblock Adaptvision: Efficient vision-language models via adaptive visual acquisition.
\newblock \emph{arXiv preprint arXiv:2512.03794}, 2025{\natexlab{a}}.

\bibitem[Liao et~al.(2026)Liao, Jiang, Hao, Tan, He, Zhao, Xu, and Liu]{liao2026resadapt}
Huanxuan Liao, Zhongtao Jiang, Yupu Hao, Yuqiao Tan, Shizhu He, Jun Zhao, Kun Xu, and Kang Liu.
\newblock Resadapt: Adaptive resolution for efficient multimodal reasoning.
\newblock \emph{arXiv preprint arXiv:2603.28610}, 2026.

\bibitem[Ma et~al.(2025{\natexlab{b}})Ma, Song, Du, Cong, Chen, Wang, Wang, and Chen]{ma2025language}
Ziyang Ma, Yakun Song, Chenpeng Du, Jian Cong, Zhuo Chen, Yuping Wang, Yuxuan Wang, and Xie Chen.
\newblock Language model can listen while speaking.
\newblock In \emph{Proceedings of the AAAI Conference on Artificial Intelligence}, volume~39, pages 24831--24839, 2025{\natexlab{b}}.

\bibitem[Zhang et~al.(2025{\natexlab{d}})Zhang, Li, Chen, Kothapally, Yu, and Yu]{zhang2025llm}
Hao Zhang, Weiwei Li, Rilin Chen, Vinay Kothapally, Meng Yu, and Dong Yu.
\newblock Llm-enhanced dialogue management for full-duplex spoken dialogue systems.
\newblock \emph{arXiv preprint arXiv:2502.14145}, 2025{\natexlab{d}}.

\bibitem[Chen et~al.(2025{\natexlab{b}})Chen, Hu, Li, Li, Liu, Li, Li, Li, Shen, Tang, et~al.]{chen2025fireredchat}
Junjie Chen, Yao Hu, Junjie Li, Kangyue Li, Kun Liu, Wenpeng Li, Xu~Li, Ziyuan Li, Feiyu Shen, Xu~Tang, et~al.
\newblock Fireredchat: A pluggable, full-duplex voice interaction system with cascaded and semi-cascaded implementations.
\newblock \emph{arXiv preprint arXiv:2509.06502}, 2025{\natexlab{b}}.

\bibitem[Wang et~al.(2025{\natexlab{d}})Wang, Yu, Chen, Tian, Zhang, Lu, and Zhang]{wang2025end}
Siyin Wang, Wenyi Yu, Xianzhao Chen, Xiaohai Tian, Jun Zhang, Lu~Lu, and Chao Zhang.
\newblock End-to-end listen, look, speak and act.
\newblock \emph{arXiv preprint arXiv:2510.16756}, 2025{\natexlab{d}}.

\bibitem[Zhang et~al.(2026{\natexlab{b}})Zhang, Tong, Lin, Wu, Sun, Ma, and Shen]{zhang2026think}
Jialiang Zhang, Junlong Tong, Junyan Lin, Hao Wu, Yirong Sun, Yunpu Ma, and Xiaoyu Shen.
\newblock Think-as-you-see: Streaming chain-of-thought reasoning for large vision-language models.
\newblock \emph{arXiv preprint arXiv:2603.02872}, 2026{\natexlab{b}}.

\bibitem[Lu et~al.(2026)Lu, Bo, Chen, Li, Guo, Guan, Liu, Xu, Sun, Sun, et~al.]{lu2026aura}
Xudong Lu, Yang Bo, Jinpeng Chen, Shuhan Li, Xintong Guo, Huankang Guan, Fang Liu, Dunyuan Xu, Peiwen Sun, Heyang Sun, et~al.
\newblock Aura: Always-on understanding and real-time assistance via video streams.
\newblock \emph{arXiv preprint arXiv:2604.04184}, 2026.

\bibitem[Ning et~al.(2024)Ning, Zhao, Jin, Ding, and Guo]{ning2024inf}
Zhenyu Ning, Jieru Zhao, Qihao Jin, Wenchao Ding, and Minyi Guo.
\newblock Inf-mllm: Efficient streaming inference of multimodal large language models on a single gpu.
\newblock \emph{arXiv preprint arXiv:2409.09086}, 2024.

\bibitem[Jiang et~al.(2025{\natexlab{b}})Jiang, Tu, Zhang, Dong, Wang, Yang, and Pan]{jiang2025m4sc}
Feibo Jiang, Siwei Tu, Jin Zhang, Li~Dong, Kezhi Wang, Kun Yang, and Cunhua Pan.
\newblock M4sc: An mllm-based multi-modal, multi-task and multi-user semantic communication system.
\newblock \emph{IEEE Wireless Communications}, 32\penalty0 (5):\penalty0 40--47, 2025{\natexlab{b}}.

\bibitem[Qiao et~al.(2025)Qiao, Mashhadi, Gao, Tafazolli, Bennis, and Niyato]{qiao2025token}
Li~Qiao, Mahdi~Boloursaz Mashhadi, Zhen Gao, Rahim Tafazolli, Mehdi Bennis, and Dusit Niyato.
\newblock Token communications: A large model-driven framework for cross-modal context-aware semantic communications.
\newblock \emph{IEEE Wireless Communications}, 32\penalty0 (5):\penalty0 80--88, 2025.

\bibitem[Yu et~al.(2025{\natexlab{c}})Yu, Zhou, Yang, Li, Wang, Hu, Xu, Xu, Shu, and Yuan]{yu2025mquant}
JiangYong Yu, Sifan Zhou, Dawei Yang, Shuoyu Li, Shuo Wang, Xing Hu, Chen Xu, Zukang Xu, Changyong Shu, and Zhihang Yuan.
\newblock Mquant: Unleashing the inference potential of multimodal large language models via static quantization.
\newblock In \emph{Proceedings of the 33rd ACM International Conference on Multimedia}, pages 1783--1792, 2025{\natexlab{c}}.

\bibitem[Li et~al.(2025{\natexlab{e}})Li, Hu, Ning, Liu, Hong, Jia, Li, Yan, Ran, Dai, et~al.]{li2025mbq}
Shiyao Li, Yingchun Hu, Xuefei Ning, Xihui Liu, Ke~Hong, Xiaotao Jia, Xiuhong Li, Yaqi Yan, Pei Ran, Guohao Dai, et~al.
\newblock Mbq: Modality-balanced quantization for large vision-language models.
\newblock In \emph{Proceedings of the Computer Vision and Pattern Recognition Conference}, pages 4167--4177, 2025{\natexlab{e}}.

\bibitem[Zhang et~al.(2026{\natexlab{c}})Zhang, Ma, Li, Liu, Zhou, Zhang, Zhang, Chao, Ji, and Zheng]{zhangmodality}
Yue Zhang, Yuexiao Ma, Guilin Li, Yuqi Liu, Jiaqi Zhou, Qingheng Zhang, Yan Zhang, Fei Chao, Rongrong Ji, and Xiawu Zheng.
\newblock Modality-aware quantization: Balancing visual and textual fidelity in multimodal compression, 2026{\natexlab{c}}.

\bibitem[Wang et~al.(2026)Wang, Jia, Zhou, Wang, Zhang, Dang, and Gu]{wang2026lqa}
Xin Wang, Hong Jia, Hualin Zhou, Sheng~Guang Wang, Yu~Zhang, Ting Dang, and Tao Gu.
\newblock Lqa: A lightweight quantized-adaptive framework for vision-language models on the edge.
\newblock \emph{arXiv preprint arXiv:2602.07849}, 2026.

\bibitem[Xue et~al.(2026)Xue, Huang, Shao, Zhu, Zhang, Li, and Zhang]{xue2025vlmq}
Yufei Xue, Yushi Huang, Jiawei Shao, Lunjie Zhu, Chi Zhang, Xuelong Li, and Jun Zhang.
\newblock Vlmq: Token saliency-driven post-training quantization for vision-language models.
\newblock \emph{arXiv preprint arXiv:2508.03351}, 2026.

\bibitem[Wang et~al.(2024{\natexlab{c}})Wang, Wang, Xu, Tang, Zhou, and Lu]{wang2024q}
Changyuan Wang, Ziwei Wang, Xiuwei Xu, Yansong Tang, Jie Zhou, and Jiwen Lu.
\newblock Q-vlm: Post-training quantization for large vision-language models.
\newblock \emph{Advances in Neural Information Processing Systems}, 37:\penalty0 114553--114573, 2024{\natexlab{c}}.

\bibitem[Das et~al.(2026)Das, La, Lau, Shrivastava, and Gwilliam]{das2026towards}
Gautom Das, Vincent La, Ethan Lau, Abhinav Shrivastava, and Matthew Gwilliam.
\newblock Towards understanding best practices for quantization of vision-language models.
\newblock \emph{arXiv preprint arXiv:2601.15287}, 2026.

\bibitem[Wang et~al.(2025{\natexlab{e}})Wang, Huang, Abdalla, Zhang, Xian, and Manocha]{wang2025bi}
Xijun Wang, Junyun Huang, Rayyan Abdalla, Chengyuan Zhang, Ruiqi Xian, and Dinesh Manocha.
\newblock Bi-vlm: Pushing ultra-low precision post-training quantization boundaries in vision-language models.
\newblock \emph{arXiv preprint arXiv:2509.18763}, 2025{\natexlab{e}}.

\bibitem[Qin et~al.(2026)Qin, Li, Chen, Zhang, Kong, and Zhang]{qin2026veq}
Guangshuo Qin, Zhiteng Li, Zheng Chen, Weihang Zhang, Linghe Kong, and Yulun Zhang.
\newblock Veq: Modality-adaptive quantization for moe vision-language models.
\newblock \emph{arXiv preprint arXiv:2602.01037}, 2026.

\bibitem[Xu et~al.(2025{\natexlab{c}})Xu, Nguyen, Mukherjee, Bagchi, Chaterji, Liang, and Li]{xu2025learning}
Zhuoyan Xu, Khoi~Duc Nguyen, Preeti Mukherjee, Saurabh Bagchi, Somali Chaterji, Yingyu Liang, and Yin Li.
\newblock Learning to inference adaptively for multimodal large language models.
\newblock In \emph{Proceedings of the IEEE/CVF International Conference on Computer Vision}, pages 3552--3563, 2025{\natexlab{c}}.

\bibitem[Cahyani et~al.(2025)Cahyani, Suartana, and Yudistira]{cahyani2025input}
Putu Indah~Githa Cahyani, Komang David~Dananjaya Suartana, and Novanto Yudistira.
\newblock Input-adaptive visual preprocessing for efficient fast vision-language model inference.
\newblock \emph{arXiv preprint arXiv:2512.20839}, 2025.

\bibitem[He and Chen(2026)]{he2026energy}
Jialuo He and Huangxun Chen.
\newblock Energy-driven adaptive visual token pruning for efficient vision-language models.
\newblock \emph{arXiv preprint arXiv:2603.05950}, 2026.

\bibitem[Debnath et~al.(2026)Debnath, Manh, Liu, and Wang]{debnath2026llmind}
Soumyaratna Debnath, Bui~Duc Manh, Zinan Liu, and Lin Wang.
\newblock Llmind: Bio-inspired training-free adaptive visual representations for vision-language models.
\newblock \emph{arXiv preprint arXiv:2603.14882}, 2026.

\bibitem[Zhang et~al.(2025{\natexlab{e}})Zhang, Zhu, Li, Tao, Liu, and Liu]{zhang2025adaptinfer}
Weichen Zhang, Zhui Zhu, Ningbo Li, Shilong Tao, Kebin Liu, and Yunhao Liu.
\newblock Adaptinfer: Adaptive token pruning for vision-language model inference with dynamical text guidance.
\newblock \emph{arXiv preprint arXiv:2508.06084}, 2025{\natexlab{e}}.

\bibitem[Liang et~al.(2025{\natexlab{b}})Liang, Guan, Lu, Chen, Wang, and Hu]{liang2025dynamic}
Xiaoyu Liang, Chaofeng Guan, Jiaying Lu, Huiyao Chen, Huan Wang, and Haoji Hu.
\newblock Dynamic token reduction during generation for vision language models.
\newblock \emph{arXiv preprint arXiv:2501.14204}, 2025{\natexlab{b}}.

\bibitem[Zhao et~al.(2024{\natexlab{b}})Zhao, Wang, Ouyang, Dong, Wang, and He]{zhao2024hallucinationsenhancinglvlmshallucinationaware}
Zhiyuan Zhao, Bin Wang, Linke Ouyang, Xiaoyi Dong, Jiaqi Wang, and Conghui He.
\newblock Beyond hallucinations: Enhancing lvlms through hallucination-aware direct preference optimization.
\newblock \emph{arXiv preprint arXiv:2311.16839}, 2024{\natexlab{b}}.

\bibitem[Goyal et~al.(2017)Goyal, Khot, Summers-Stay, Batra, and Parikh]{goyal2017vqav2}
Yash Goyal, Tejas Khot, Douglas Summers-Stay, Dhruv Batra, and Devi Parikh.
\newblock Making the v in vqa matter: Elevating the role of image understanding in visual question answering.
\newblock In \emph{Proceedings of the IEEE conference on computer vision and pattern recognition}, pages 6904--6913, 2017.

\bibitem[Li et~al.(2024{\natexlab{c}})Li, Ge, Ge, Wang, Wang, Zhang, and Shan]{li2024seedbench}
Bohao Li, Yuying Ge, Yixiao Ge, Guangzhi Wang, Rui Wang, Ruimao Zhang, and Ying Shan.
\newblock Seed-bench: Benchmarking multimodal large language models.
\newblock In \emph{Proceedings of the IEEE/CVF Conference on Computer Vision and Pattern Recognition}, pages 13299--13308, 2024{\natexlab{c}}.

\bibitem[Liu et~al.(2024{\natexlab{c}})Liu, Duan, Zhang, Li, Zhang, Zhao, Yuan, Wang, He, Liu, et~al.]{liu2023mmbench}
Yuan Liu, Haodong Duan, Yuanhan Zhang, Bo~Li, Songyang Zhang, Wangbo Zhao, Yike Yuan, Jiaqi Wang, Conghui He, Ziwei Liu, et~al.
\newblock Mmbench: Is your multi-modal model an all-around player?
\newblock In \emph{European conference on computer vision}, pages 216--233. Springer, 2024{\natexlab{c}}.

\bibitem[Chen et~al.(2024{\natexlab{d}})Chen, Li, Dong, Zhang, Zang, Chen, Duan, Wang, Qiao, Lin, et~al.]{chen2024mmstar}
Lin Chen, Jinsong Li, Xiaoyi Dong, Pan Zhang, Yuhang Zang, Zehui Chen, Haodong Duan, Jiaqi Wang, Yu~Qiao, Dahua Lin, et~al.
\newblock Are we on the right way for evaluating large vision-language models?
\newblock \emph{Advances in Neural Information Processing Systems}, 37:\penalty0 27056--27087, 2024{\natexlab{d}}.

\bibitem[Yue et~al.(2024)Yue, Ni, Zhang, Zheng, Liu, Zhang, Stevens, Jiang, Ren, Sun, et~al.]{yue2024mmmu}
Xiang Yue, Yuansheng Ni, Kai Zhang, Tianyu Zheng, Ruoqi Liu, Ge~Zhang, Samuel Stevens, Dongfu Jiang, Weiming Ren, Yuxuan Sun, et~al.
\newblock Mmmu: A massive multi-discipline multimodal understanding and reasoning benchmark for expert agi.
\newblock In \emph{Proceedings of the IEEE/CVF conference on computer vision and pattern recognition}, pages 9556--9567, 2024.

\bibitem[Lu et~al.(2023)Lu, Bansal, Xia, Liu, Li, Hajishirzi, Cheng, Chang, Galley, and Gao]{lu2023mathvista}
Pan Lu, Hritik Bansal, Tony Xia, Jiacheng Liu, Chunyuan Li, Hannaneh Hajishirzi, Hao Cheng, Kai-Wei Chang, Michel Galley, and Jianfeng Gao.
\newblock Mathvista: Evaluating mathematical reasoning of foundation models in visual contexts.
\newblock \emph{arXiv preprint arXiv:2310.02255}, 2023.

\bibitem[Li et~al.(2023{\natexlab{c}})Li, Du, Zhou, Wang, Zhao, and Wen]{li2023evaluatingobjecthallucinationlarge}
Yifan Li, Yifan Du, Kun Zhou, Jinpeng Wang, Wayne~Xin Zhao, and Ji-Rong Wen.
\newblock Evaluating object hallucination in large vision-language models.
\newblock \emph{arXiv preprint arXiv:2305.10355}, 2023{\natexlab{c}}.

\bibitem[Liu et~al.(2024{\natexlab{d}})Liu, Li, Huang, Yang, Yu, Li, Yin, Liu, Jin, and Bai]{liu2024ocrbench}
Yuliang Liu, Zhang Li, Mingxin Huang, Biao Yang, Wenwen Yu, Chunyuan Li, Xu-Cheng Yin, Cheng-Lin Liu, Lianwen Jin, and Xiang Bai.
\newblock Ocrbench: on the hidden mystery of ocr in large multimodal models.
\newblock \emph{Science China Information Sciences}, 67\penalty0 (12):\penalty0 220102, 2024{\natexlab{d}}.

\bibitem[Ghosh et~al.(2023)Ghosh, Hajishirzi, and Schmidt]{ghosh2024geneval}
Dhruba Ghosh, Hannaneh Hajishirzi, and Ludwig Schmidt.
\newblock Geneval: An object-focused framework for evaluating text-to-image alignment.
\newblock \emph{Advances in Neural Information Processing Systems}, 36:\penalty0 52132--52152, 2023.

\bibitem[Hu et~al.(2024)Hu, Wang, Fang, Fu, Cheng, and Yu]{hu2024dpgbench}
Xiwei Hu, Rui Wang, Yixiao Fang, Bin Fu, Pei Cheng, and Gang Yu.
\newblock Ella: Equip diffusion models with llm for enhanced semantic alignment.
\newblock \emph{arXiv preprint arXiv:2403.05135}, 2024.

\bibitem[Huang et~al.(2023)Huang, Sun, Xie, Li, and Liu]{huang2023t2icompbench}
Kaiyi Huang, Kaiyue Sun, Enze Xie, Zhenguo Li, and Xihui Liu.
\newblock T2i-compbench: A comprehensive benchmark for open-world compositional text-to-image generation.
\newblock \emph{Advances in Neural Information Processing Systems}, 36:\penalty0 78723--78747, 2023.

\bibitem[Heusel et~al.(2017)Heusel, Ramsauer, Unterthiner, Nessler, and Hochreiter]{heusel2017gans}
Martin Heusel, Hubert Ramsauer, Thomas Unterthiner, Bernhard Nessler, and Sepp Hochreiter.
\newblock Gans trained by a two time-scale update rule converge to a local nash equilibrium.
\newblock \emph{Advances in neural information processing systems}, 30, 2017.

\bibitem[Hessel et~al.(2021)Hessel, Holtzman, Forbes, Le~Bras, and Choi]{hessel2021clipscore}
Jack Hessel, Ari Holtzman, Maxwell Forbes, Ronan Le~Bras, and Yejin Choi.
\newblock Clipscore: A reference-free evaluation metric for image captioning.
\newblock In \emph{Proceedings of the 2021 conference on empirical methods in natural language processing}, pages 7514--7528, 2021.

\bibitem[Conneau et~al.(2023)Conneau, Ma, Khanuja, Zhang, Axelrod, Dalmia, Riesa, Rivera, and Bapna]{conneau2023fleurs}
Alexis Conneau, Min Ma, Simran Khanuja, Yu~Zhang, Vera Axelrod, Siddharth Dalmia, Jason Riesa, Clara Rivera, and Ankur Bapna.
\newblock Fleurs: Few-shot learning evaluation of universal representations of speech.
\newblock In \emph{2022 IEEE Spoken Language Technology Workshop (SLT)}, pages 798--805. IEEE, 2023.

\bibitem[Huang et~al.(2026)Huang, Cooper, and Toda]{huang2026mos}
Wen-Chin Huang, Erica Cooper, and Tomoki Toda.
\newblock Mos-bench: Benchmarking generalization abilities of subjective speech quality assessment models.
\newblock \emph{IEEE Transactions on Audio, Speech and Language Processing}, 2026.

\bibitem[Yan et~al.(2026)Yan, Chen, Liu, Ma, Lin, Wen, Xie, Wu, Liang, Zhao, et~al.]{yan2026soulx}
Ruiqi Yan, Wenxi Chen, Zhanxun Liu, Ziyang Ma, Haopeng Lin, Hanlin Wen, Hanke Xie, Jun Wu, Yuzhe Liang, Yuxiang Zhao, et~al.
\newblock Soulx-duplug: Plug-and-play streaming state prediction module for realtime full-duplex speech conversation.
\newblock \emph{arXiv preprint arXiv:2603.14877}, 2026.

\bibitem[Lin et~al.(2025{\natexlab{b}})Lin, Lian, Li, Wang, Anumanchipalli, Liu, and Lee]{lin2025full}
Guan-Ting Lin, Jiachen Lian, Tingle Li, Qirui Wang, Gopala Anumanchipalli, Alexander~H Liu, and Hung-yi Lee.
\newblock Full-duplex-bench: A benchmark to evaluate full-duplex spoken dialogue models on turn-taking capabilities.
\newblock \emph{arXiv preprint arXiv:2503.04721}, 2025{\natexlab{b}}.

\bibitem[Fu et~al.(2025)Fu, Dai, Luo, Li, Ren, Zhang, Wang, Zhou, Shen, Zhang, et~al.]{fu2025video}
Chaoyou Fu, Yuhan Dai, Yongdong Luo, Lei Li, Shuhuai Ren, Renrui Zhang, Zihan Wang, Chenyu Zhou, Yunhang Shen, Mengdan Zhang, et~al.
\newblock Video-mme: The first-ever comprehensive evaluation benchmark of multi-modal llms in video analysis.
\newblock In \emph{Proceedings of the IEEE/CVF conference on computer vision and pattern recognition}, pages 24108--24118, 2025.

\bibitem[Mangalam et~al.(2023)Mangalam, Akshulakov, and Malik]{mangalam2023egoschema}
Karttikeya Mangalam, Raiymbek Akshulakov, and Jitendra Malik.
\newblock Egoschema: A diagnostic benchmark for very long-form video language understanding.
\newblock \emph{Advances in Neural Information Processing Systems}, 36:\penalty0 46212--46244, 2023.

\bibitem[Li et~al.(2024{\natexlab{d}})Li, Wang, He, Li, Wang, Liu, Wang, Xu, Chen, Luo, et~al.]{li2024mvbench}
Kunchang Li, Yali Wang, Yinan He, Yizhuo Li, Yi~Wang, Yi~Liu, Zun Wang, Jilan Xu, Guo Chen, Ping Luo, et~al.
\newblock Mvbench: A comprehensive multi-modal video understanding benchmark.
\newblock In \emph{Proceedings of the IEEE/CVF Conference on Computer Vision and Pattern Recognition}, pages 22195--22206, 2024{\natexlab{d}}.

\bibitem[Patraucean et~al.(2023)Patraucean, Smaira, Gupta, Recasens, Markeeva, Banarse, Koppula, Malinowski, Yang, Doersch, et~al.]{patraucean2023perception}
Viorica Patraucean, Lucas Smaira, Ankush Gupta, Adria Recasens, Larisa Markeeva, Dylan Banarse, Skanda Koppula, Mateusz Malinowski, Yi~Yang, Carl Doersch, et~al.
\newblock Perception test: A diagnostic benchmark for multimodal video models.
\newblock \emph{Advances in Neural Information Processing Systems}, 36:\penalty0 42748--42761, 2023.

\bibitem[Wu et~al.(2024{\natexlab{b}})Wu, Li, Chen, and Li]{wu2024longvideobench}
Haoning Wu, Dongxu Li, Bei Chen, and Junnan Li.
\newblock Longvideobench: A benchmark for long-context interleaved video-language understanding.
\newblock \emph{Advances in Neural Information Processing Systems}, 37:\penalty0 28828--28857, 2024{\natexlab{b}}.

\bibitem[Zhou et~al.(2025)Zhou, Shu, Zhao, Wu, Liang, Xiao, Qin, Yang, Xiong, Zhang, et~al.]{zhou2025mlvu}
Junjie Zhou, Yan Shu, Bo~Zhao, Boya Wu, Zhengyang Liang, Shitao Xiao, Minghao Qin, Xi~Yang, Yongping Xiong, Bo~Zhang, et~al.
\newblock Mlvu: Benchmarking multi-task long video understanding.
\newblock In \emph{Proceedings of the IEEE/CVF Conference on Computer Vision and Pattern Recognition}, pages 13691--13701, 2025.

\bibitem[Niu et~al.(2025)Niu, Li, Miao, Ge, Zhou, He, Dong, Duan, Ding, Qian, et~al.]{niu2025ovo}
Junbo Niu, Yifei Li, Ziyang Miao, Chunjiang Ge, Yuanhang Zhou, Qihao He, Xiaoyi Dong, Haodong Duan, Shuangrui Ding, Rui Qian, et~al.
\newblock Ovo-bench: How far is your video-llms from real-world online video understanding?
\newblock In \emph{Proceedings of the Computer Vision and Pattern Recognition Conference}, pages 18902--18913, 2025.

\bibitem[Lin et~al.(2026)Lin, Fang, Chen, Cheng, Wan, Luo, Wang, Li, Liu, and Sun]{lin2026streamingbench}
Junming Lin, Zheng Fang, Chi Chen, Haoxuan Cheng, Zihao Wan, Fuwen Luo, Ziyue Wang, Peng Li, Yang Liu, and Maosong Sun.
\newblock Streamingbench: Assessing the gap for mllms to achieve streaming video understanding.
\newblock In \emph{ICASSP 2026-2026 IEEE International Conference on Acoustics, Speech and Signal Processing (ICASSP)}, pages 12147--12151. IEEE, 2026.

\bibitem[Wang et~al.(2025{\natexlab{f}})Wang, Wang, Chen, Wu, Zhao, and Zheng]{wang2025omnimmi}
Yuxuan Wang, Yueqian Wang, Bo~Chen, Tong Wu, Dongyan Zhao, and Zilong Zheng.
\newblock Omnimmi: A comprehensive multi-modal interaction benchmark in streaming video contexts.
\newblock In \emph{Proceedings of the IEEE/CVF Conference on Computer Vision and Pattern Recognition}, pages 18925--18935, 2025{\natexlab{f}}.

\bibitem[Soomro et~al.(2012)Soomro, Zamir, and Shah]{soomro2012ucf101}
Khurram Soomro, Amir~Roshan Zamir, and Mubarak Shah.
\newblock Ucf101: A dataset of 101 human actions classes from videos in the wild.
\newblock \emph{arXiv preprint arXiv:1212.0402}, 2012.

\bibitem[Carreira et~al.(2018)Carreira, Noland, Banki-Horvath, Hillier, and Zisserman]{carreira2018short}
Joao Carreira, Eric Noland, Andras Banki-Horvath, Chloe Hillier, and Andrew Zisserman.
\newblock A short note about kinetics-600.
\newblock \emph{arXiv preprint arXiv:1808.01340}, 2018.

\bibitem[Seedance et~al.(2026)Seedance, Chen, Chen, Chen, Chen, Chen, Chen, Cheng, Cheng, Cheng, et~al.]{seedance2026seedance}
Team Seedance, De~Chen, Liyang Chen, Xin Chen, Ying Chen, Zhuo Chen, Zhuowei Chen, Feng Cheng, Tianheng Cheng, Yufeng Cheng, et~al.
\newblock Seedance 2.0: Advancing video generation for world complexity.
\newblock \emph{arXiv preprint arXiv:2604.14148}, 2026.

\bibitem[{Arena AI}(2026)]{arenaai}
{Arena AI}.
\newblock {Arena AI}.
\newblock \url{https://arena.ai/}, 2026.
\newblock Accessed: 2026-05-20.

\bibitem[Sheng et~al.(2026)Sheng, Du, Zhang, Yan, and Chen]{sheng2025syncspeech}
Zhengyan Sheng, Zhihao Du, Shiliang Zhang, Zhijie Yan, and Liping Chen.
\newblock Syncspeech: Efficient and low-latency text-to-speech based on temporal masked transformer.
\newblock \emph{arXiv preprint arXiv:2502.11094}, 2026.

\bibitem[Gu et~al.(2025)Gu, Mao, and Shou]{gu2025long}
Yuchao Gu, Weijia Mao, and Mike~Zheng Shou.
\newblock Long-context autoregressive video modeling with next-frame prediction.
\newblock \emph{arXiv preprint arXiv:2503.19325}, 2025.

\bibitem[Liu et~al.(2026{\natexlab{b}})Liu, Guo, Li, Zhen, He, Ji, Ren, Zhang, Lu, and Liu]{liu2026thinking}
Zikang Liu, Longteng Guo, Handong Li, Ru~Zhen, Xingjian He, Ruyi Ji, Xiaoming Ren, Yanhao Zhang, Haonan Lu, and Jing Liu.
\newblock Thinking in streaming video.
\newblock \emph{arXiv preprint arXiv:2603.12938}, 2026{\natexlab{b}}.

\end{thebibliography}
